\documentclass[10pt,twocolumn,letterpaper]{article}

\usepackage{cvpr}
\usepackage{times}
\usepackage{epsfig}
\usepackage{graphicx}
\usepackage{amsmath}
\usepackage{amssymb}
\usepackage{multirow}
\usepackage{algorithm}
\usepackage{algorithmic}
\usepackage{booktabs}
\usepackage{caption}
\usepackage{makecell}

\usepackage[breaklinks=true,bookmarks=false]{hyperref}

\cvprfinalcopy 

\def\cvprPaperID{****} 

\begin{document}

\title{End-to-End Single Image Fog Removal using Enhanced Cycle Consistent Adversarial Networks}

\author{
Wei Liu,Xianxu Hou,Jiang Duan,Guoping Qiu
}

\maketitle

\begin{abstract}
  Single image defogging is a classical and challenging problem in computer vision. Existing methods towards this problem mainly include handcrafted priors based methods that rely on the use of the atmospheric degradation model and learning based approaches that require paired fog-fogfree training example images. In practice, however, prior-based methods are prone to failure due to their own limitations and paired training data are extremely difficult to acquire. Inspired by the principle of CycleGAN network, we have developed an end-to-end learning system that uses unpaired fog and fogfree training images, adversarial discriminators and cycle consistency losses to automatically construct a fog removal system. Similar to CycleGAN, our system has two transformation paths; one maps fog images to a fogfree image domain and the other maps fogfree images to a fog image domain. Instead of one stage mapping, our system uses a two stage mapping strategy in each transformation path to enhance the effectiveness of fog removal. Furthermore, we make explicit use of prior knowledge in the networks by embedding the atmospheric degradation principle and a sky prior for mapping fogfree images to the fog images domain. In addition, we also contribute the first real world nature fog-fogfree image dataset for defogging research. Our multiple real fog images dataset (MRFID) contains images of 200 natural outdoor scenes. For each scene, there are one clear image and corresponding four foggy images of different fog densities manually selected from a sequence of images taken by a fixed camera over the course of one year. Qualitative and quantitative comparison against several state-of-the-art methods on both synthetic and real world images demonstrate that our approach is effective and performs favorably for recovering a clear image from a foggy image.
\end{abstract}
\section{Introduction}
\begin{figure}[t]\footnotesize
	\begin{center}
		\begin{tabular}{@{}cc@{}}
			\includegraphics[width = 0.23\textwidth]{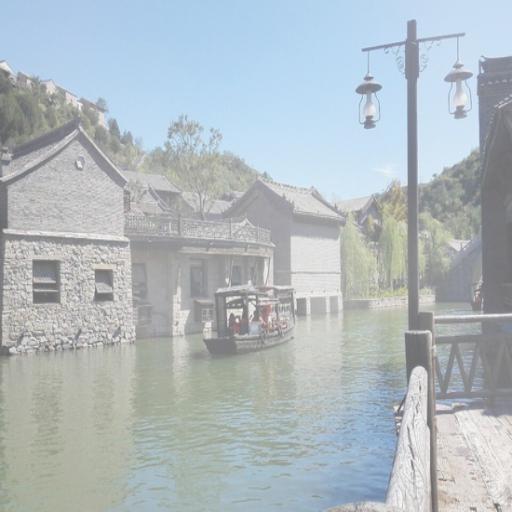} & \hspace{-0.4cm}
			\includegraphics[width = 0.23\textwidth]{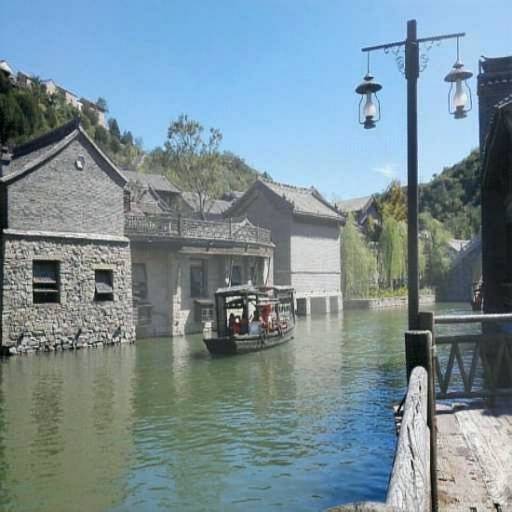}\\
			(a) & \hspace{-0.4cm}
			(b) \\
			\includegraphics[width = 0.23\textwidth]{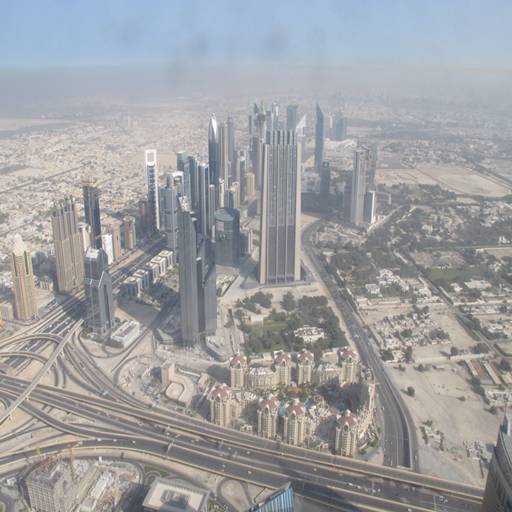} & \hspace{-0.4cm}
			\includegraphics[width = 0.23\textwidth]{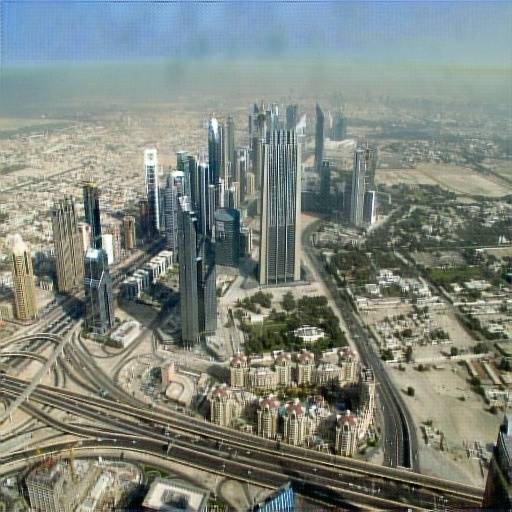} \\
			(c)  & \hspace{-0.4cm}
			(d) 
		\end{tabular}
	\end{center}
	\vspace{-0.5cm}
	\caption{Examples of image defogging results of the proposed Cycle-Defog2Refog method on both synthetic and real-world foggy images. (a) Synthetic foggy image. (b) The defogged result of (a). (c) Real foggy image. (d) The defogged result of (c).
	}
	\vspace{-0.1cm}
	\label{sample-results}
\end{figure}

Fog is an atmospheric phenomenon caused by very small particles in the air that obscure the clarity of the atmosphere. In computer vision, fog can cause serious degradation in image quality which in turn can  affect the performances of image analysis algorithms. Fog removal from a single image is a challenging problem which aims to recover a clean image from a given foggy image. Many defogging techniques in the literature are based on the atmospheric degradation model \cite{McCartneyModel}\cite{mittal2012TIP}\cite{nayar} which can be formulated as:
\begin{equation}
I(x) = J(x)T(x)+A[1-T(x)]
\end{equation}
where $J(x)$ denotes the recovered image, $I(x)$ is the observed hazy image, while the transmission map is denoted by $T(x)$, and \textit{A} corresponds to the atmospheric light. Due to the unknown parameters \textit{A}, \textit{T} and \textit{J}, single image defogging is a mathematically ill-posed problem when we want to recover \textit{J} from \textit{I}. Thus, most single image defogging methods try to estimate the atmospheric light \textit{A} and transmission \textit{T} in order to recover a foggy image by using this model.

There are mainly two classes of defogging methods, prior-based and learning-based. Prior-based methods\cite{he2011PAMI} \cite{zhu2015TIP} \cite{fattal2014colorline} \cite{ancuti2010ACCV} \cite{Non-local2016CVPR} obtain fog-related features through observation and statistics. It can be also called hand-crafted priors. However, these methods are usually under strict constraint conditions which can result in undesired fog artifacts in the recovered images. For example, He \textit{et.al.} \cite{he2011PAMI} assume that in a clear natural image, at least one channel in the RGB space is close to zero. This method may fail when dealing with scene objects which are similar to atmospheric light, such as sky or white building. Recently, deep learning-based methods \cite{cai2016dehazenet}
\cite{MSCNN2016ECCV} \cite{Aod-net2017ICCV} \cite{denselynet2018CVPR} are proposed to address the disadvantages of methods based on hand-crafted priors. They exploit the Convolutional Neural Network (CNN) to estimate the transmission and atmospheric light. However, these methods still use the defogging model (1) to recover the clear images. When the parameters of the model are not estimated accurately, defogging can introduce artifacts such as  color distortion and halo. Furthermore, these methods require fog-fogfree image pairs to train their networks. Thus, they have to use the model in Equation (1) and labeled depth maps such as those from the NYU depth dataset\cite{NYU2012ECCV} to synthesize fog-fogfree training image pairs. Synthesizing the data requires the depth maps of the fogfree images which are not always available, particularly for the outdoor scenes. For example, the NYU data \cite{NYU2012ECCV} often used in the defogging literature is an indoor dataset and models trained with indoor data are not best suited for outdoor scenes. 

\begin{figure*}[t]\footnotesize
	\begin{center}
		\begin{tabular}{@{}c@{}}
			\includegraphics[width = 0.85\textwidth]{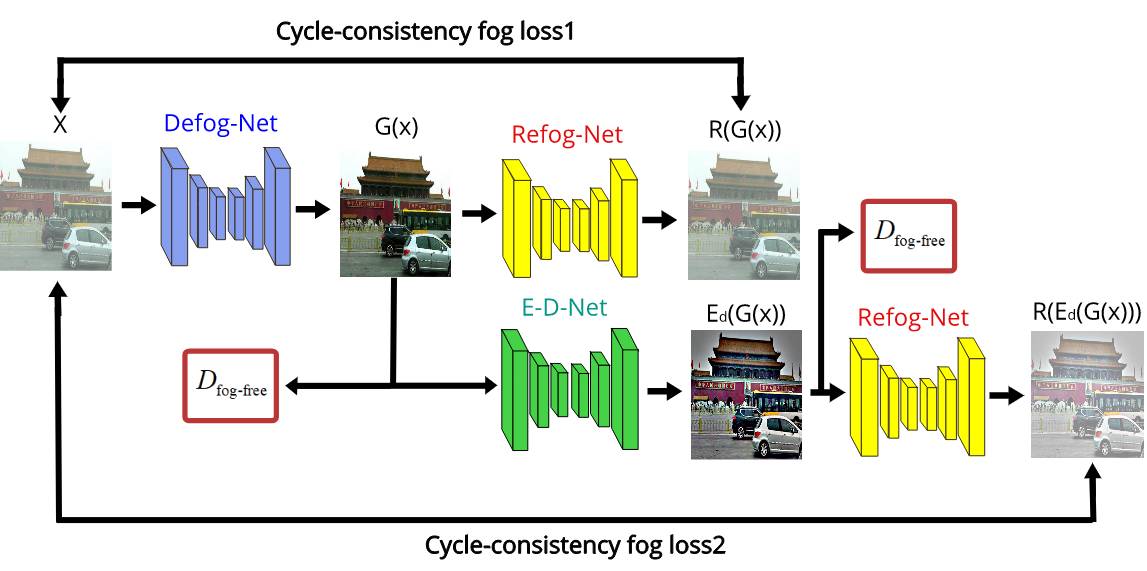}\\
			(a) Defog architecture\\
			\includegraphics[width = 0.85\textwidth]{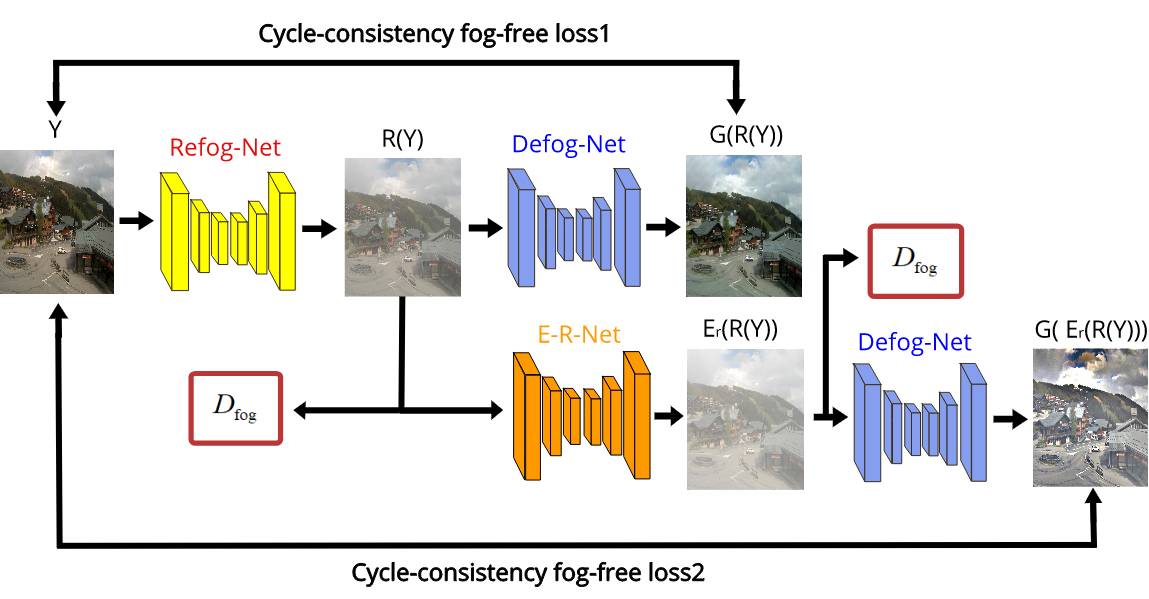}\\
			(b)  Refog architecture
		\end{tabular}
	\end{center}
	\vspace{-0.5cm}
	\caption{The architectures of Cycle-Defog2Refog network. X is the input foggy image. \textit{G} denotes the generator Defog-Net, and \textit{G}(X) is the defogged image. Y is the clear image. \textit{R} denotes the generator Refog-Net, and \textit{R}(Y) is the synthetic foggy image. $E_{d}$ denotes the generator Enhancer-Defog-Net(E-D-Net), which enhances the defogged image. $E_{r}$ denotes the generator Enhancer-Refog-Net(E-R-Net), which enhances the synthetic image. $D_{fog}$ is the adversarial discriminator for classifying the real foggy image and the generated foggy image. $D_{fogfree}$ is the adversarial  discriminator for distinguishing the real fog-free image and the defogged image.
	}
	\vspace{-0.5cm}
	\label{refog2defog-structure}
\end{figure*}

In this paper, we present a novel fog removal system based on recent developments in adversarial generative network (GAN). Specifically, inspired by the principle of CycleGAN \cite{CycleGAN2017ICCV}, we have developed an end-to-end learning system that uses unpaired fog and fogfree training images, adversarial discriminators and cycle consistency losses to automatically construct a defogging system. Similar to CycleGAN, our system has two transformation paths, one maps fog images to a fogfree image domain and the other maps fogfree images to a fog image domain. Instead of one stage mapping, our system uses a two stage mapping stategy in each transformation path (as shown in Figure \ref{refog2defog-structure}). In the fog to fogfree transformation path, fog images are mapped to a first fogfree domain. As the output of the first stage mapping may still contain residual fog, they are passed onto a second mapping network to remove the fog further. Similarly, in the fogfree to fog transformation path, fogfree images are transformed to a fog image domain first and the results are then passed onto a second transformation network to the fog image domain.  In constructing the fogfree to fog domain transformation, we explicitly embed the atmospheric degradation model (1) in the learning process.  Furthermore, a sky prior is introduced to reduce artifacts. We present experimental results on synthetic foggy and natural foggy image datasets to show the effectiveness of the new defogging technique. 

To our knowledge, there is currently no real world fog-fogfree nature image dataset suitable for defogging research publicly available. We believe using clear and foggy images that occur naturally is valuable for developing practically useful defogging technologies. In this work, we have collected clear and foggy images of 200 natural outdoor scenes. Each scene was imaged by a fixed camera over the course of one year. One clear image and four foggy images of different fog intensities were manually selected to form the Multiple Real Fog Images Dataset (MRFID). We believe this dataset will be useful for the research community and we will make this dataset publicly available to researchers to facilitate the development of defogging techniques. The dataset will be made available online in due course.

The organization of the paper is as follows. In Section 2, we briefly review relevant literature. In Section 3, we present the new cycle fog to fogfree learning framework, its components, learning cost functions and training procedure. Section 4 presents experimental results and Section 5 concludes the paper.

%

\section{Related work}

In this section, we mainly review relevant single image defogging methods, which can be roughly grouped as prior-based methods and learning-based methods.

\textbf{Prior-based methods.} These methods have been widely used in the past few years and are also known as hand-crafted feature based methods. These methods often leverage the statistics of the natural image to characterize the transmission map, such as dark-channel prior \cite{he2011PAMI}, color attenuation prior \cite{zhu2015TIP}, contrast color-lines \cite{fattal2014colorline}, hue disparity prior \cite{ancuti2010ACCV} and haze-line prior \cite{Non-local2016CVPR}. Particularly, the method of the dark-channel has shown its excellent defogging performance, which has led many researchers to improve this method to achieve single image defogging. Despite the remarkable defogging performance by these methods, hand-crafted features (such as textural, contrast and so on.) also have limitations. For instance, dark-channel prior \cite{he2011PAMI} does not work well for some scene objects (such as sky, white building and so on.) which are inherently similar to the atmospheric light. Using haze-line prior \cite{Non-local2016CVPR} can cause color distortion when the fog density is high.

\textbf{Learning-based methods.} Recently, some learning-based methods have drawn significant attention in the defogging reseach community. Tang \textit{et.al.} \cite{tang2014CVPR} proposed a method by using the random forest methods to train dark primary colors and other multiple color features to improve the estimation accuracy of transmittance. Mai \textit{et.al.} \cite{mai2014ROBIO} found that the RGB color feature of the haze image had a strong linear relationship with the depth of the scene, and established the intrinsic relation between the color feature and the scene depth through the back propagation neural network to effectively restore the scene depth. Cai \textit{et.al.} \cite{cai2016dehazenet} proposed a concept of a dehazing network that used a convolutional neural network to train the color characteristics (such as dark primary colors, color fading, maximum contrast, etc.) of foggy images and to optimize the transmission. All of those methods can achieve better defogging effect. However, they still have to estimate the transmission map and atmospheric light first, and then remove the fog with the atmospheric degradation model. Thus, the artifacts could not be avoided in the final defogged results when the transmission or atmospheric light is wrongly estimated.

To address the above problem, networks based on encoder-decoder structure \cite{cGAN2018dehazing} \cite{Aod-net2017ICCV} \cite{denselynet2018CVPR} have been used to directly recover clear images. Among these methods, generative adversarial network (GAN) \cite{GAN2014NIPS} based defogging algorithms have achieved remarkable results. Li \textit{et.al.} \cite{cGAN2018dehazing} modify the basic GAN to directly restore a clear image from a foggy image. However, all these methods required fog-fogfree pair images to train the network. In practice, it is difficult to obtain a large number of paired fog-fogfree images. A method based on CycleGAN \cite{CycleGAN2017ICCV} has been proposed in \cite{Cycle-dehazing2018} where cycle-consistency and VGG perceptual losses are used to directly remove fog. A significant advantage of using CycleGAN is that there is no need to use paired fog-fogfree images to train the system.
	
\section{Cycle-Defog2Refog Network}

In this section, we introduce the details of the proposed network. It consists of two parts, a defog architecture and a refog architecture, as shown in Figure \ref{refog2defog-structure}. In the defog architecture, we use a refog-net (\textit{R}) and an enhancer-defog-net ($E_{d}$) to constrain the defogging mapping function with two consistency fog loss functions and an adversarial discriminator $D_{fog-free}$, as shown in Figure \ref{refog2defog-structure} (a). In the refog architecture, a defog-net (\textit{G}) and an enhancer-refog-net($E_{r}$) are used to supervise the refogging mapping function with two consistency fog-free loss functions and an adversarial discriminator $D_{fog}$, as shown in Figure \ref{refog2defog-structure} (b). 
\subsection{Defog-net}
\begin{figure}[t]\footnotesize
	\begin{center}
		\begin{tabular}{@{}c@{}}
			\includegraphics[width = 0.35\textwidth]{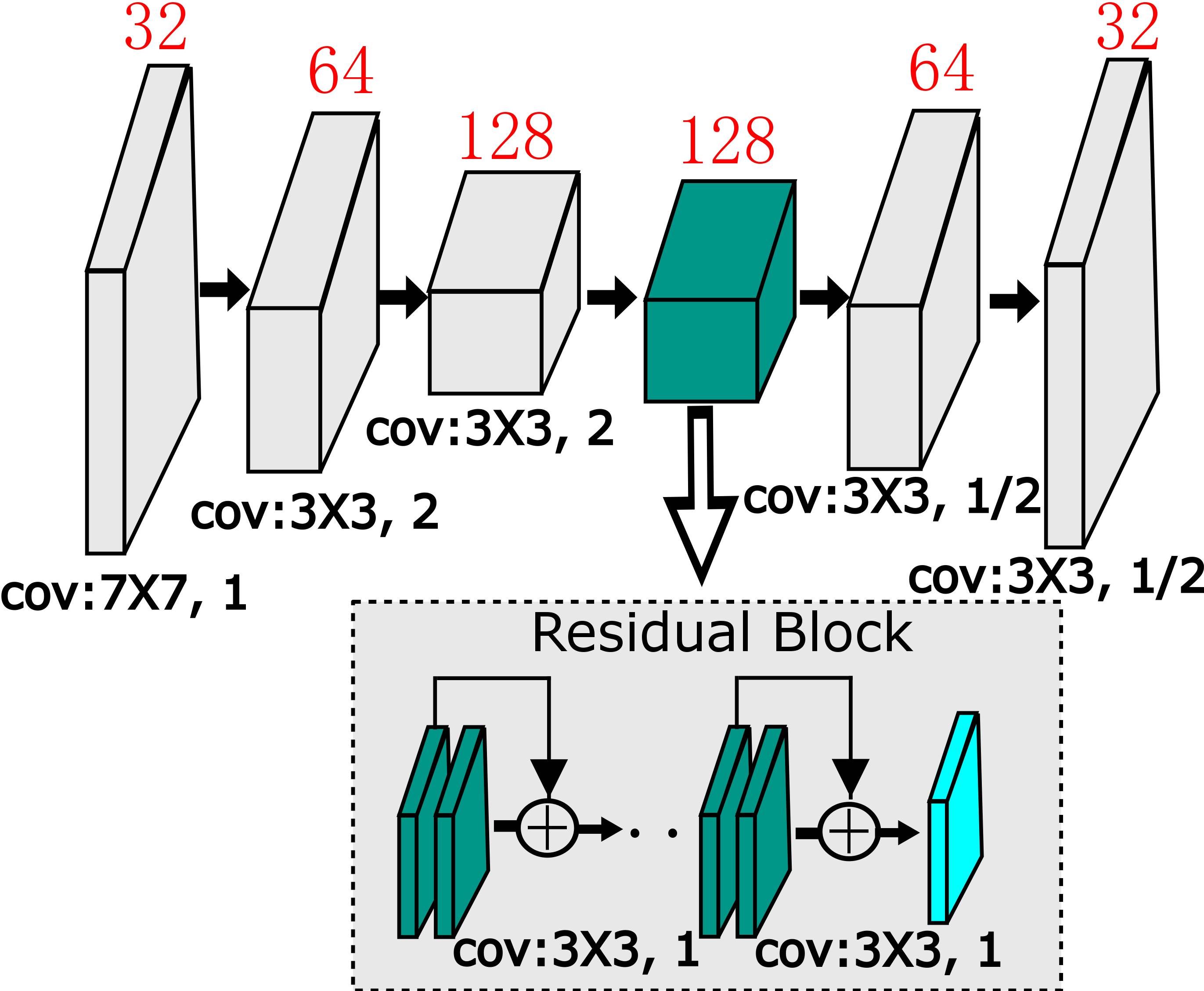}
		\end{tabular}
	\end{center}
	\vspace{-0.5cm}
	\caption{The defog network architecture. Here, (Conv:3 $\times$ 3,1/2) denotes a convolutional layer with kernel size of 3 and stride of 1/2.
	}
	\vspace{-0.5cm}
	\label{defog-structure}
\end{figure}

The details of the defog network are shown in Figure \ref{defog-structure}. For the generator \textit{G}, we adopt an encoder-decoder network from Johnson \cite{johnson2016ECCV}. We use 3 convolutional blocks in the encoding process and 3 deconvolutional blocks in the decoding process. In the encoder network, the first layer has 32 filters with a kernel size 7$\times$7 and a stride of 1, the second layer has 64 filters with a kernel size of 3$\times$3 and a stride of 2, and the third layer has 128 filters with a kernel size of 3$\times$3 and a stride of 2. Each layer of the decoder network has the same number of filters as its symmetric layer in the encoder but their convolutions have a stride of 1/2. Moreover, similar to CycleGAN, we have used 9 residual blocks \cite{resnet2016CVPR}.
%
\subsection{Refog-net}
In the traditional CycleGAN, we can use a CNN to directly generate the foggy images to contrain the defog mapping function. However, in practice, it is extemely difficult to fit the distribution of foggy image by only using a convolutional neural network due to the diversity and complexity of a variety of foggy image contents. Instead, we introduce a CNN based atmospheric degradation model to synthesize foggy images. Specifically, we use a CNN to estimate the transmission map \textit{T} and use a sky prior to estimate the atmospheric light \textit{A} . 

\begin{figure}[t]\footnotesize
	\begin{center}
		\begin{tabular}{@{}c@{}}
			\includegraphics[width = 0.25\textwidth]{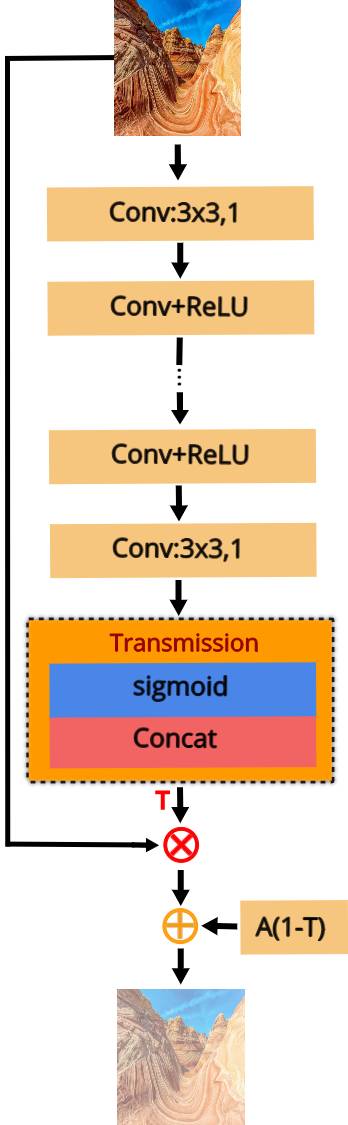}
		\end{tabular}
	\end{center}
	\vspace{-0.5cm}
	\caption{The details of the refog network. Here, (Conv:3 $\times$ 3,1) denotes a convolutional layer with kernel size of 3 and stride of 1. \textit{T} denotes the transimission map. \textit{A} denotes the atmospheric light. 
	}
	\vspace{-0.5cm}
	\label{refog-details}
\end{figure}

The details of the refog network are shown in Figure \ref{refog-details}. We introduce a CNN to estimate the transmission map \textit{T}. Each layer of this network has 64 filters with a kernel size 3$\times$3 and a stride of 1. In our network, we use 5 layers to estimate the transmission \textit{T}. 

\begin{figure*}[t]\footnotesize
	\begin{center}
		\begin{tabular}{@{}c@{}}
			\includegraphics[width = 0.85\textwidth]{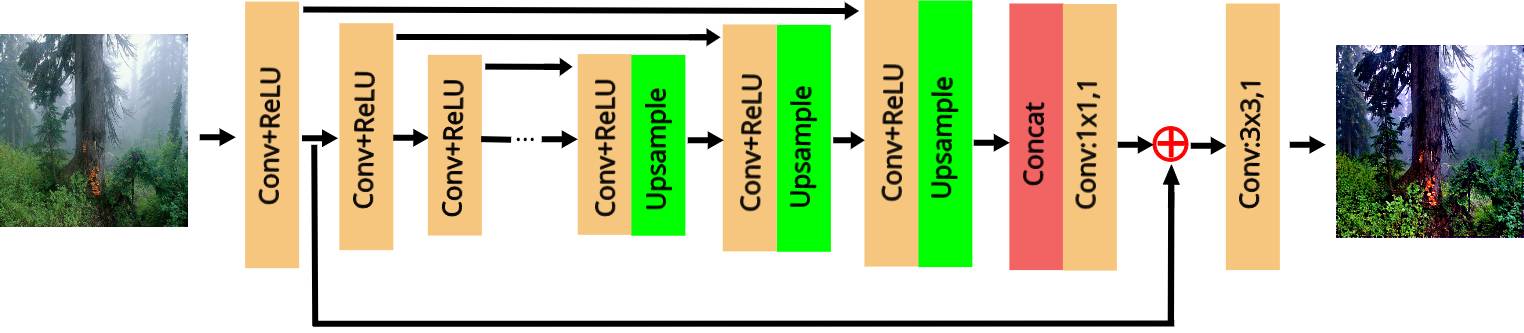}
		\end{tabular}
	\end{center}
	\vspace{-0.5cm}
	\caption{The architecture of E-Net.
	}
	\vspace{-0.5cm}
	\label{enhanceNet-structure}
\end{figure*}
In addition to \textit{T}, we need to estimate \textit{A} in order to use the atmospheric degradation model (1). It is well known that the more accurately \textit{A} is estimated, the better a defogging performance can be obtained. He \textit{et.al.} \cite{he2011PAMI} first selected the top 0.1 percent brightest pixels in the dark channel of the foggy image and  then took the maximum of these top pixels as \textit{A}. Tan \textit{et.al.}\cite{tan2008CVPR} picked the highest intensity of foggy image as the atmospheric light value. However, taking the maximum value of the pixels as \textit{A} is prone to color distortion or halo artifact in the defogged results. To address this problem, we first segment the sky region from the foggy images and then take the average pixel values of the sky region as \textit{A}. The depth of the sky in an image is regarded as infinity, i.e., 
\begin{equation}
d_{sky}(x) \to +\infty
\end{equation}

In model (1), the transmission $T(x)$ is usually described as the portion of light which reaches the camera from the object and can be expressed as follows \cite{McCartneyModel}:
\begin{equation}
T(x) = e^{-\beta d(x)}
\end{equation}
where $\beta$ is the scattering coefficient of the atmosphere and it is always assumed constant. $d(x)$ is the depth from scene point to the camera. Substitute equation (2) into (3) and calculate the limit of both sides:
\begin{equation}
T_{sky}(x) \thickapprox 0
\end{equation}

Substitute equation (4) into (1):
\begin{equation}
I(x) \thickapprox A
\end{equation}

The value of \textit{A} in equation (5) can be considered as the intensity of pixels in the area of maximum fog density \cite{tan2008CVPR}. Thus, it is reasonable to take the average value of the sky region as \textit{A}. Then, the atmospheric light can be calculated as:
\begin{equation}
	A_{sky} = \mathop{mean}\limits_{c \in \{r,g,b\}}(I_{sky}^{c}(x)) 
\end{equation}
where $I_{sky}^{c}$ is a color channel of the sky region in a foggy image, $\mathop{mean}\limits_{c \in \{r,g,b\}}$ is an average filter to process each pixel for each RGB channel. Moreover, for the sky segmentation algorithm, we can choose the method of image matting \cite{closed-matting2008PAMI}\cite{fast-matting2010CVPR}\cite{poisson_matting2004ToG} or OSTU \cite{ostu1979threshold}. For a foggy image with no sky region or few sky region, we take the atmospheric light value \textit{A} follows the method of He \cite{he2011PAMI}. As shown in Figure  \ref{refog-details}, when the \textit{T} and \textit{A} are obtained, we synthesize the foggy image by using atmospheric degradation model (1). 

\subsection{E-Net}
\begin{figure}[t]\footnotesize
	\begin{center}
		\begin{tabular}{@{}ccc@{}}
			\includegraphics[width = 0.16\textwidth]{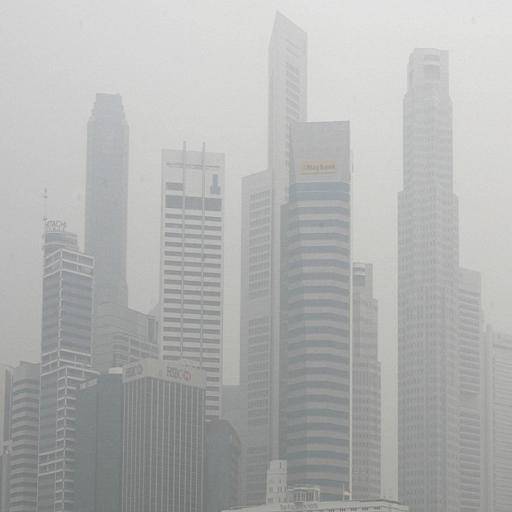} & \hspace{-0.4cm}
			\includegraphics[width = 0.16\textwidth]{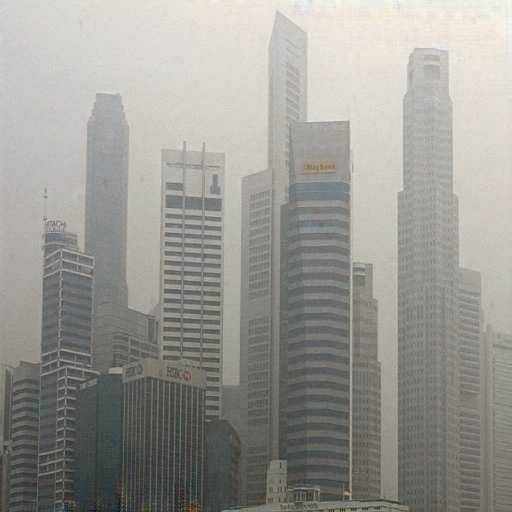} & \hspace{-0.4cm}
			\includegraphics[width = 0.16\textwidth]{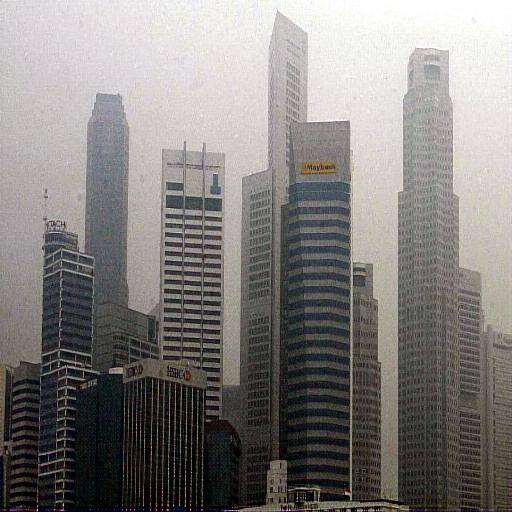} \\
			(a) Foggy input& \hspace{-0.4cm}
			(b) without E-net& \hspace{-0.4cm}
			(c) with E-net 
		\end{tabular}
	\end{center}
	\begin{center}
	\begin{tabular}{@{}ccc@{}}
		\includegraphics[width = 0.165\textwidth]{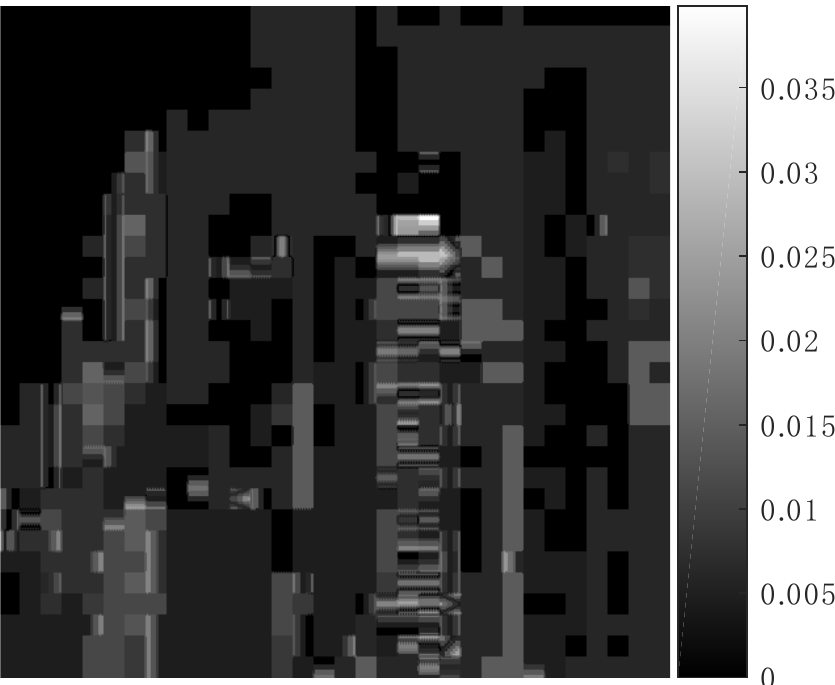} & \hspace{-0.4cm}
		\includegraphics[width = 0.165\textwidth]{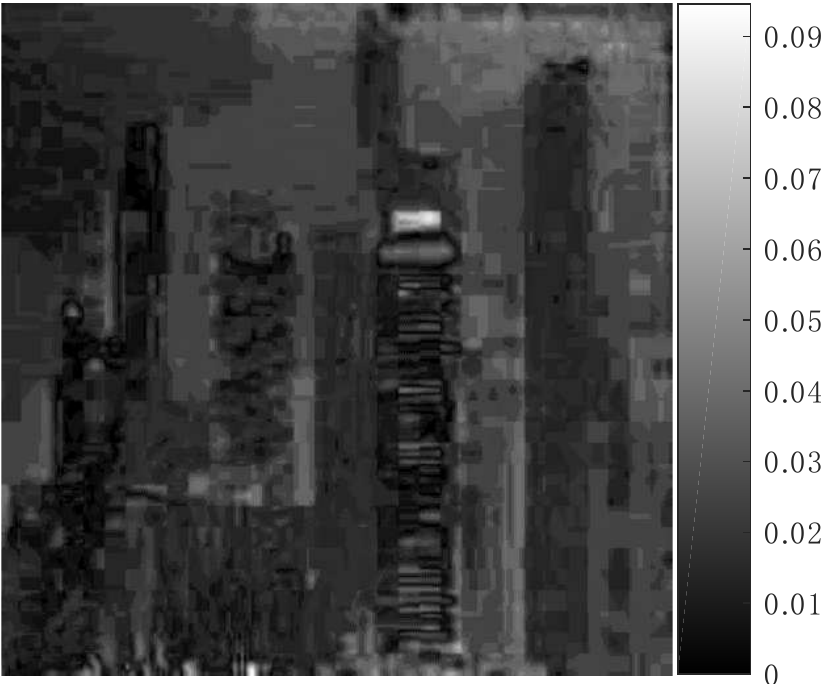} & \hspace{-0.4cm}
		\includegraphics[width = 0.165\textwidth]{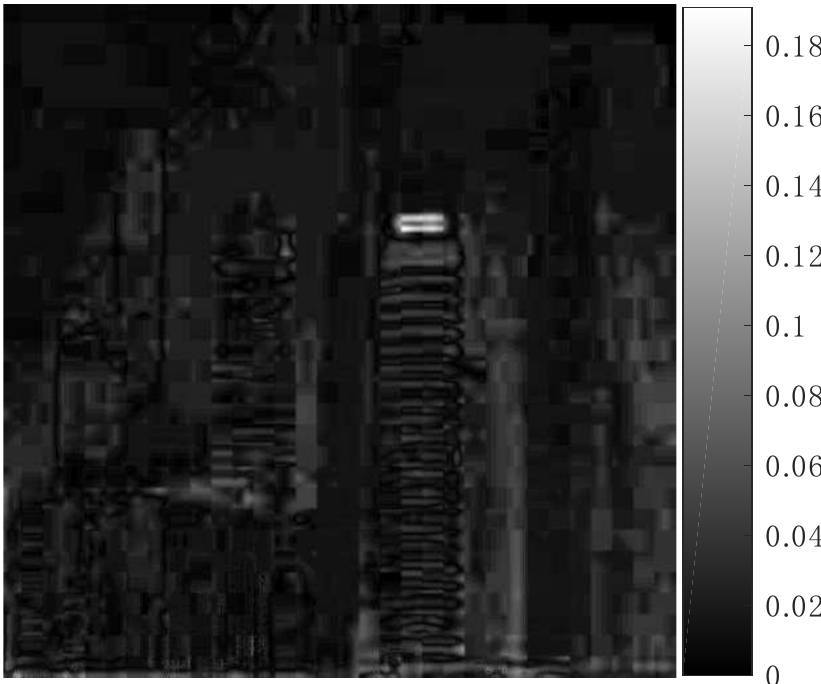} \\
		(d) Luminance of (a) & \hspace{-0.4cm}
		(e) Luminance of (b)& \hspace{-0.4cm}
		(f) Luminance of (c) 
	\end{tabular}
\end{center}
	\vspace{-0.5cm}
	\caption{Defogged results by using the Cycle-Defog2Refog associate with E-net and the corresponding luminance weight map.
	}
	\vspace{-0.5cm}
	\label{Enet-results}
\end{figure}

Although the sky prior strategy can prevent the artifacts, the defogged images generally have a low contrast and can loss some of the texture information by the remaining fog. As shown in Figure \ref{Enet-results} (b), the whole image looks dim and the details of edges are not clear. To overcome this shortcoming, we introduce an enhancer network to improve the quality of the generated images. It is another important new feature in our approach. We refer to this network as E-Net, and its architecture is shown in Figure \ref{enhanceNet-structure}. In E-Net, we also use the encoder-decoder structure. In the encoder network, we used 3 convolutional blocks, each layer is of the same type: 64 filters of a kernel size  3$\times$3 with stride 2. Different from the encoder, the decoder network has 5 deconvolutional blocks, each layer of the first three blocks has the same type: 64 filters of a kernel size 3$\times$3 with stride 1. Note that, in each deconvolutional block, there are several skip connections to be added by element-wise summation from the convolutional blocks, which enforces the network to learn more details. In the fourth deconvolutional block, it consists of a concatenation function and a convolutional layer with 64 filters of a kernel size 1$\times$1. Then, we multiply the output of the first convolutional blocks and the fourth deconvolutional blocks as the input to the last layer. The last deconvolutional layer is with a kernel size of 3$\times$3 and stride of 1. 

In practice, this network is used to enhance the image texture features for the two generators \textit{G} and \textit{R} respectively. Thus, as shown in Figure \ref{refog2defog-structure}, we refer it as Enhancer-Defog Net (E-D-Net,$E_{d}$) in the Defog architecture. In the Refog architecture, we refer it as Enhancer-Refog Net (E-R-Net,$E_{r}$). Figure \ref{Enet-results} shows an example which illustrates the advantage of including the E-Net in the system. We can see that the defogged result with the E-net looks clearer than the one without the E-net. Moreover, this is reflected in the Luminance weight map. This map is used to measure the visibility of each pixel in the image and assigns high values to areas with good visibility and low values to areas with poor visibility. As shown in Figure \ref{Enet-results} (f), the result with E-net has larger dynamic range of luminance values (0$\sim$0.18) than the other in Figure \ref{Enet-results} (e) (0$\sim$0.09). Due to the remaining fog in the result without E-net (Figure \ref{Enet-results}(b)), the texture detials in its luninance weight map are not clear (as shown in Figure \ref{Enet-results} (e)). In constrast, as shown in Figure \ref{Enet-results} (f), the texture detials are clearer.

Once the generated images \textit{G(X)} and \textit{R(Y)} are enhanced by E-D-Net and E-R-Net respectively, the loss function is defined as follows:
\begin{equation}
\begin{split}
\mathcal{L}_{Enhancer} = \Vert E_{d}(X)-R(E_{d}(G(X))) \Vert_{2}^{2} \\ + \Vert E_{r}(Y)-G(E_{r}(R(Y))) \Vert_{2}^{2}
\end{split}
\end{equation}
where $E_{d}$ denotes the E-D-Net, $E_{r}$ denotes the E-R-Net. \textit{X} is the foggy image, \textit{Y} is the fog-free image.

%
\subsection{Discriminator}

The function of the discriminator is to distinguish whether an image is real or fake. In our network, we have two discriminators. As shown in Figure \ref{refog2defog-structure}, \textit{$D_{fog}$} is used to distinguish between the input foggy images and the generated images from \textit{R}; in the same way, $D_{fog-free}$ is used to discriminate between the generate images from \textit{G} and the input clear images. For the discriminator networks, we use 5 convolutional blocks to classify whether the images are real or fake. Each layer has the same kernel size 4$\times$4 with stride of 2, and the filters are 64, 128, 256, 512, 1 from lowest to highest.
\begin{algorithm}[h]
	\caption{The training procedure for our network}
	\begin{algorithmic}[1]
		\REQUIRE ~~\\ 
		The foggy image set, $X_{n}$;\\
		The clear image set, $Y_{n}$;
		\ENSURE ~~\\ 
		The defogged image, $I$; \\
		The refogged image, $II$; 
		
		\LOOP
		\FOR{$x$ $\in$ $X_{n}$,$y$ $\in$ $Y_{n}$} 
		\STATE $I$ = $f_{defog}(x)$ $\longleftarrow$ defog with Defog-Net; \\
		\STATE $I^{'}$ = $f_{refog}(I)$ $\longleftarrow$ refog with Refog-Net;\\
		\STATE $I^{''}$ = $f_{enhancer}(I)$ $\longleftarrow$ enhance with E-D-Net;\\
		\STATE $I^{'''}$ = $f_{refog}(I^{''})$ $\longleftarrow$ refog with Refog-Net;\\
		\STATE loss1 = $mean \parallel x-I^{'} \parallel$ ;\\
		\STATE loss2 = $mean \parallel x-I^{'''} \parallel$;\\
		\STATE loss3 = $mean \parallel 1-D_{fog-free}(y) \parallel$ + $mean \parallel D_{fog-free}(I) \parallel$$\longleftarrow$ GAN fog-free loss1;\\
		\STATE loss4 = $mean \parallel 1-D_{fog-free}(y) \parallel$ + $mean \parallel D_{fog-free}(I^{''}) \parallel$ $\longleftarrow$ GAN fog-free loss2;\\
		\STATE loss5 = mean$(VGG(x,y,f_{defog},f_{refog}))$ $\longleftarrow$ VGG loss ;\\
		\STATE $\mathcal{L}_{defog}$ = loss1+loss2+loss3+loss4+loss5
		\RETURN $I$
		\ENDFOR
		
		\FOR{$y$ $\in$ $Y_{n}$,$x$ $\in$ $X_{n}$}
		\STATE $II$ = $f_{refog}(y)$ $\longleftarrow$ refog with Defog-Net; \\
		\STATE $II^{'}$ = $f_{defog}(II)$ $\longleftarrow$ defog with Refog-Net;\\
		\STATE $II^{''}$ = $f_{enhancer}(II)$ $\longleftarrow$ enhance with E-R-Net;\\
		\STATE $II^{'''}$ = $f_{defog}(II^{''})$ $\longleftarrow$ Defog with Refog-Net;\\
		\STATE loss1 = $mean \parallel y-II^{'} \parallel$; \\
		\STATE loss2 = $mean \parallel y-II^{'''} \parallel$ ;\\
		\STATE loss3 = $mean \parallel 1-D_{fog}(x) \parallel$ +$ mean \parallel D_{fog}(II) \parallel$$\longleftarrow$ GAN fog loss1;\\
		\STATE loss4 = $mean\parallel 1-D_{fog}(x)\parallel$ + $mean \parallel D_{fog}(II^{''}) \parallel$ $\longleftarrow$ GAN fog loss2;\\
		\STATE loss5 = $mean(VGG(y,x,f_{defog},f_{refog}))$ $\longleftarrow$ VGG loss ;\\
		\STATE $\mathcal{L}_{refog}$ = loss1+loss2+loss3+loss4+loss5
		\RETURN $II$
		\ENDFOR
		\ENDLOOP
	\end{algorithmic}
	\label{algorithm}
\end{algorithm}
\subsection{Full objective of Cycle-Defog2Refog}

In this work, we use the hybrid loss function. We first use the generative adversarial loss \cite{GAN2014NIPS} to supervise the network. For the refog mapping function \textit{R} and its discriminator $D_{fog}$, the adversarial loss of refog is defined as follows:
\begin{equation}
	\begin{split}
	\mathcal{L}_{r-adv}(R,D_{fog},Y,X)= \mathbb{E}_{X \sim p_{fog}(X)}\left[\log D_{fog}(X)\right] \\
	+\mathbb{E}_{Y \sim p_{clear}(Y)}\left[\log(1 - D_{fog}(R(Y)))\right]
	\end{split}
\end{equation}
where, \textit{R} is our refog network in Figure \ref{refog2defog-structure}, while \textit{$D_{fog}$} aims to distinguish between the generated foggy images by \textit{R(Y)} and the real foggy images \textit{X}. In a similar way, the defog adversarial loss is defined as follows:
\begin{equation}
\begin{split}
\mathcal{L}_{d-adv}(G,D_{fog-free},X,Y)=\\ \mathbb{E}_{Y \sim p_{clear}(Y)}\left[\log D_{fog-free}(Y)\right]+\\
\mathbb{E}_{X \sim p_{fog}(X)}\left[\log(1 - D_{fog-free}(G(X)))\right]
\end{split}
\end{equation}

In the above equation, \textit{G} and \textit{R} are able to minimize the objective against the adversary $D_{fog}$ and $D_{fog-free}$ that try to maximize it. However, there will be some artifacts in the generated results when we only use the adversarial loss. We also found that it is difficult to remove the fog from foggy images by only using this loss function. The details are discussed later in Section 5.

In order to better remove fog and preserve more detail of texture information, we introduce another loss function, which is defined as Cycle-Refog loss. This function is used to minimize the objective between the foggy images \textit{X} and its reconstructed foggy images \textit{R(G(X))}, the clear images \textit{Y} and its reconstructed clear images \textit{G(R(Y))} (as shown in Figure \ref{refog2defog-structure}). We formulate this objective as:
\begin{equation}
\mathcal{L}_{Cycle-Refog} = \Vert X-R(G(X)) \Vert_{2}^{2} + \Vert Y-G(R(Y)) \Vert_{2}^{2}
\end{equation}

Moreover, in order to learn more textural information from foggy images, a perceptual loss based on pre-trained $VGG16$ \cite{VGG16} is introduced to further constrain the generators, which is defined as:
\begin{equation}
\begin{split}
\mathcal{L}_{VGG} = \Vert \mathcal{P}_{i}(X)-\mathcal{P}_{i}(R(G(X))) \Vert_{2}^{2} \\ + \Vert \mathcal{P}_{i}(Y)-\mathcal{P}_{i}(G(R(Y))) \Vert_{2}^{2}
\end{split}
\end{equation}
where $\mathcal{P}_{i}$ denotes the feature maps of the i-th layer of the \textit{VGG16} network. 

Finally, by combining the refog adversarial loss, defog adversarial loss, enhancer loss and Cycle-Refog loss, our overall loss function is:
\begin{equation}
	\begin{split}
		\mathcal{L}_{Cycle-Defog2Refog}= &\gamma_{1}\mathcal{L}_{r-adv}+\gamma_{2}\mathcal{L}_{d-adv}\\
		&+\gamma_{3}\mathcal{L}_{Cycle-Refog}+\gamma_{4}\mathcal{L}_{Enhancer}\\
		&+\gamma_{5}\mathcal{L}_{VGG}
	\end{split}
\end{equation}
where $\gamma_{1}$, $\gamma_{2}$, $\gamma_{3}$, $\gamma_{4}$ and $\gamma_{5}$ are the positive weights. To optimize our network, we aim to solve:
\begin{equation}
D^{*},R^{*} = arg\min\limits_{G,R,E} \max\limits_{D_{fog},D_{fog-free}}\mathcal{L}_{Cycle-Defog2Refog}
\end{equation}
We use the generator $D^{*}$ to remove the fog from the testing image. The overview of our traning procedure for Cycle-Defog2Refog is shown in Algorithm \ref{algorithm}.

\begin{center}
	\begin{table*}[htbp]
		\caption{Quantitative comparison with other methods on synthetic images.}
		\vspace{-0.3cm}
		\setlength{\tabcolsep}{0.8mm}{
			\begin{tabular}{ccccc|cccc|cccc|cccc|cccc}
				\toprule 
				\multirow{2}{*}{} & \multicolumn{4}{c}{S1} &\multicolumn{4}{c}{S2} &\multicolumn{4}{c}{S3}
				&\multicolumn{4}{c}{S4} &\multicolumn{4}{c}{S5}\\
				\cline{2-21}
				& \textit{e} & $\overline{\textit{r}}$ & $\delta(\%)$& $\mathcal{F}$ & 
				\textit{e} & $\overline{\textit{r}}$ & $\delta(\%)$& $\mathcal{F}$ &
				\textit{e} & $\overline{\textit{r}}$ & $\delta(\%)$& $\mathcal{F}$ & 
				\textit{e} & $\overline{\textit{r}}$ & $\delta(\%)$& $\mathcal{F}$ & 
				\textit{e} & $\overline{\textit{r}}$ & $\delta(\%)$& $\mathcal{F}$ \\
				\midrule
				He~\cite{he2011PAMI} &  0.69  &  1.81  &  0.07  &  0.70 &  1.36  &  2.23  &  0.03  &  0.97  &  1.49  &  2.09  &  0.01 &  1.65  &  0.94 & 1.66 & 0.12 & 0.64  & 1.01 & 1.58 &	0.09 &	0.67\\
				\midrule
				Meng~\cite{meng2013ICCV} & 0.26  &  2.52  &  0.43  &  0.69  &  1.40  &  3.06  &  0.50  &  1.01  &  4.13  &  4.95  &  0.34  &  1.12  & 0.71 & 2.65 & 0.66 &	0.68  &	1.04 &	2.28 &	0.19 &	0.62  \\
				\midrule
				Zhu~\cite{zhu2015TIP} & 0.25  &  1.66  &  0.01  &  0.87 &  0.66  &   1.57  &  6.29 &  1.12 &  0.93  &   1.73  &   0.02  &  2.47  & 0.62 & 1.53 &	0.02 & 0.85  & 	0.82 &	1.54 &	0.05 &	0.84 \\
				\midrule
				Cai~\cite{cai2016dehazenet} & 0.21	& 1.77 &	0.01 &	1.26 & 0.68 &	1.66 & 3.74 & 1.35 &  0.82 &	1.79 &	0.01 &	2.85 & 0.64&	1.68 &	0.01&	0.99& 0.87&	1.54&	0.19 &	0.87\\
				\midrule
				Ren~\cite{MSCNN2016ECCV} & 0.22 & 1.90 &	0.02 &	0.96  &	0.45 &	1.67 &	0.01 &	1.90 &	0.93 &	1.96 &	0.09 &	2.39 &	0.43 &	1.95 &	0.31&	0.97 &	0.68	&1.59&	0.02&	1.02\\
				\midrule
				Ours  &	0.79 &	2.14 &	0.03 &	0.76 &	1.68 &	2.69 &	0.01 &	1.03  &	1.89 &	2.34 &	0.03 &	1.73 &	1.65 &	2.05 &	0.07 &	0.71 &	1.01&	1.77&	1.81 &	0.44 \\
				\bottomrule
		\end{tabular}}
		\label{synthetic-quantitative}    
	\end{table*}
	\vspace{-0.5cm}
\end{center}
\begin{figure*}[htbp]\scriptsize
	\begin{center}
		 \resizebox{\textwidth}{11cm}{
		\begin{tabular}{@{}ccccccccc@{}}
			& \hspace{-0.4cm}
			\includegraphics[width = 0.13\textwidth]{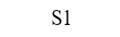} & \hspace{-0.4cm}
			\includegraphics[width = 0.13\textwidth]{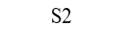} & \hspace{-0.4cm}
			\includegraphics[width = 0.13\textwidth]{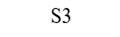} & \hspace{-0.4cm}
			\includegraphics[width = 0.13\textwidth]{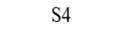} & \hspace{-0.4cm}
			\includegraphics[width = 0.13\textwidth]{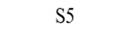} \\
			\rotatebox[origin=lt]{90}{\Large{(a) Inputs}} & \hspace{-0.4cm}
			\includegraphics[width = 0.2\textwidth]{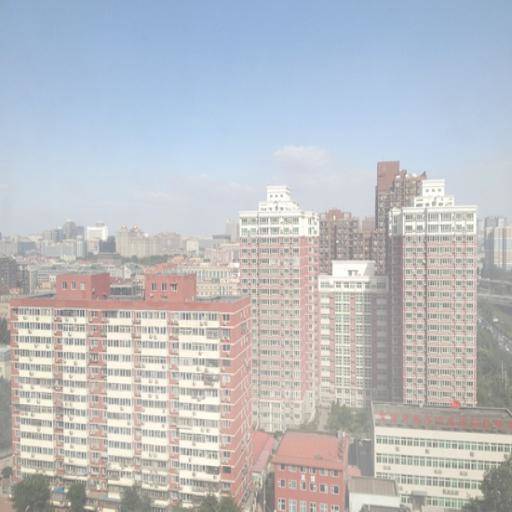} & \hspace{-0.4cm}
			\includegraphics[width = 0.2\textwidth]{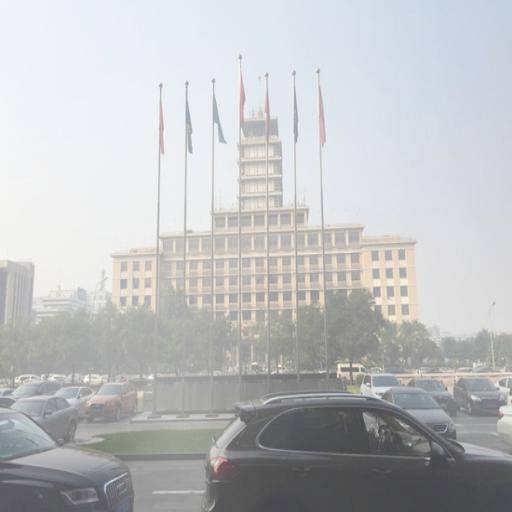} & \hspace{-0.4cm}
			\includegraphics[width = 0.2\textwidth]{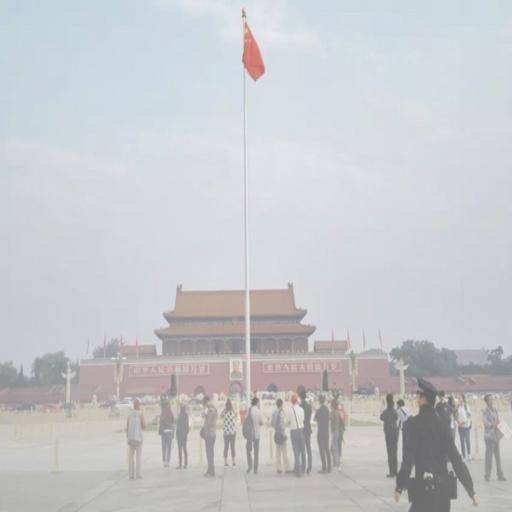} & \hspace{-0.4cm}
			\includegraphics[width = 0.2\textwidth]{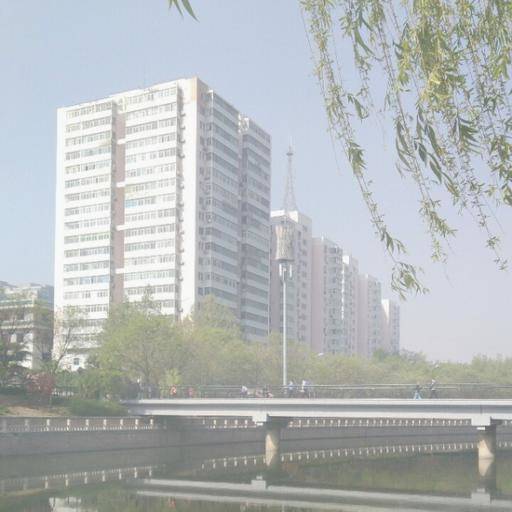} & \hspace{-0.4cm}
			\includegraphics[width = 0.2\textwidth]{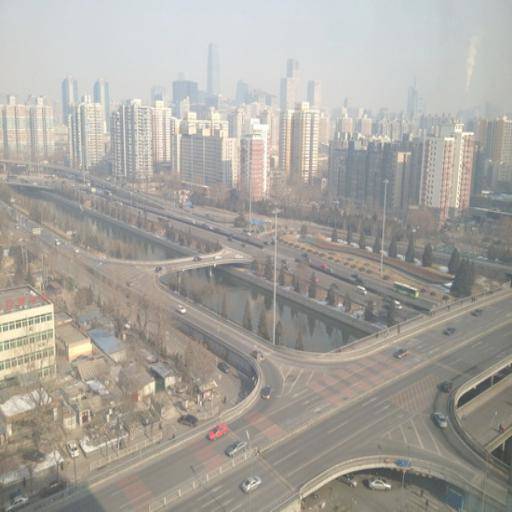} \\
			\rotatebox[origin=lt]{90}{\Large{(b) He \cite{he2011PAMI}}} & \hspace{-0.4cm}
			\includegraphics[width = 0.2\textwidth]{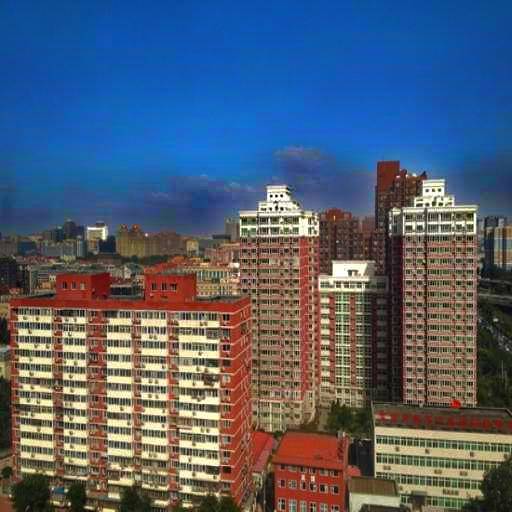} & \hspace{-0.4cm}
			\includegraphics[width = 0.2\textwidth]{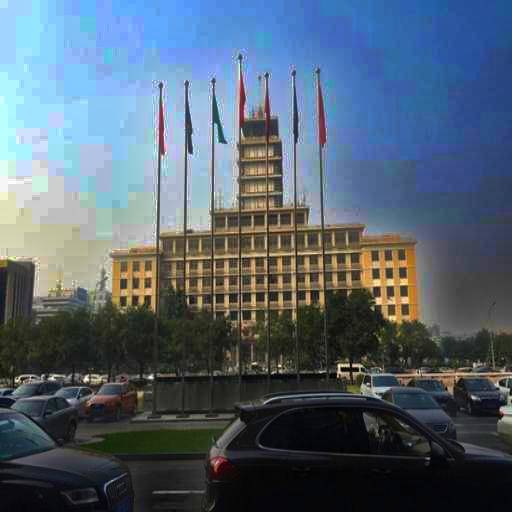} & \hspace{-0.4cm}
			\includegraphics[width = 0.2\textwidth]{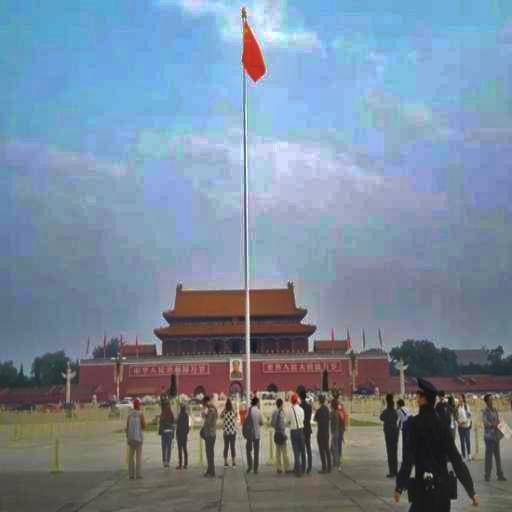} & \hspace{-0.4cm}
			\includegraphics[width = 0.2\textwidth]{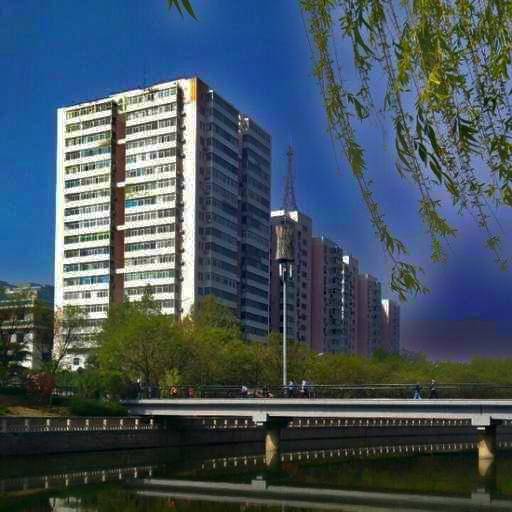} & \hspace{-0.4cm}
			\includegraphics[width = 0.2\textwidth]{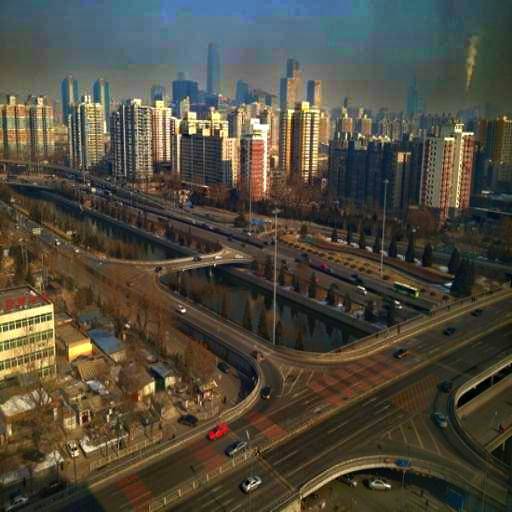} \\
			\rotatebox[origin=lt]{90}{\Large{(c) Meng \cite{meng2013ICCV}}} & \hspace{-0.4cm}
			\includegraphics[width = 0.2\textwidth]{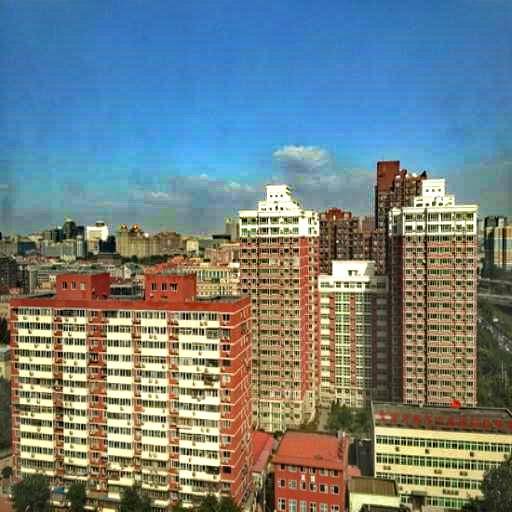} & \hspace{-0.4cm}
			\includegraphics[width = 0.2\textwidth]{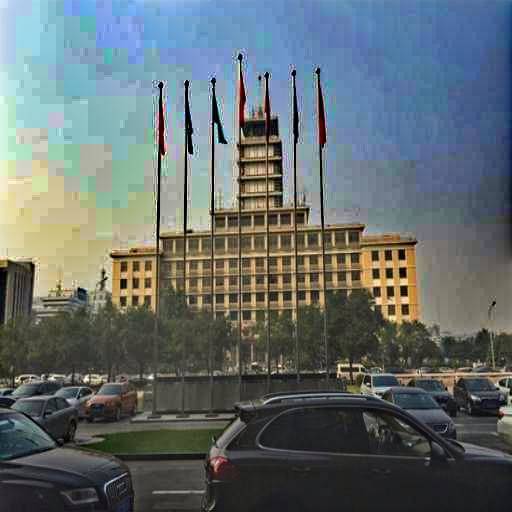} & \hspace{-0.4cm}
			\includegraphics[width = 0.2\textwidth]{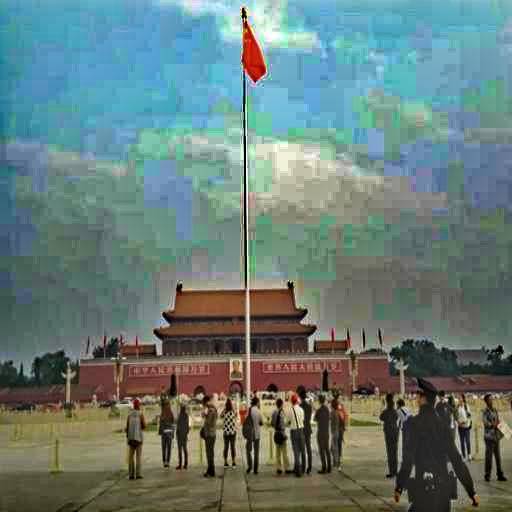} & \hspace{-0.4cm}
			\includegraphics[width = 0.2\textwidth]{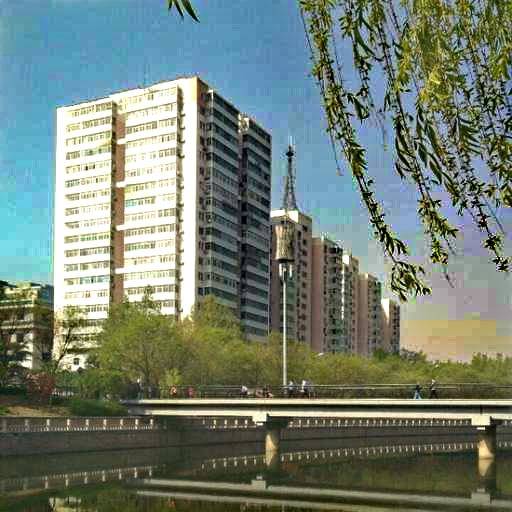} & \hspace{-0.4cm}
			\includegraphics[width = 0.2\textwidth]{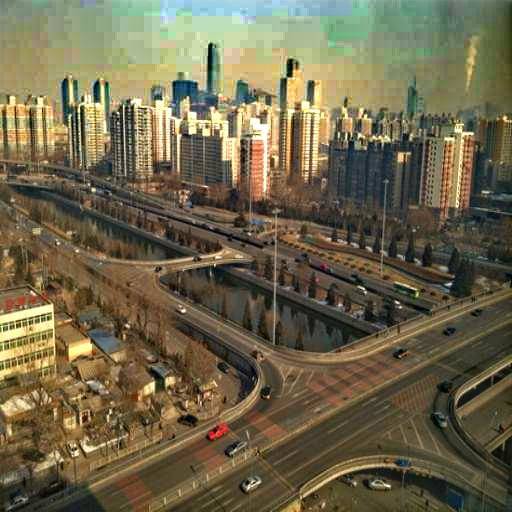} \\
			\rotatebox[origin=lt]{90}{\Large{(d) Zhu \cite{zhu2015TIP}}} & \hspace{-0.4cm}
			\includegraphics[width = 0.2\textwidth]{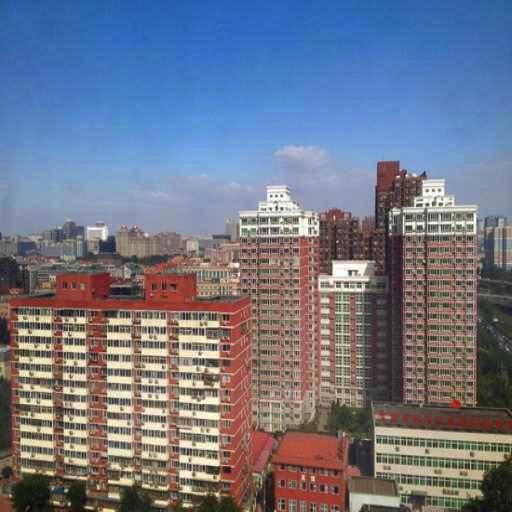} & \hspace{-0.4cm}
			\includegraphics[width = 0.2\textwidth]{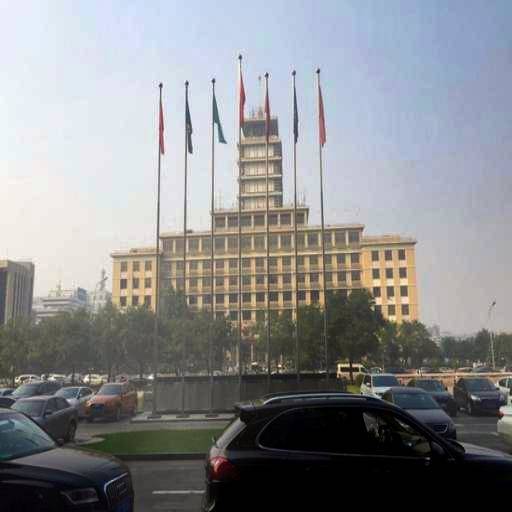} & \hspace{-0.4cm}
			\includegraphics[width = 0.2\textwidth]{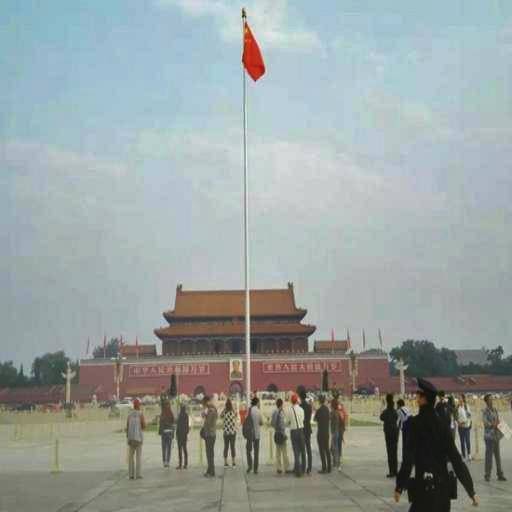} & \hspace{-0.4cm}
			\includegraphics[width = 0.2\textwidth]{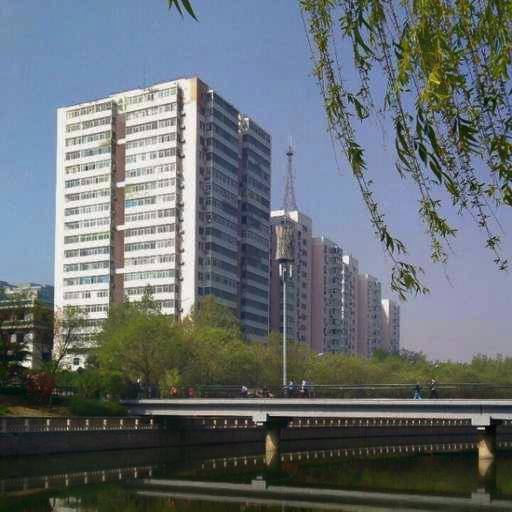} & \hspace{-0.4cm}
			\includegraphics[width = 0.2\textwidth]{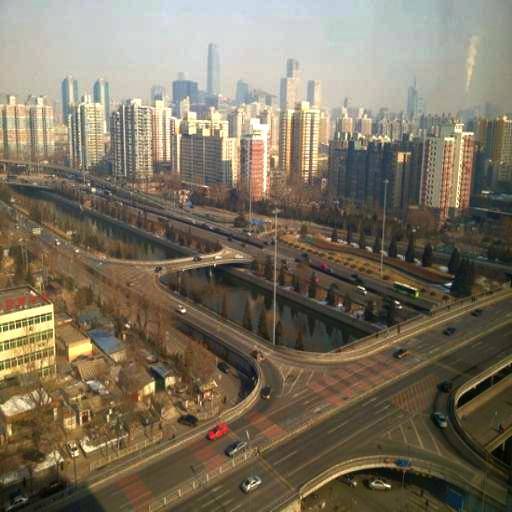} \\
			\rotatebox[origin=lt]{90}{\Large{(e) Cai \cite{cai2016dehazenet}}} & \hspace{-0.4cm}
			\includegraphics[width = 0.2\textwidth]{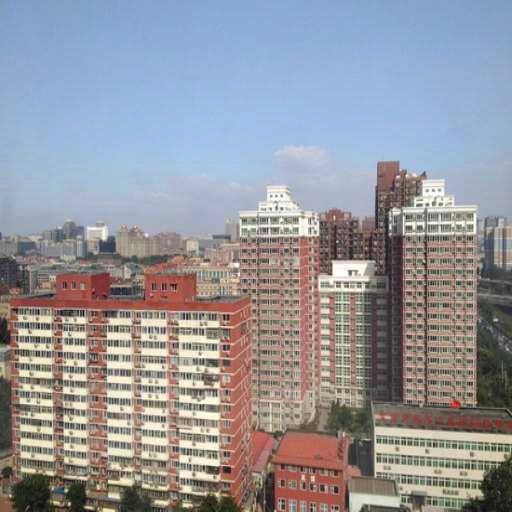} & \hspace{-0.4cm}
			\includegraphics[width = 0.2\textwidth]{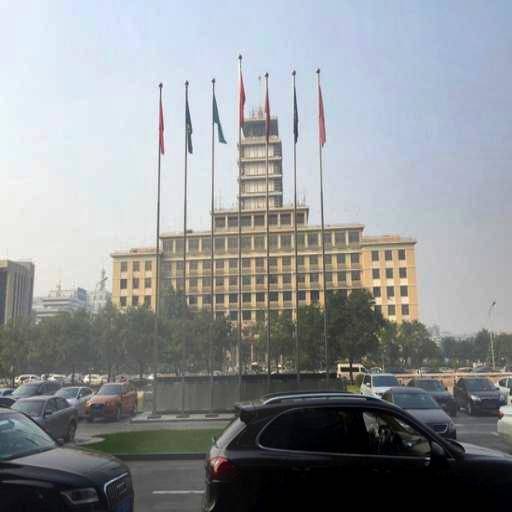} & \hspace{-0.4cm}
			\includegraphics[width = 0.2\textwidth]{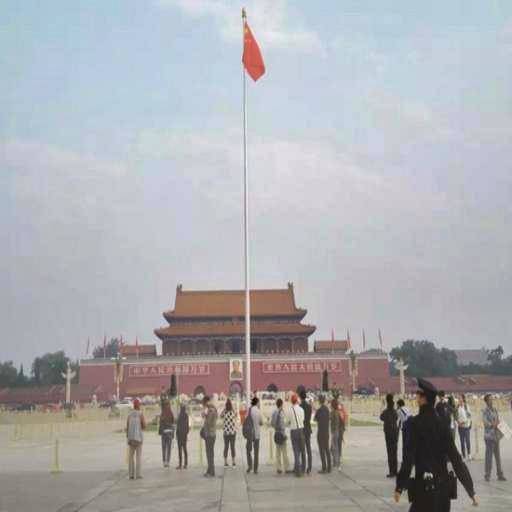} & \hspace{-0.4cm}
			\includegraphics[width = 0.2\textwidth]{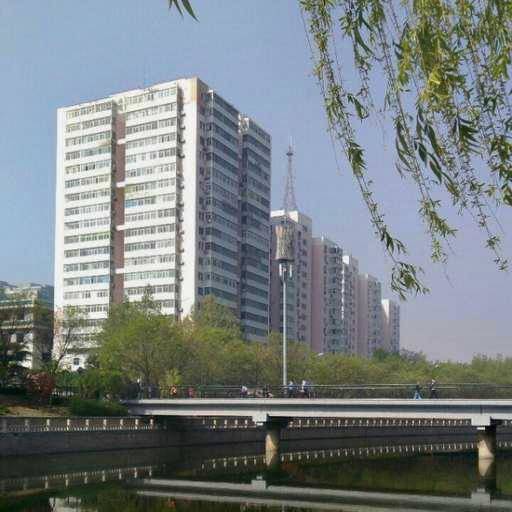} & \hspace{-0.4cm}
			\includegraphics[width = 0.2\textwidth]{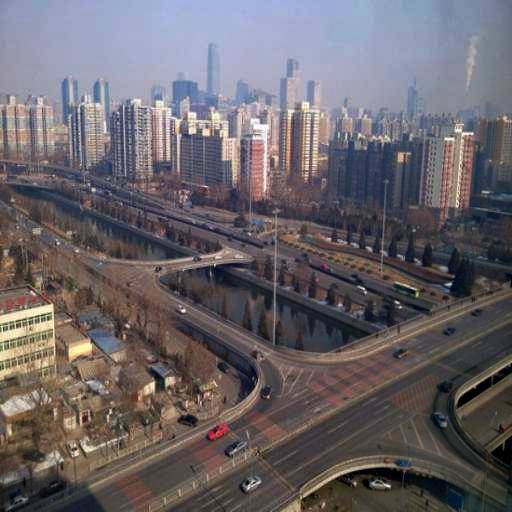} \\
			\rotatebox[origin=lt]{90}{\Large{(f) Ren \cite{MSCNN2016ECCV}}} & \hspace{-0.4cm}
			\includegraphics[width = 0.2\textwidth]{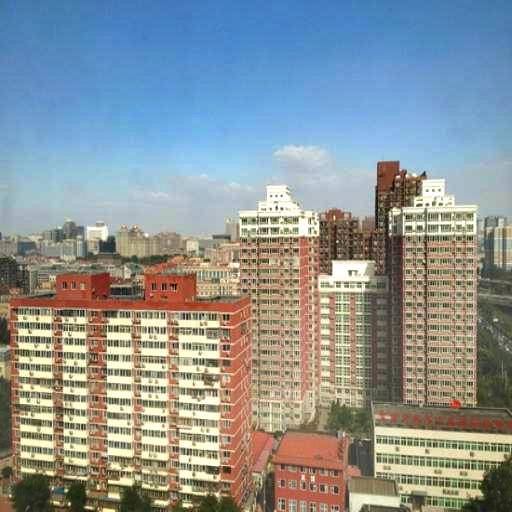} & \hspace{-0.4cm}
			\includegraphics[width = 0.2\textwidth]{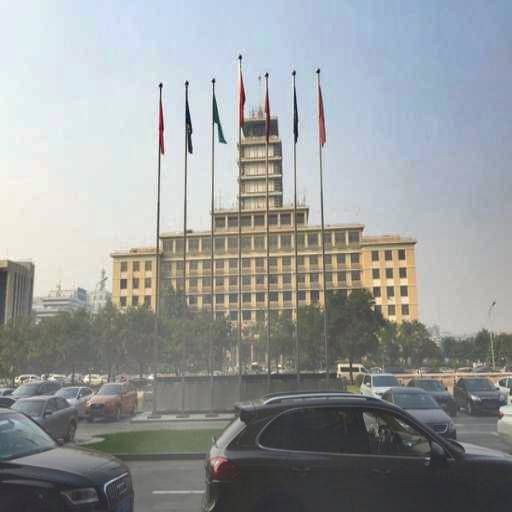} & \hspace{-0.4cm}
			\includegraphics[width = 0.2\textwidth]{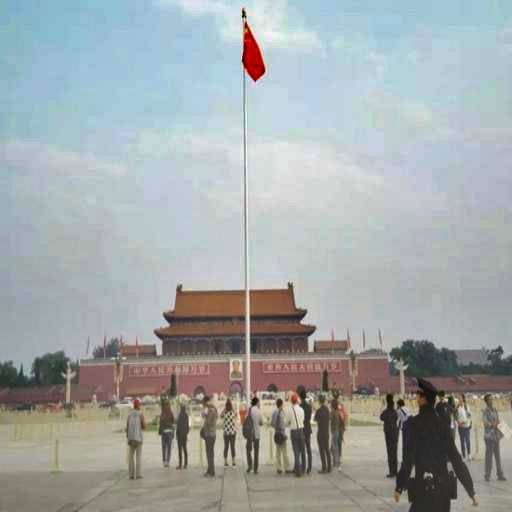} & \hspace{-0.4cm}
			\includegraphics[width = 0.2\textwidth]{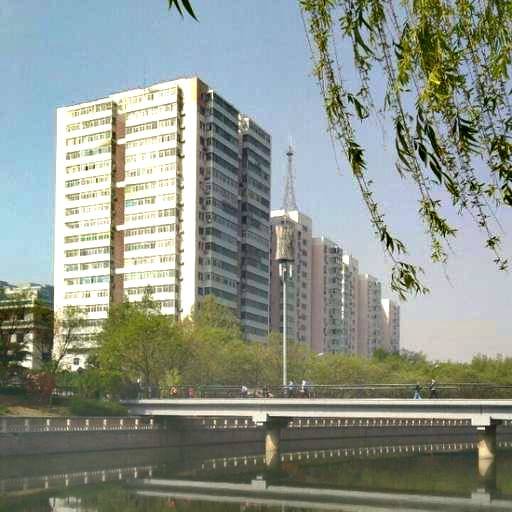} & \hspace{-0.4cm}
			\includegraphics[width = 0.2\textwidth]{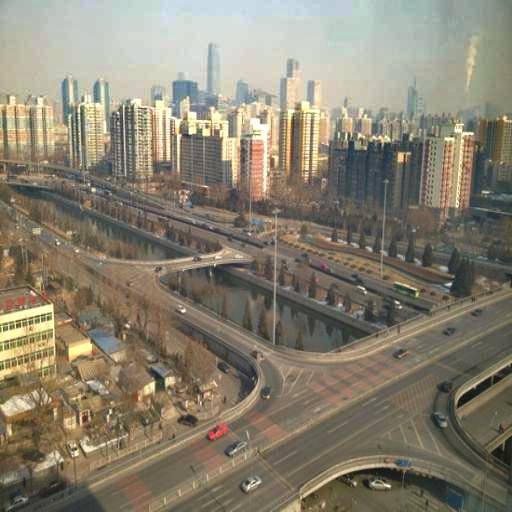} \\
			\rotatebox[origin=lt]{90}{\Large{(g) Ours}} & \hspace{-0.4cm}
			\includegraphics[width = 0.2\textwidth]{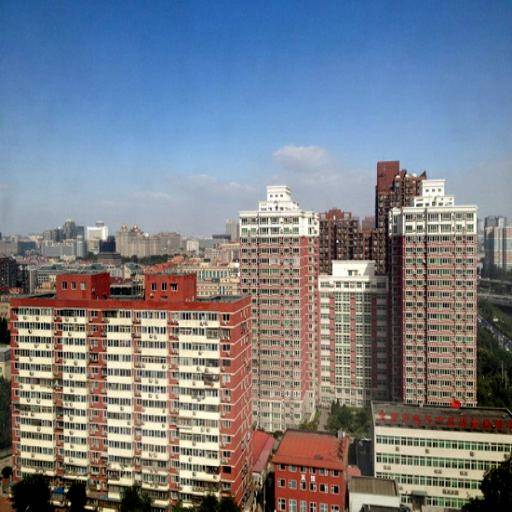} & \hspace{-0.4cm}
			\includegraphics[width = 0.2\textwidth]{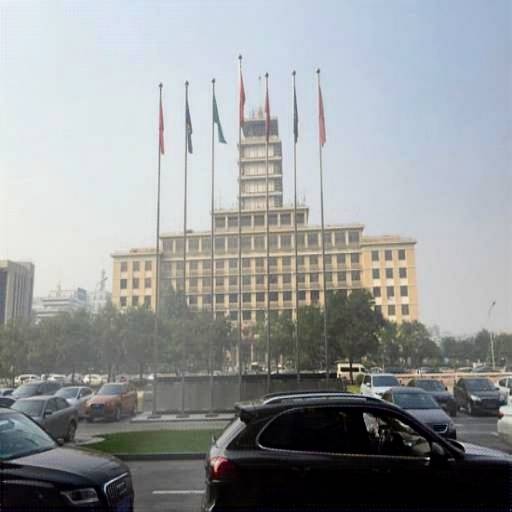} & \hspace{-0.4cm}
			\includegraphics[width = 0.2\textwidth]{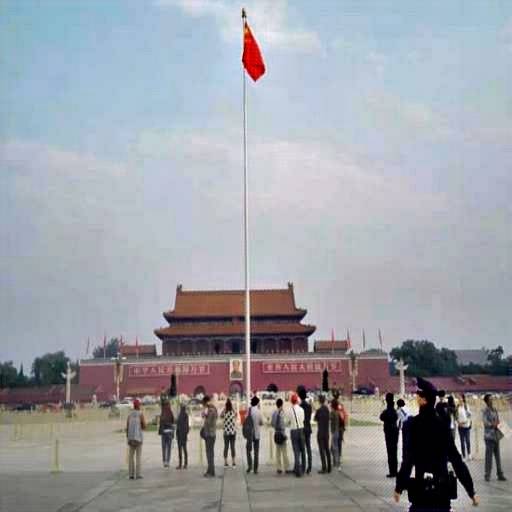} & \hspace{-0.4cm}
			\includegraphics[width = 0.2\textwidth]{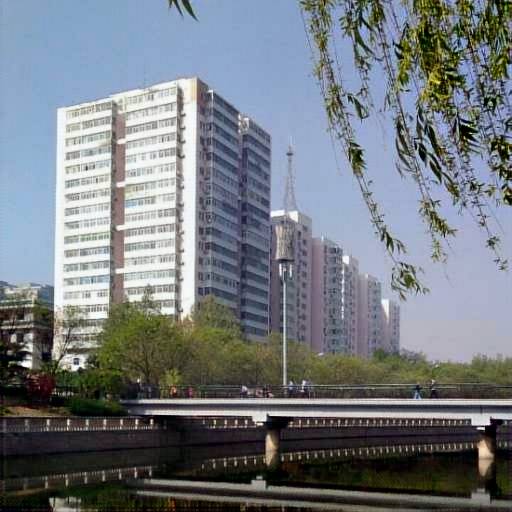} & \hspace{-0.4cm}
			\includegraphics[width = 0.2\textwidth]{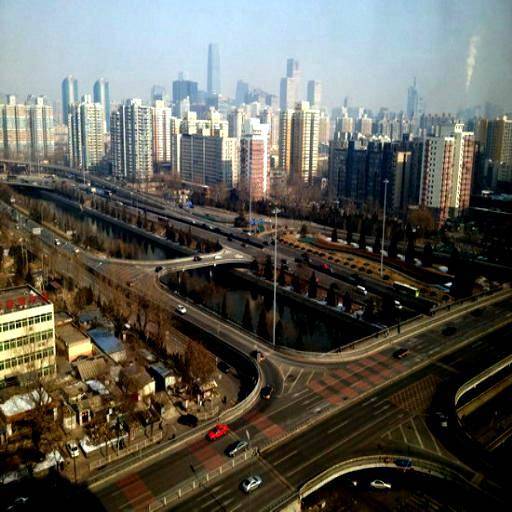} \\
			
		\end{tabular}}
	\end{center}
	\vspace{-0.3cm}
	\caption{Qualitative evaluate the defogged results on the synthetic images.
	}
	\vspace{-2.5mm}
	\label{synthetic-results}
\end{figure*}
\begin{figure}[t]\footnotesize
	\begin{center}
		\begin{tabular}{@{}c@{}}
			\includegraphics[width = 0.45\textwidth]{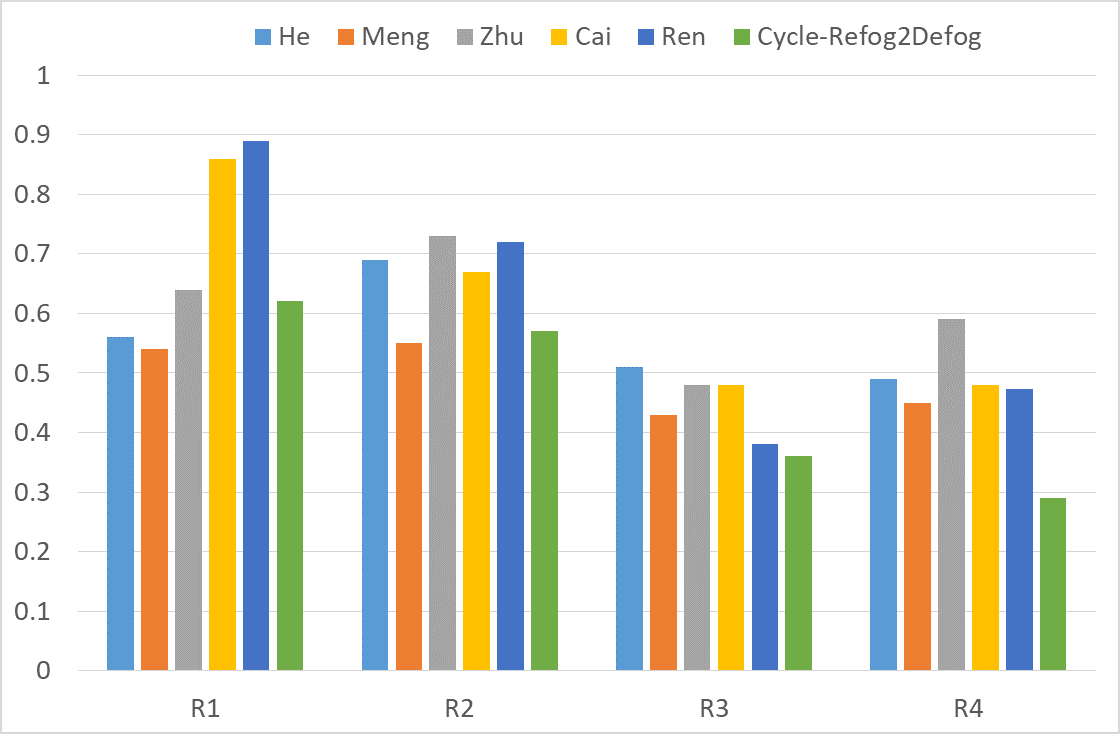} 
		\end{tabular}
	\end{center}
	\vspace{-0.5cm}
	\caption{Quantitative evaluate $\mathcal{F}$ of the defog results in Figure \ref{real-results}.
	}
	\vspace{-0.3cm}
	\label{D-real-quantitative}
\end{figure}
\begin{figure}[t]\footnotesize
	\begin{center}
		\begin{tabular}{@{}cc@{}}
			\includegraphics[width = 0.22\textwidth]{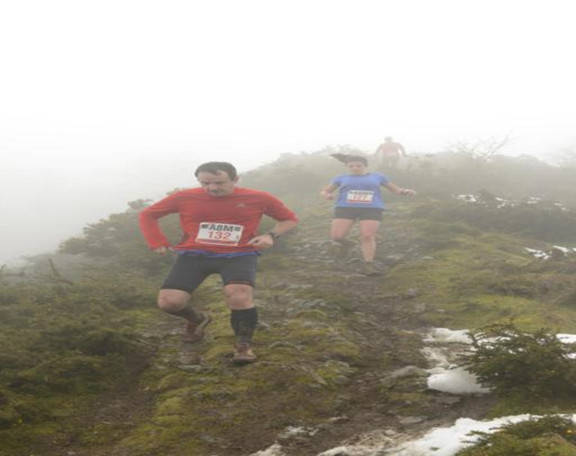} & \hspace{-0.4cm}
			\includegraphics[width = 0.22\textwidth]{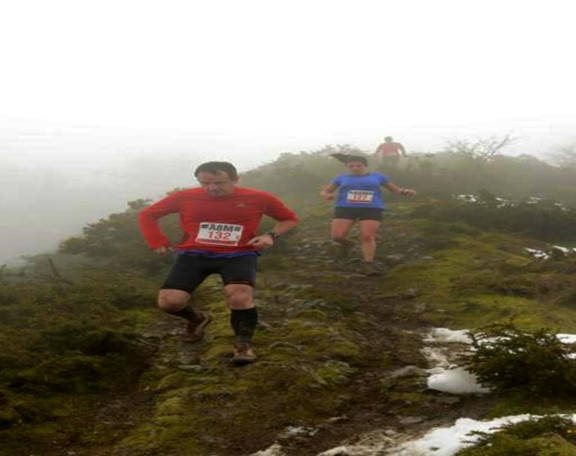} \\
			Foggy image & \hspace{-0.4cm}
			Cai \cite{cai2016dehazenet}\\
			\includegraphics[width = 0.22\textwidth]{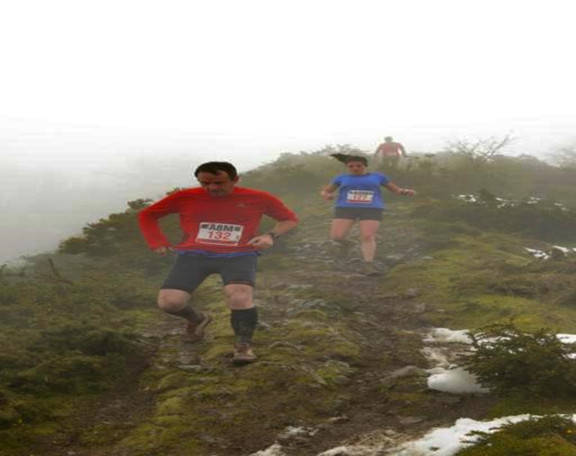} & \hspace{-0.4cm}
			\includegraphics[width = 0.22\textwidth]{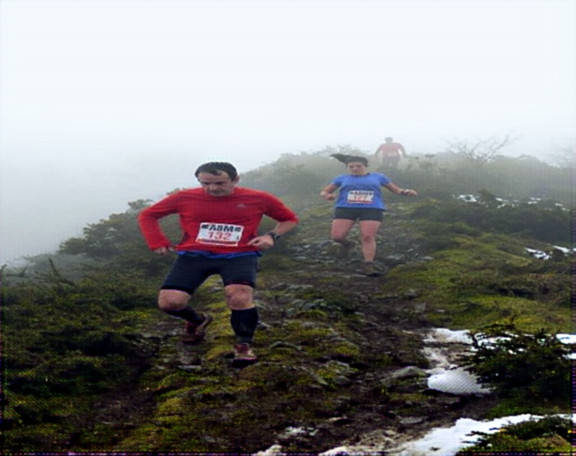} \\
			Ren \cite{MSCNN2016ECCV} & \hspace{-0.4cm}
			Our result\\
			\includegraphics[width = 0.22\textwidth]{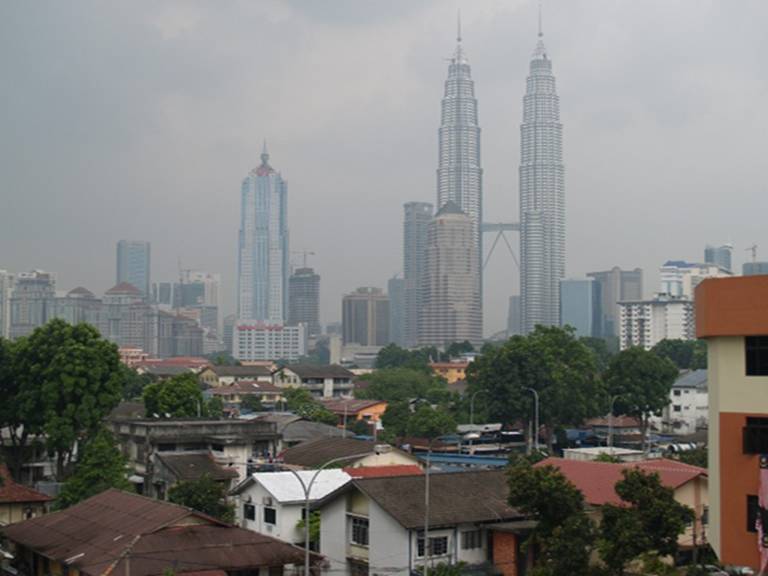} & \hspace{-0.4cm}
			\includegraphics[width = 0.22\textwidth]{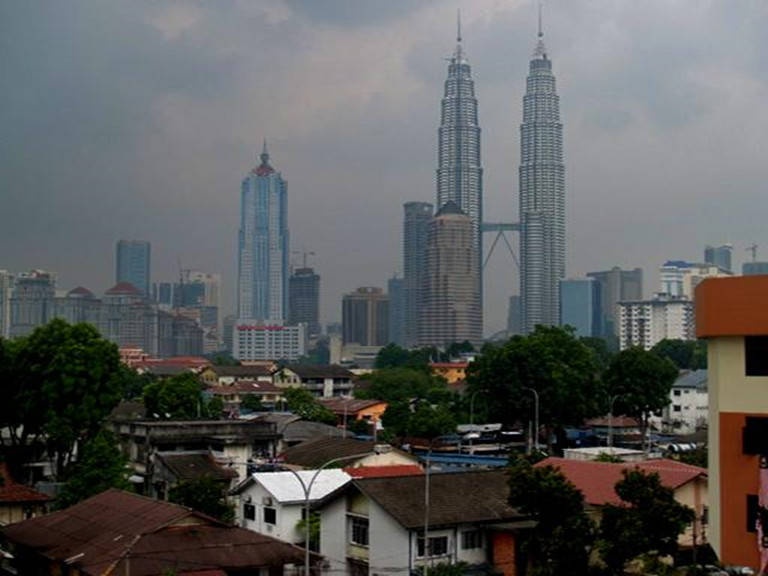} \\
			Foggy image & \hspace{-0.4cm}
			Cai \cite{cai2016dehazenet}\\
			\includegraphics[width = 0.22\textwidth]{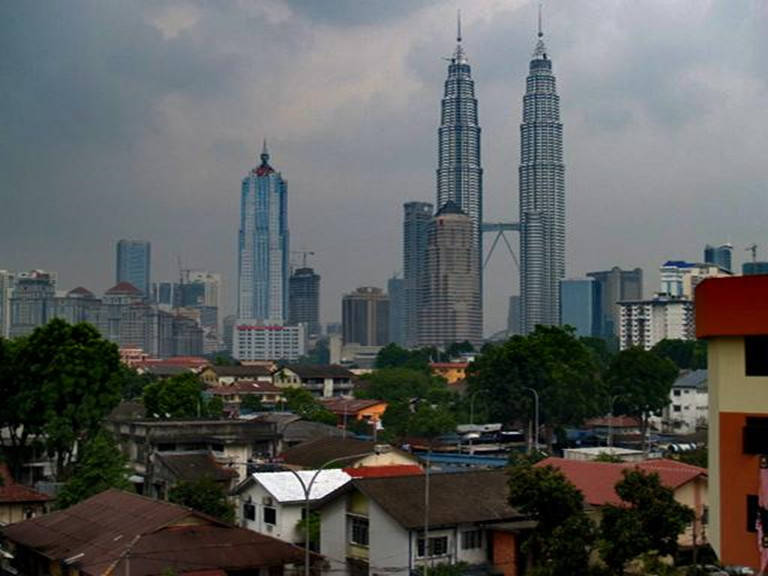} & \hspace{-0.4cm}
			\includegraphics[width = 0.22\textwidth]{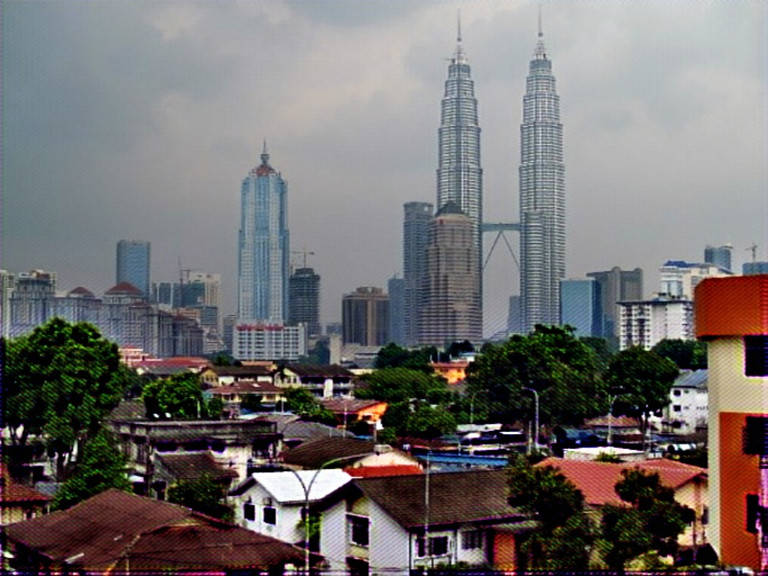} \\
			Ren \cite{MSCNN2016ECCV} & \hspace{-0.4cm}
			Our result\\
		\end{tabular}
	\end{center}
	\vspace{-0.5cm}
	\caption{Comparison with the learning-based defogging techniques of Cai\cite{cai2016dehazenet} and Ren \cite{MSCNN2016ECCV} on real-world images.
	}
	\vspace{-0.35cm}
	\label{learning-comparison}
\end{figure}
\begin{center}
	\begin{table*}[htbp]
		\caption{Quantitative comparison with other methods on real world images.}
		\vspace{-0.3cm}
		\setlength{\tabcolsep}{3.2mm}{
			\begin{tabular}{cccc|ccc|ccc|ccc}
				\toprule 
				\multirow{2}{*}{} & \multicolumn{3}{c}{R1} &\multicolumn{3}{c}{R2} &\multicolumn{3}{c}{R3}
				&\multicolumn{3}{c}{R4}\\
				\cline{2-13}
				& \textit{e} & $\overline{\textit{r}}$ & $\delta(\%)$ & 
				\textit{e} & $\overline{\textit{r}}$ & $\delta(\%)$ &
				\textit{e} & $\overline{\textit{r}}$ & $\delta(\%)$&
				\textit{e} & $\overline{\textit{r}}$ & $\delta(\%)$ \\
				\midrule
				He~\cite{he2011PAMI} &  1.09 &	1.22 &	0.21 & 0.09 & 1.09 & 0.07 & 0.29 &	1.13 &	1.35 & 0.13 &	0.99 &	1.01  \\
				\midrule
				Meng~\cite{meng2013ICCV} &	0.72&	1.94&	0.79&	0.39 &	1.65 &	0.14 &	0.40 &	1.62 &	3.83 &  0.14 &	1.24 &	1.23  \\
				\midrule
				Zhu~\cite{zhu2015TIP} & 0.94 &	1.34&	0.81 &	0.07 &	1.14 &	0.06 &	0.24 &	0.98 &	0.76 & 0.06 &	0.61 &	0.85  \\
				\midrule
				Cai~\cite{cai2016dehazenet} & 0.41 & 1.46 &	0.58 &	0.01 &	1.20 &	5.78 &	0.19 &	1.10 &	9.49 &	0.13 &	0.89 &	1.62  \\
				\midrule
				Ren~\cite{MSCNN2016ECCV} &	0.41 &	1.48 &	0.45 &	0.06 &	1.18 &	1.41 &	0.13 &	1.34 &	2.74 &	0.16 &	0.90 &	1.95  \\
				\midrule
				Ours & 0.67 &	1.54 &	0.48 &	0.47 &	1.89 &	0.06 & 0.36 & 1.63 & 1.52 &	0.38 &	1.49 &	1.04   \\
				\bottomrule
		\end{tabular}}
		\label{real-quantitative}    
	\end{table*}
	\vspace{-0.8cm}
\end{center}
\begin{figure*}[htbp]\scriptsize
	\begin{center}
		\resizebox{\textwidth}{8cm}{
			\begin{tabular}{@{}ccccc@{}}
				& \hspace{-0.4cm}
				\includegraphics[width = 0.08\textwidth]{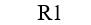} & \hspace{-0.4cm}
				\includegraphics[width = 0.08\textwidth]{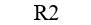} & \hspace{-0.4cm}
				\includegraphics[width = 0.08\textwidth]{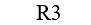} & \hspace{-0.4cm}
				\includegraphics[width = 0.08\textwidth]{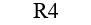} \\
				\rotatebox[origin=lt]{90}{\large {(a) Inputs}} & \hspace{-0.4cm}
				\includegraphics[width = 0.2\textwidth]{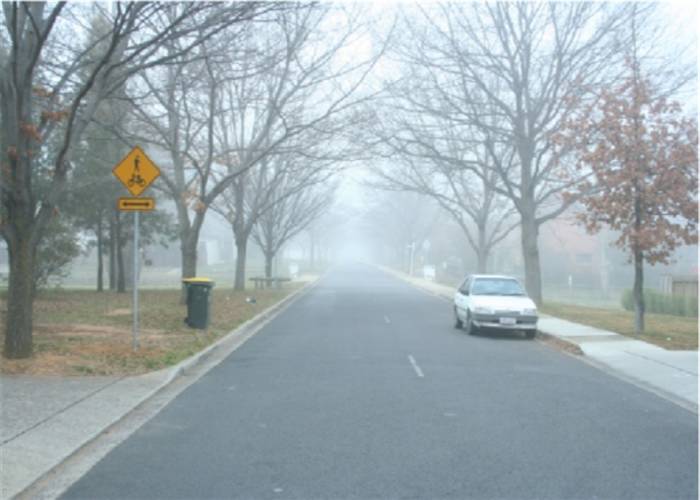} & \hspace{-0.4cm}
				\includegraphics[width = 0.2\textwidth]{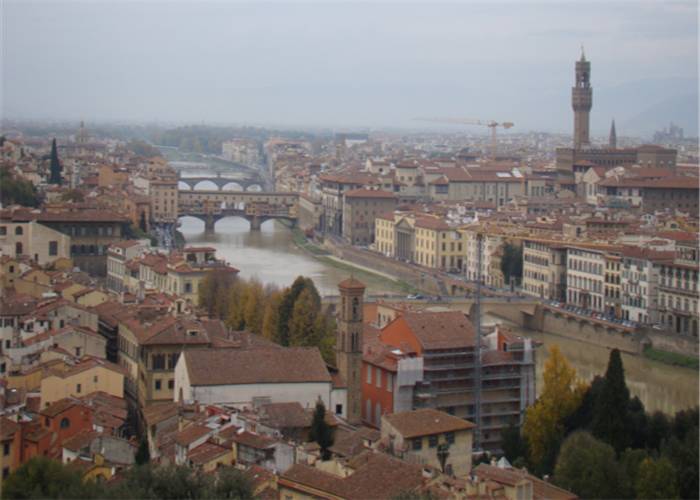} & \hspace{-0.4cm}
				\includegraphics[width = 0.2\textwidth]{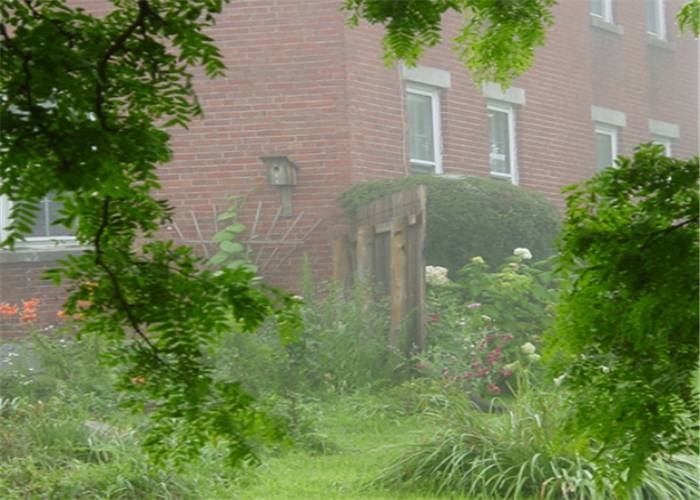} & \hspace{-0.4cm}
				\includegraphics[width = 0.2\textwidth]{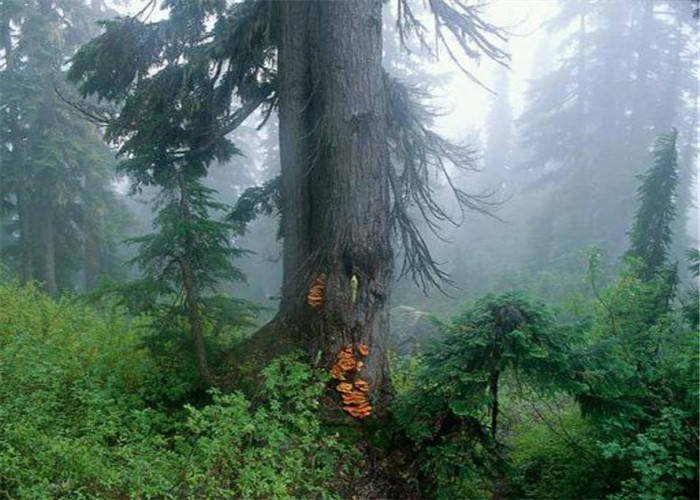} \\
				\rotatebox[origin=lt]{90}{\large{(b) He \cite{he2011PAMI}}} & \hspace{-0.4cm}
				\includegraphics[width = 0.2\textwidth]{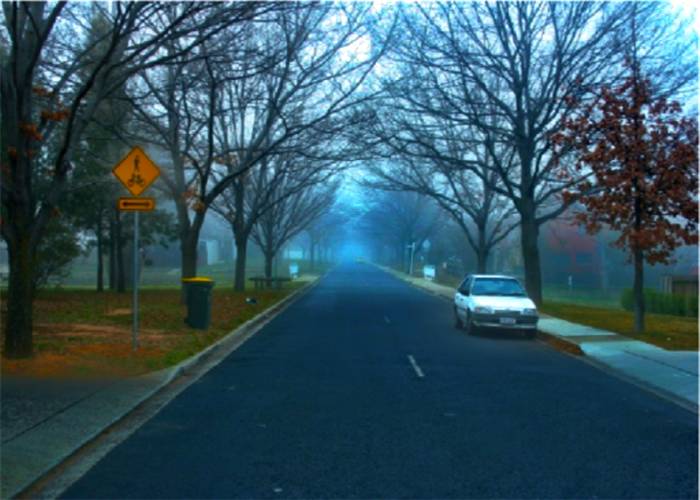} & \hspace{-0.4cm}
				\includegraphics[width = 0.2\textwidth]{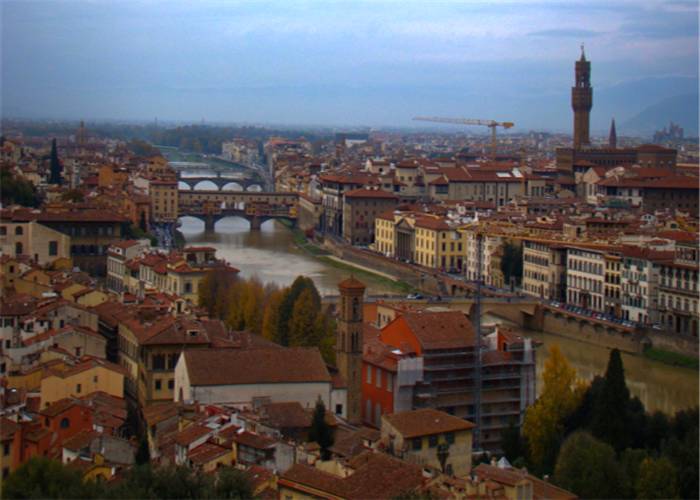} & \hspace{-0.4cm}
				\includegraphics[width = 0.2\textwidth]{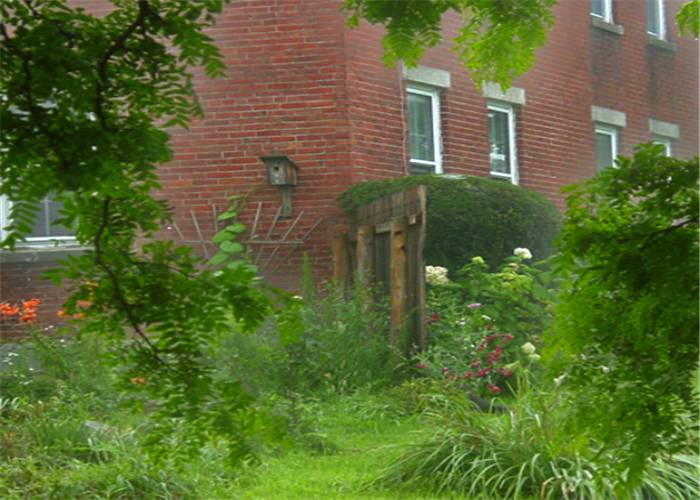} & \hspace{-0.4cm}
				\includegraphics[width = 0.2\textwidth]{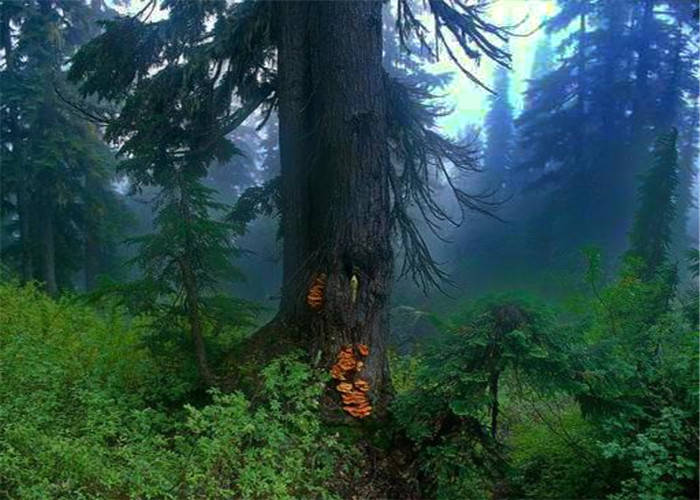} \\
				\rotatebox{90}{\large{(c) Meng \cite{meng2013ICCV}}} & \hspace{-0.4cm}
				\includegraphics[width = 0.2\textwidth]{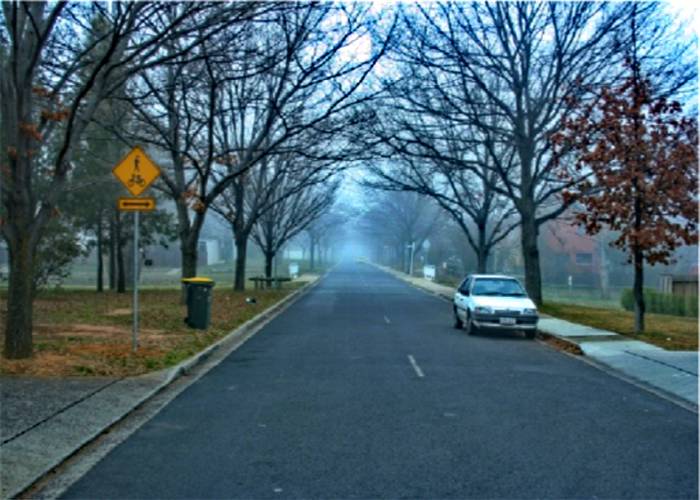} & \hspace{-0.4cm}
				\includegraphics[width = 0.2\textwidth]{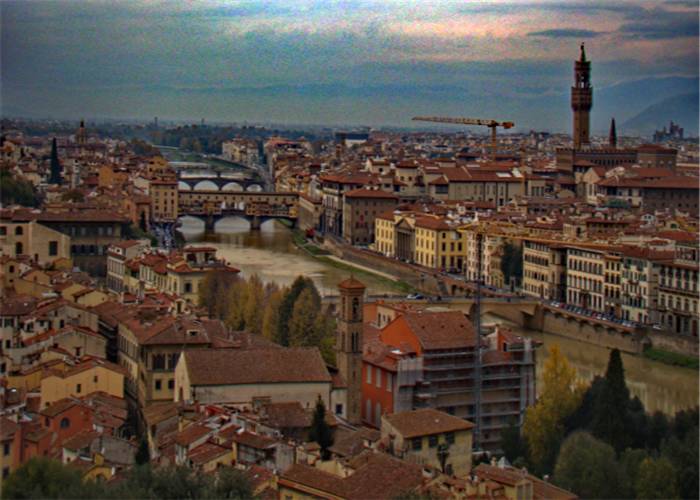} & \hspace{-0.4cm}
				\includegraphics[width = 0.2\textwidth]{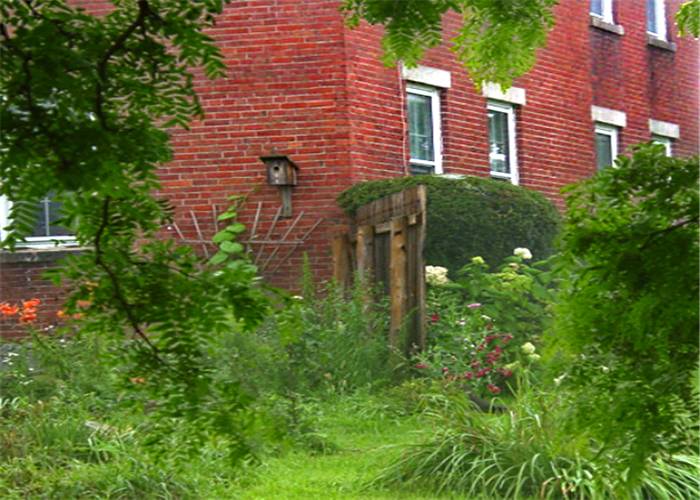} & \hspace{-0.4cm}
				\includegraphics[width = 0.2\textwidth]{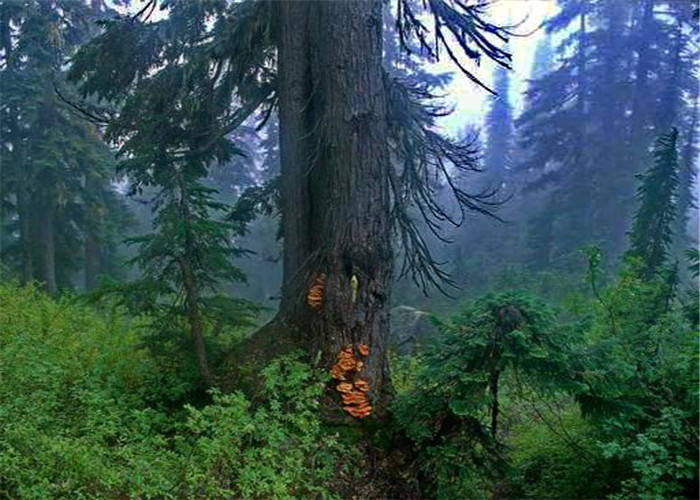} \\
				\rotatebox[origin=lt]{90}{\large{(d) Zhu \cite{zhu2015TIP}}} & \hspace{-0.4cm}
				\includegraphics[width = 0.2\textwidth]{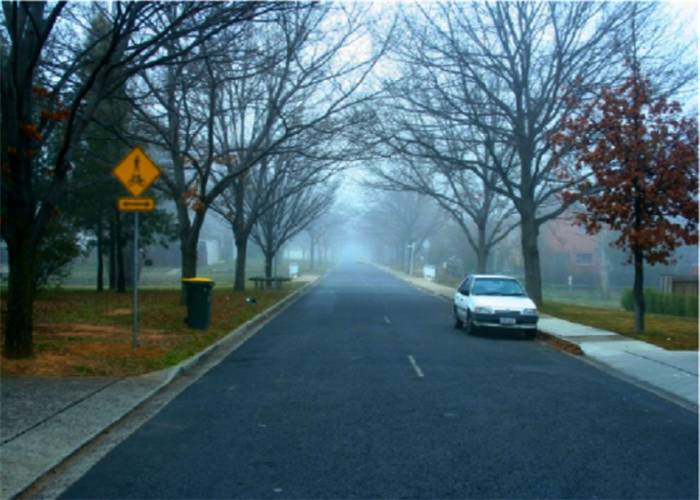} & \hspace{-0.4cm}
				\includegraphics[width = 0.2\textwidth]{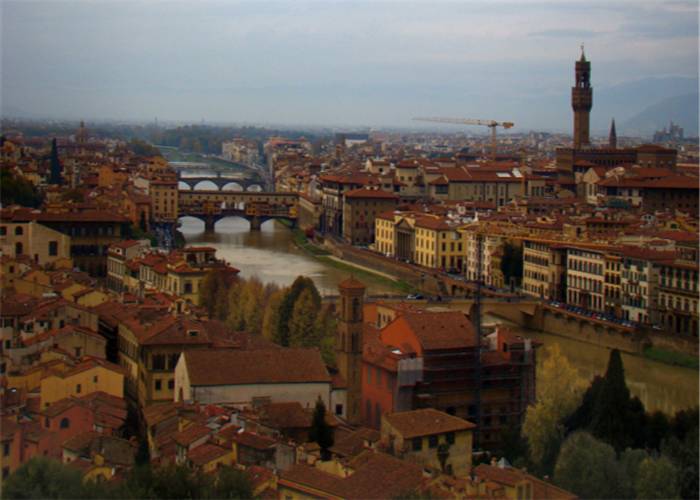} & \hspace{-0.4cm}
				\includegraphics[width = 0.2\textwidth]{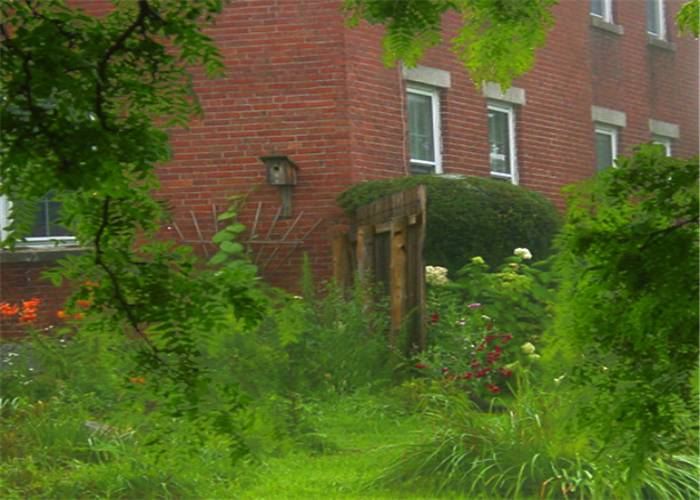} & \hspace{-0.4cm}
				\includegraphics[width = 0.2\textwidth]{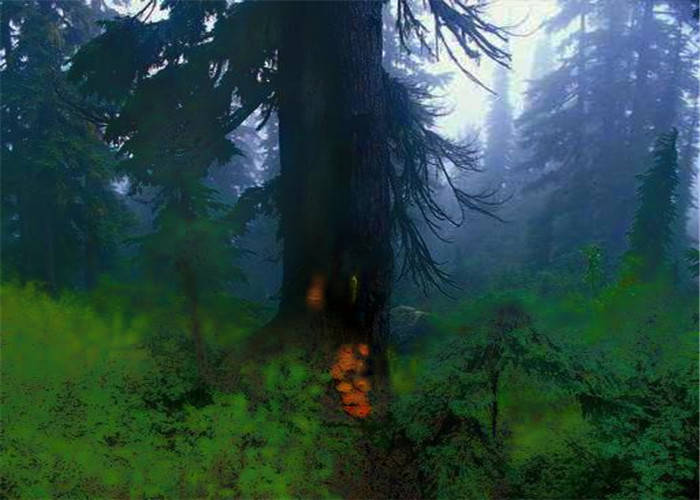} \\
				\rotatebox[origin=lt]{90}{\large{(e) Cai \cite{cai2016dehazenet}}} & \hspace{-0.4cm}
				\includegraphics[width = 0.2\textwidth]{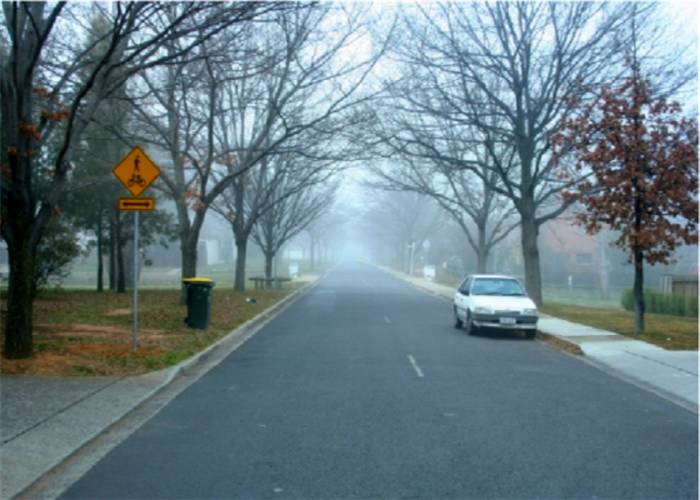} & \hspace{-0.4cm}
				\includegraphics[width = 0.2\textwidth]{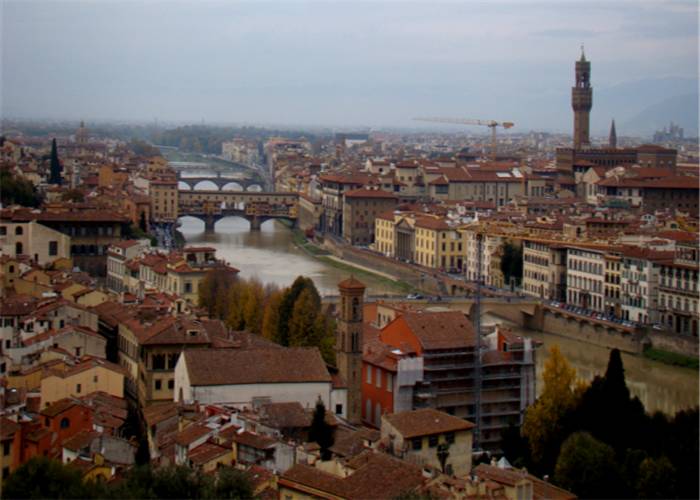} & \hspace{-0.4cm}
				\includegraphics[width = 0.2\textwidth]{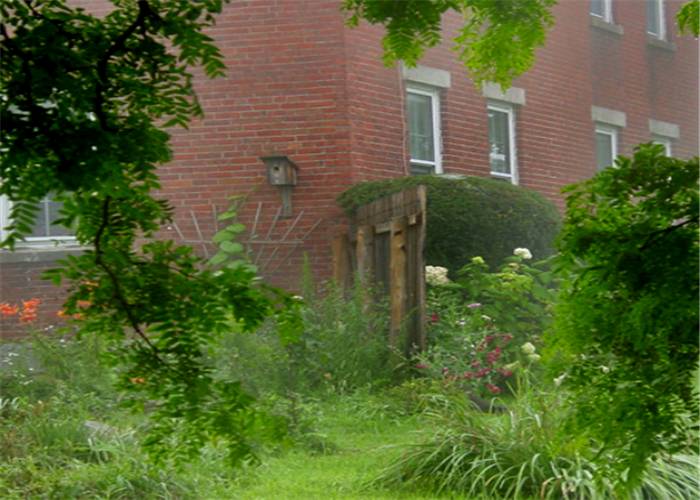} & \hspace{-0.4cm}
				\includegraphics[width = 0.2\textwidth]{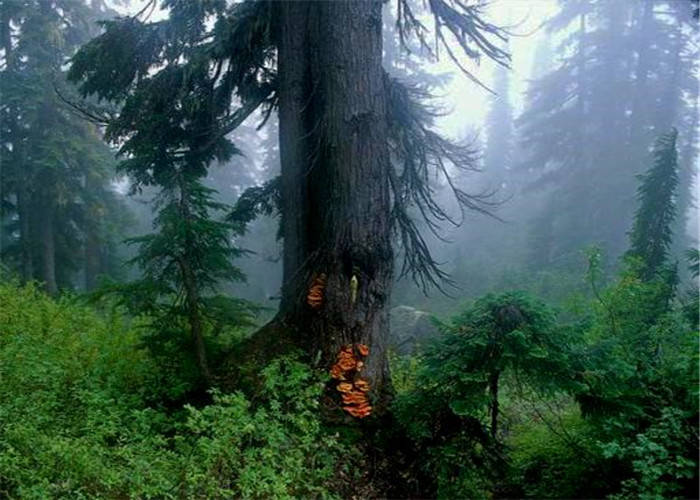} \\
				\rotatebox[origin=lt]{90}{\large{(f) Ren \cite{MSCNN2016ECCV}}} & \hspace{-0.4cm}
				\includegraphics[width = 0.2\textwidth]{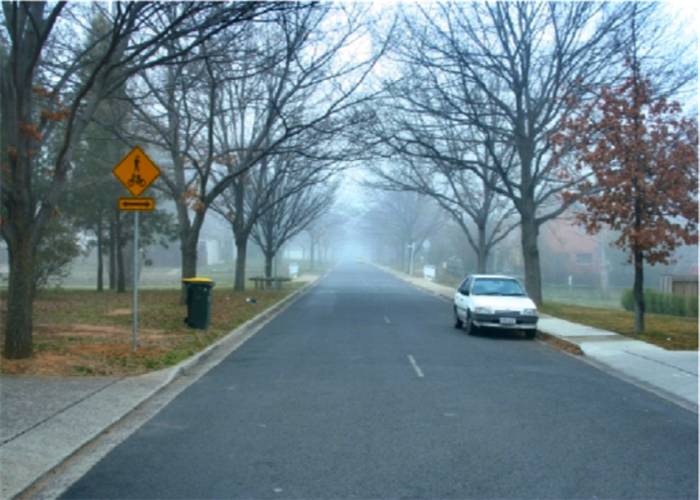} & \hspace{-0.4cm}
				\includegraphics[width = 0.2\textwidth]{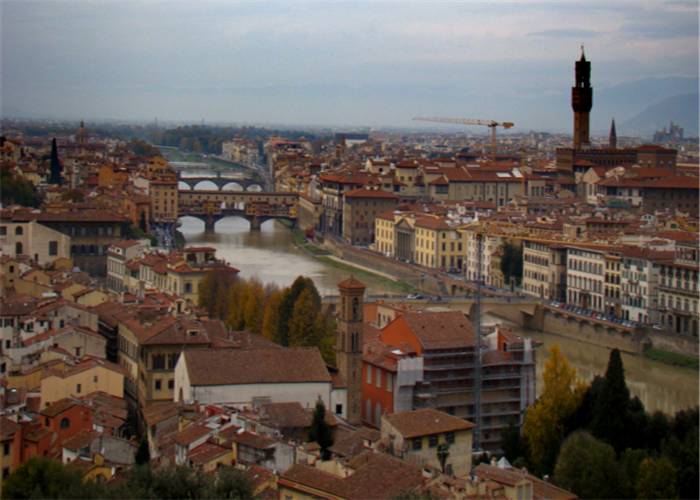} & \hspace{-0.4cm}
				\includegraphics[width = 0.2\textwidth]{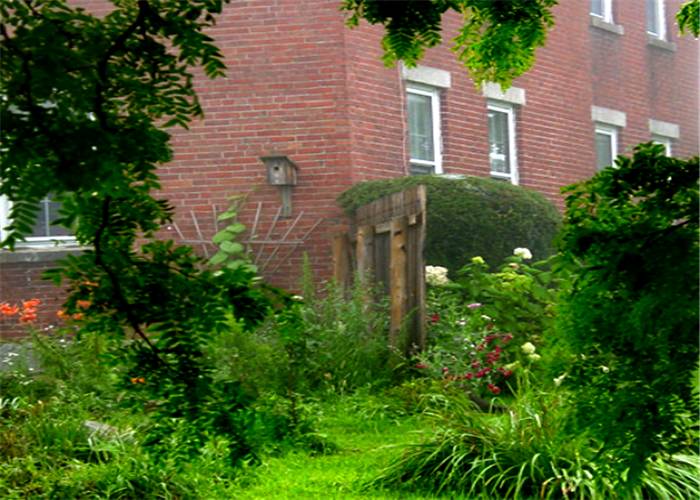} & \hspace{-0.4cm}
				\includegraphics[width = 0.2\textwidth]{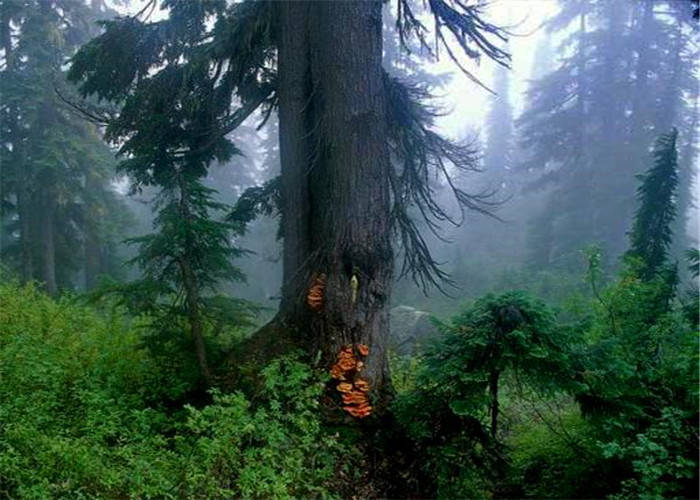} \\
				\rotatebox[origin=lt]{90}{\large{(g) Ours}} & \hspace{-0.4cm}
				\includegraphics[width = 0.2\textwidth]{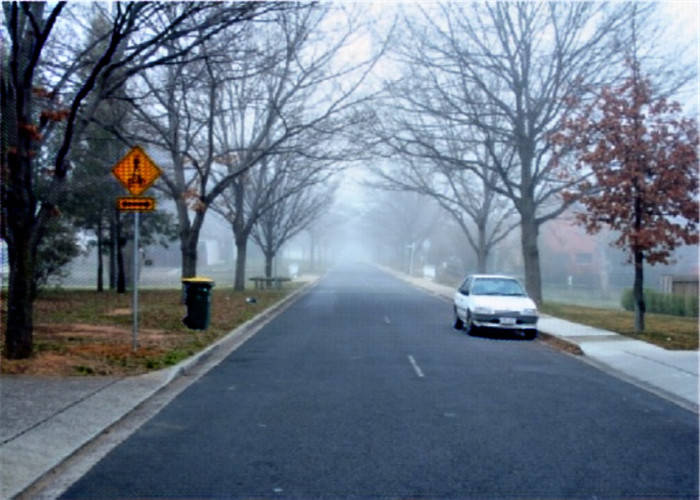} & \hspace{-0.4cm}
				\includegraphics[width = 0.2\textwidth]{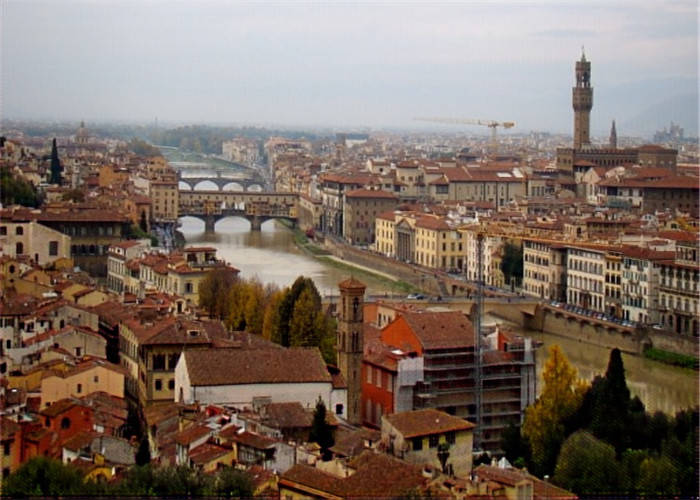} & \hspace{-0.4cm}
				\includegraphics[width = 0.2\textwidth]{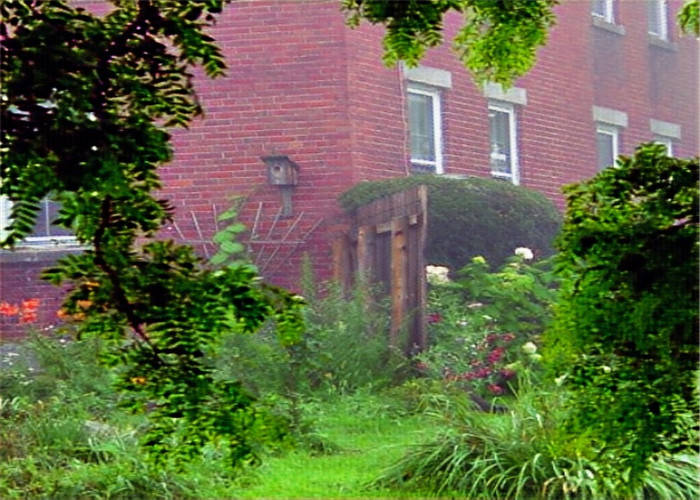} & \hspace{-0.4cm}
				\includegraphics[width = 0.2\textwidth]{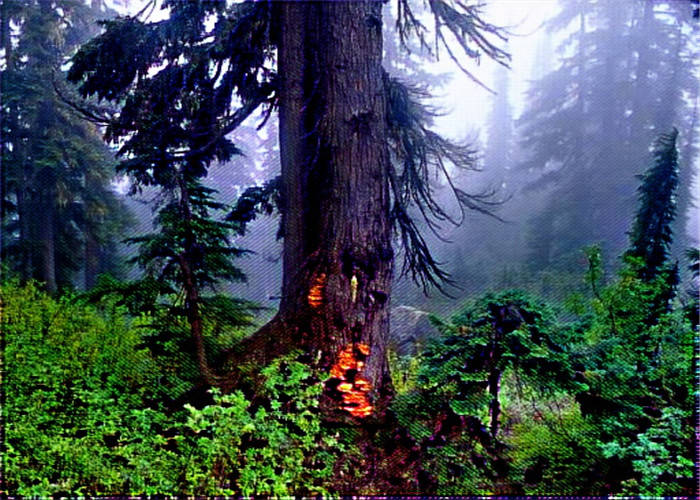} \\
				
		\end{tabular}}
	\end{center}
	\vspace{-0.3cm}
	\caption{Qualitative evaluate the defogged results on the real images.
	}
	\vspace{-0.1cm}
	\label{real-results}
\end{figure*}
\begin{figure}[t]\footnotesize
	\begin{center}
		\begin{tabular}{@{}ccccc@{}}
			\includegraphics[width = 0.09\textwidth]{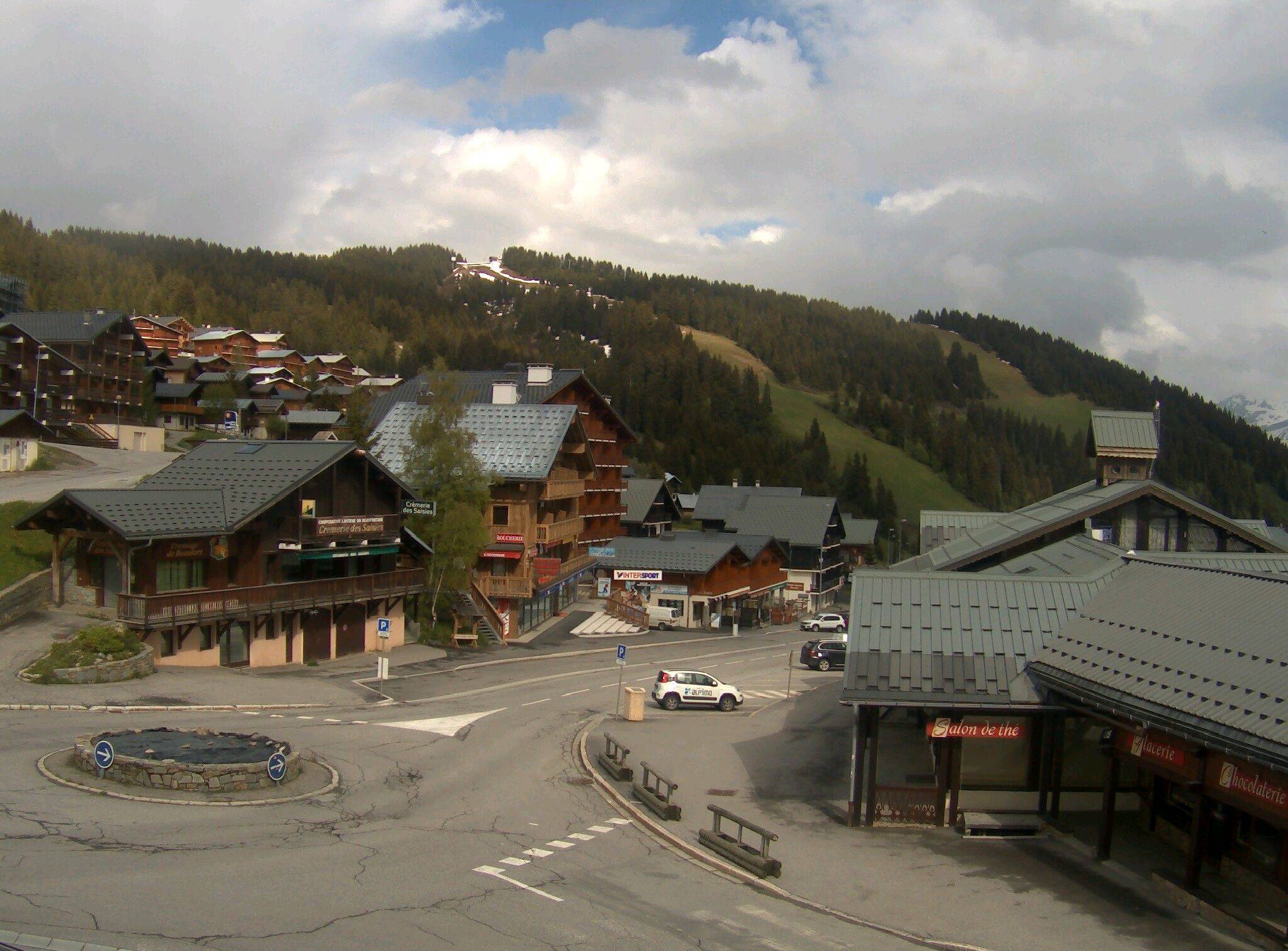} &\hspace{-0.4cm}
			\includegraphics[width = 0.09\textwidth]{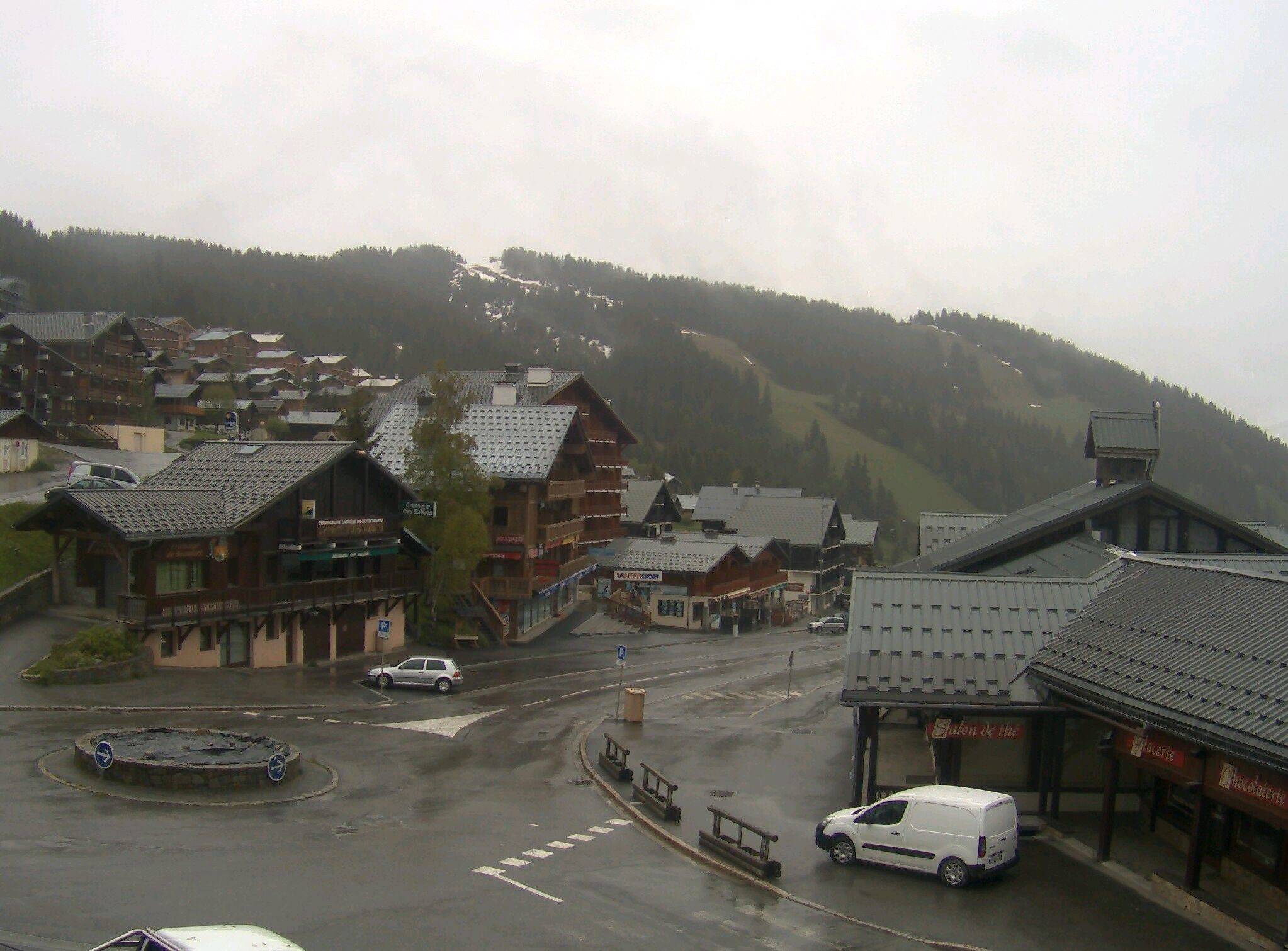} &\hspace{-0.5cm}
			\includegraphics[width = 0.09\textwidth]{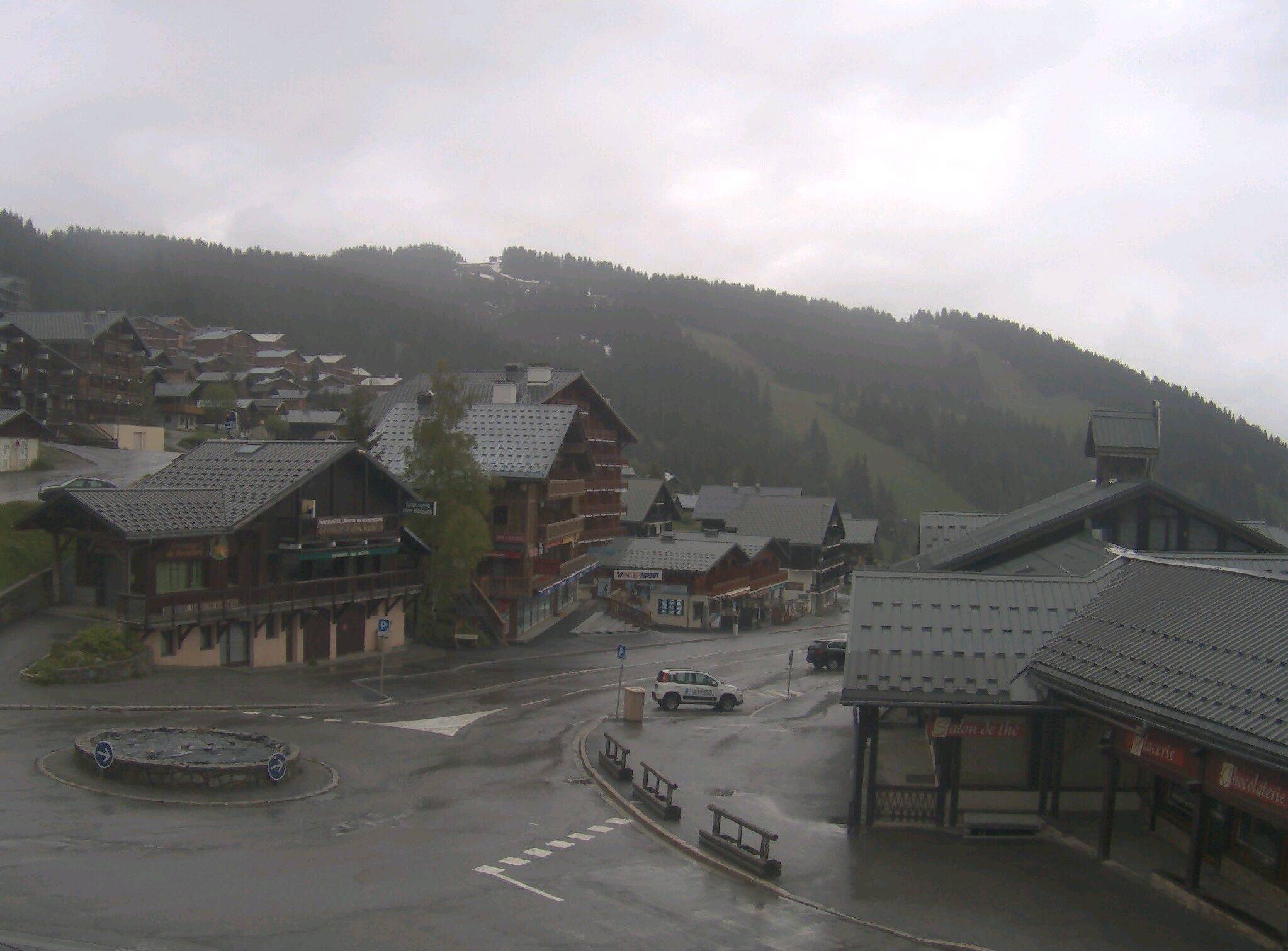} &\hspace{-0.5cm}
			\includegraphics[width = 0.09\textwidth]{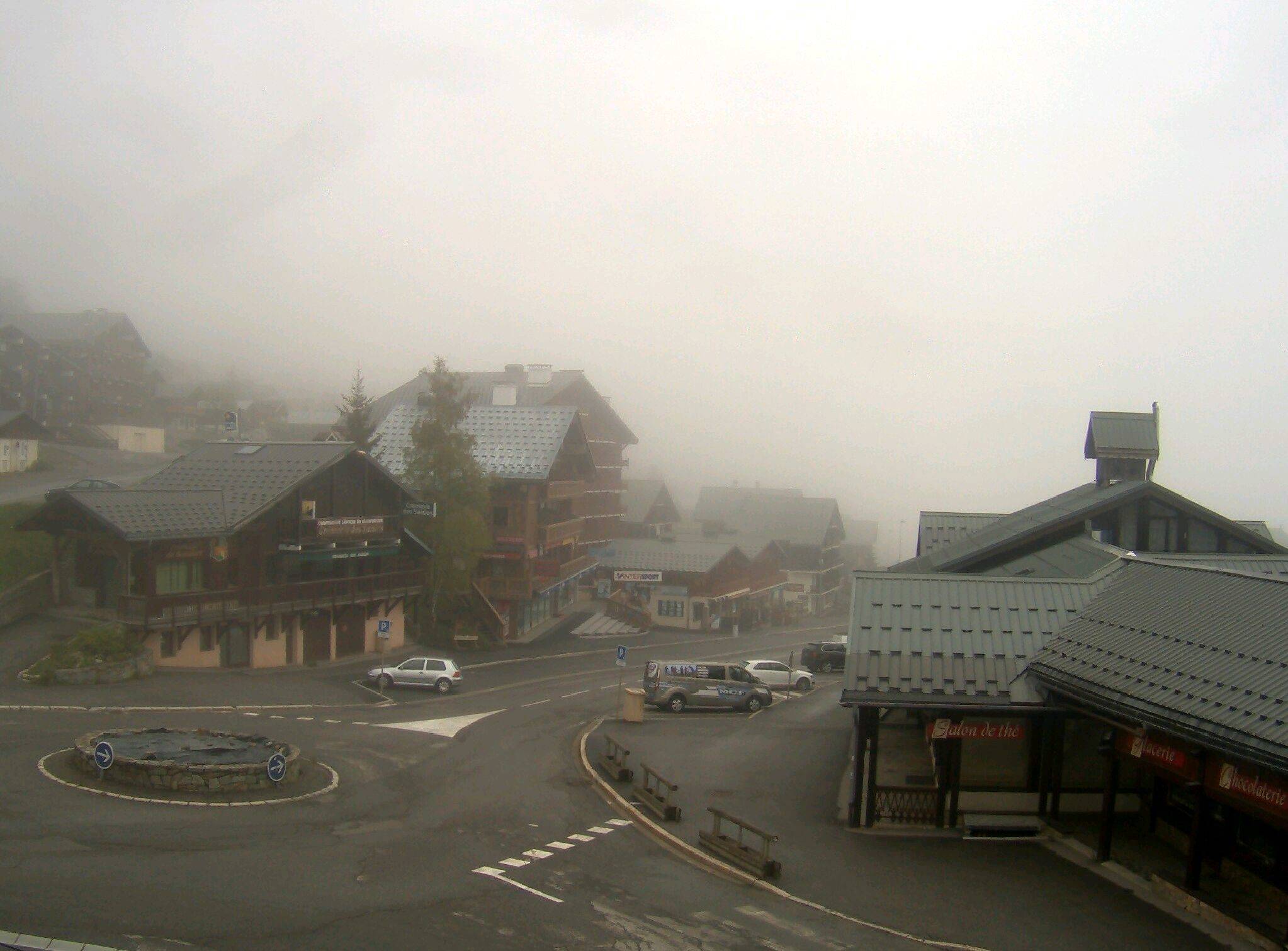} &\hspace{-0.4cm}
			\includegraphics[width = 0.09\textwidth]{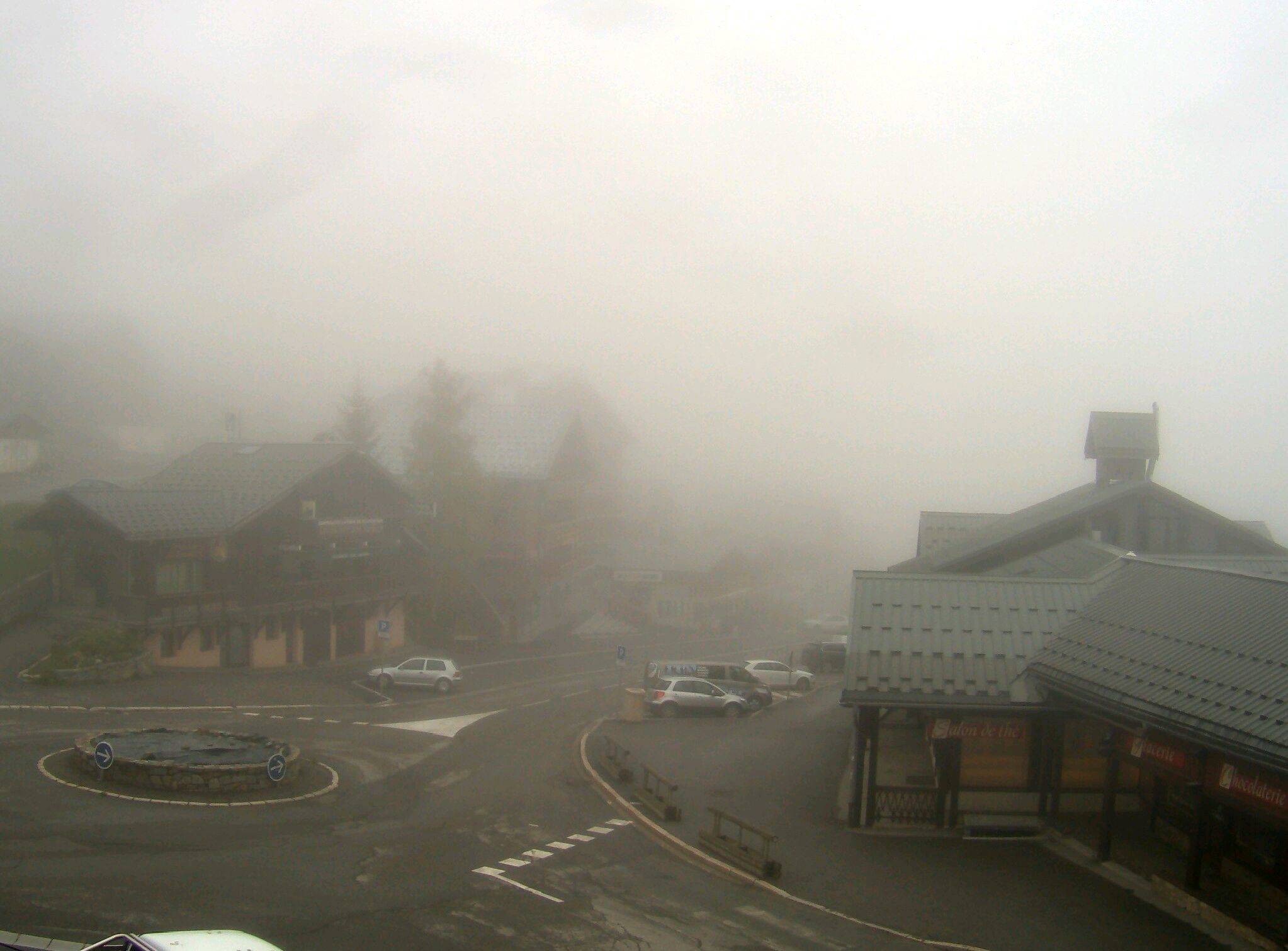} \\
			\includegraphics[width = 0.09\textwidth]{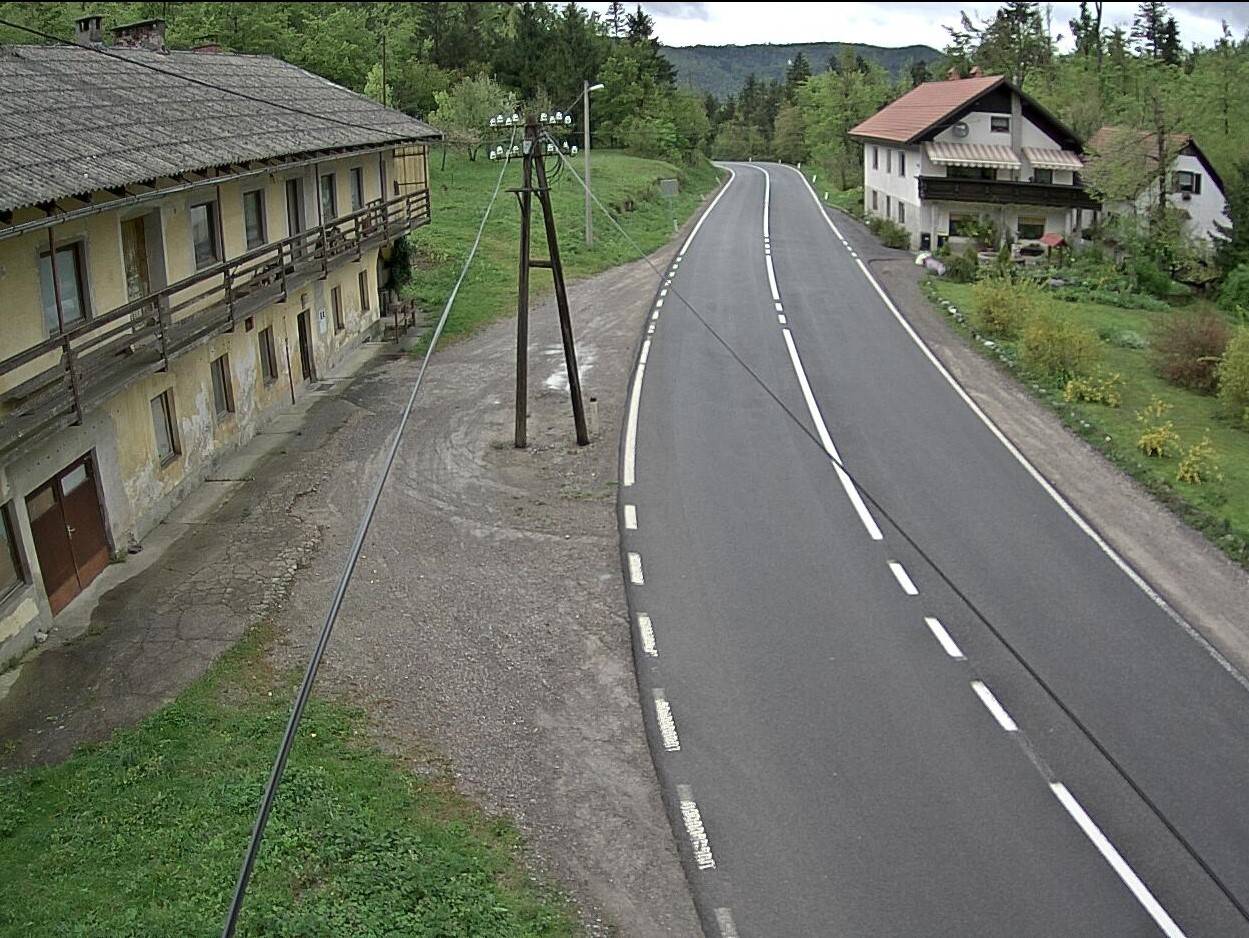} &\hspace{-0.4cm}
			\includegraphics[width = 0.09\textwidth]{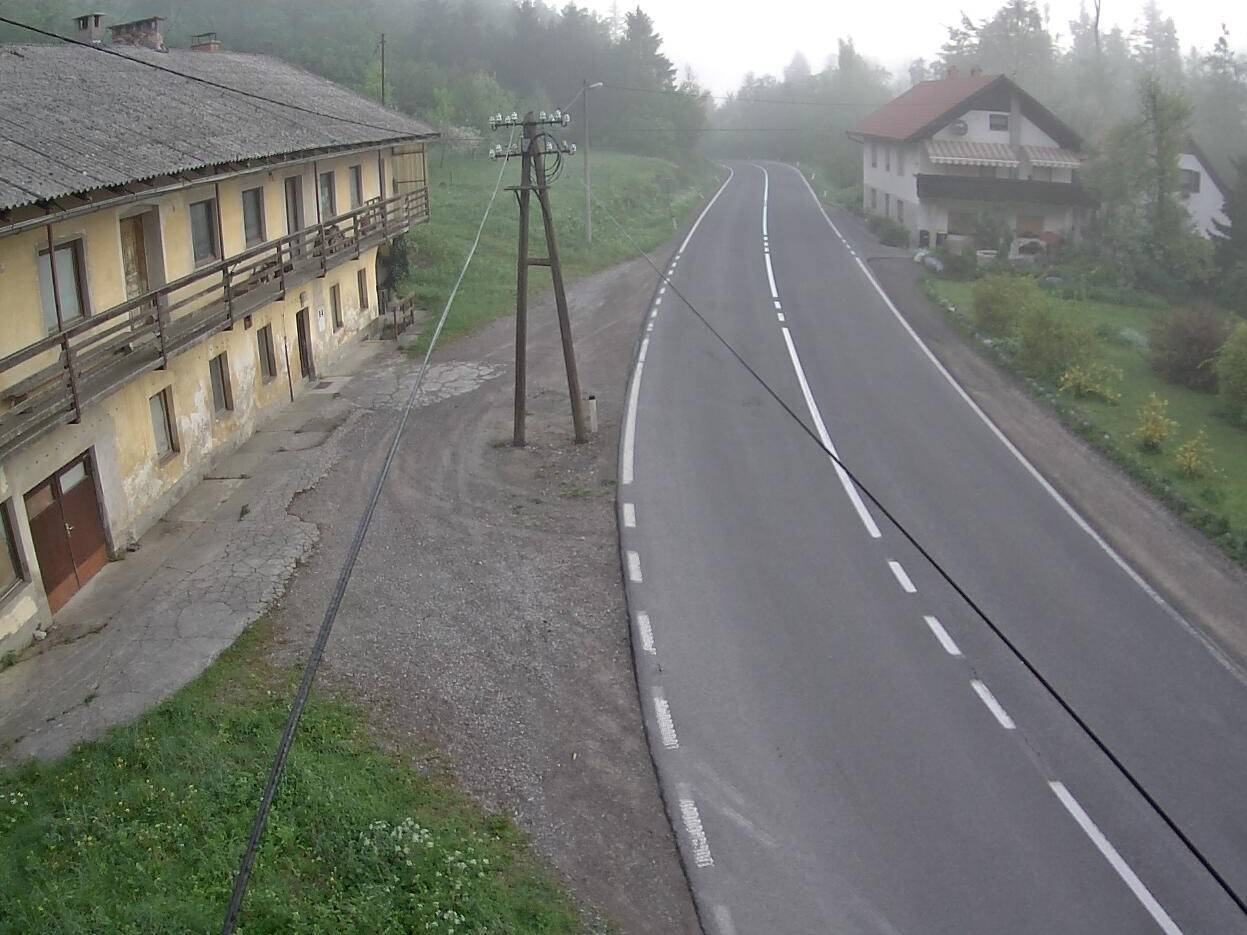} &\hspace{-0.5cm}
			\includegraphics[width = 0.09\textwidth]{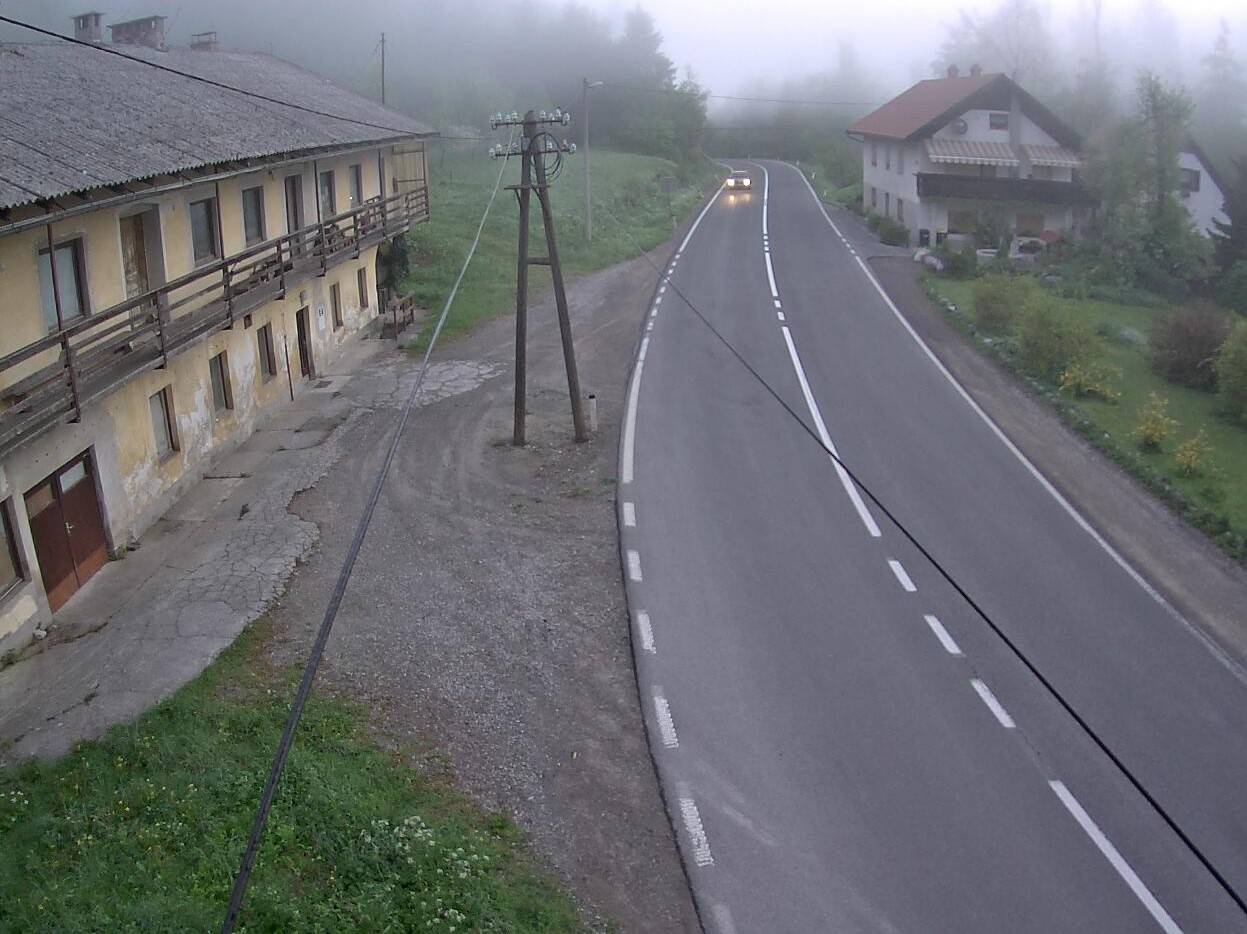} &\hspace{-0.5cm}
			\includegraphics[width = 0.09\textwidth]{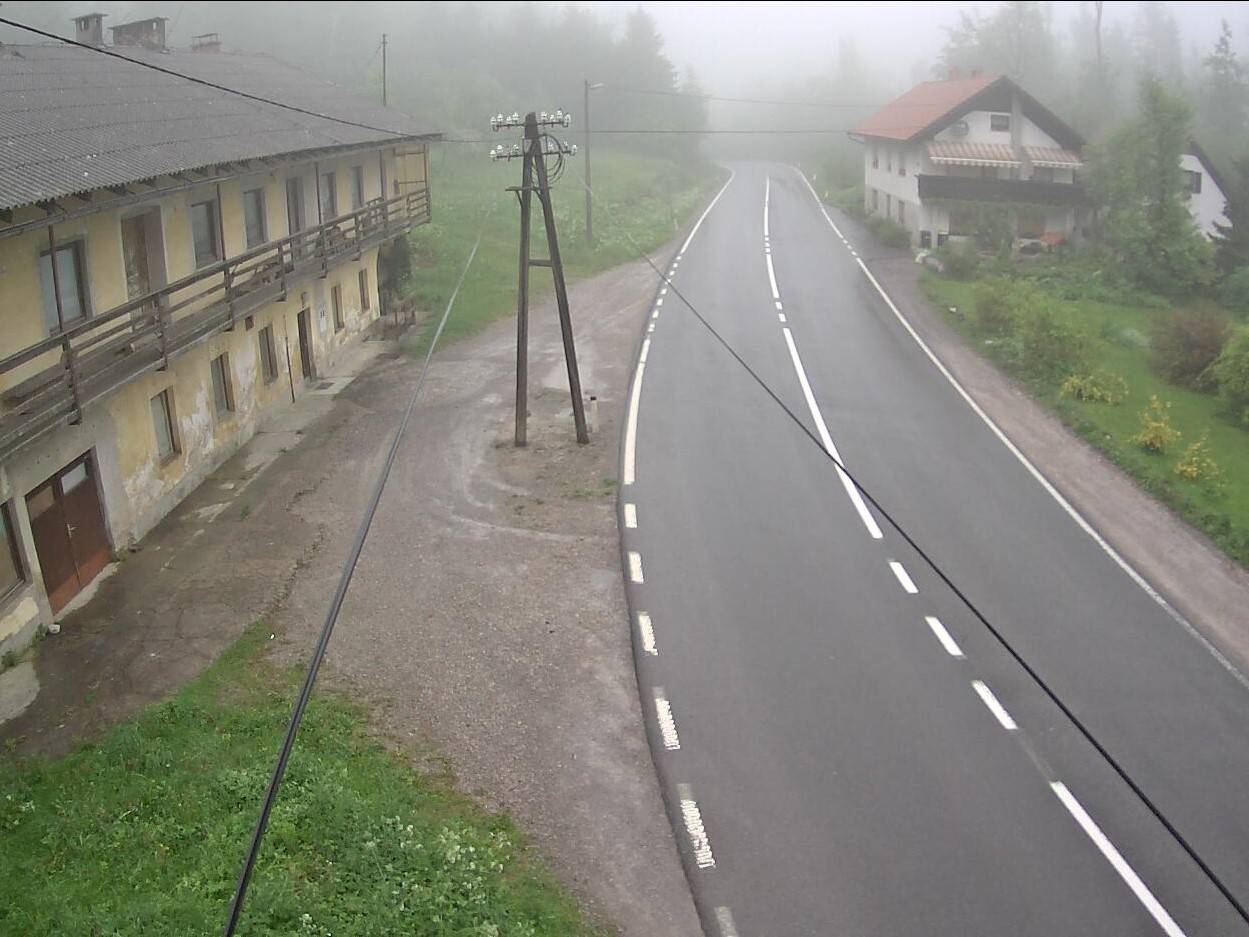} &\hspace{-0.4cm}
			\includegraphics[width = 0.09\textwidth]{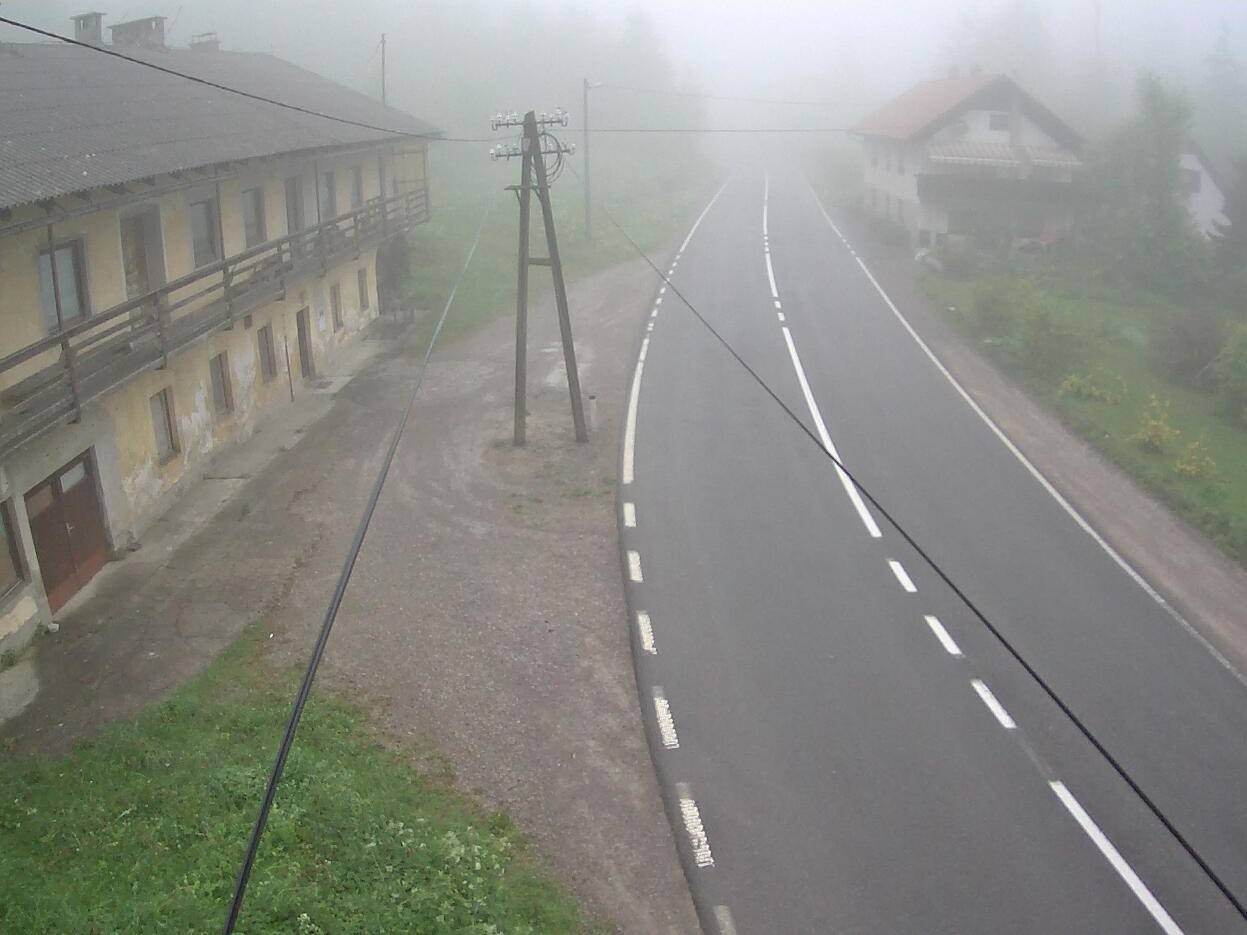} \\
			\includegraphics[width = 0.09\textwidth]{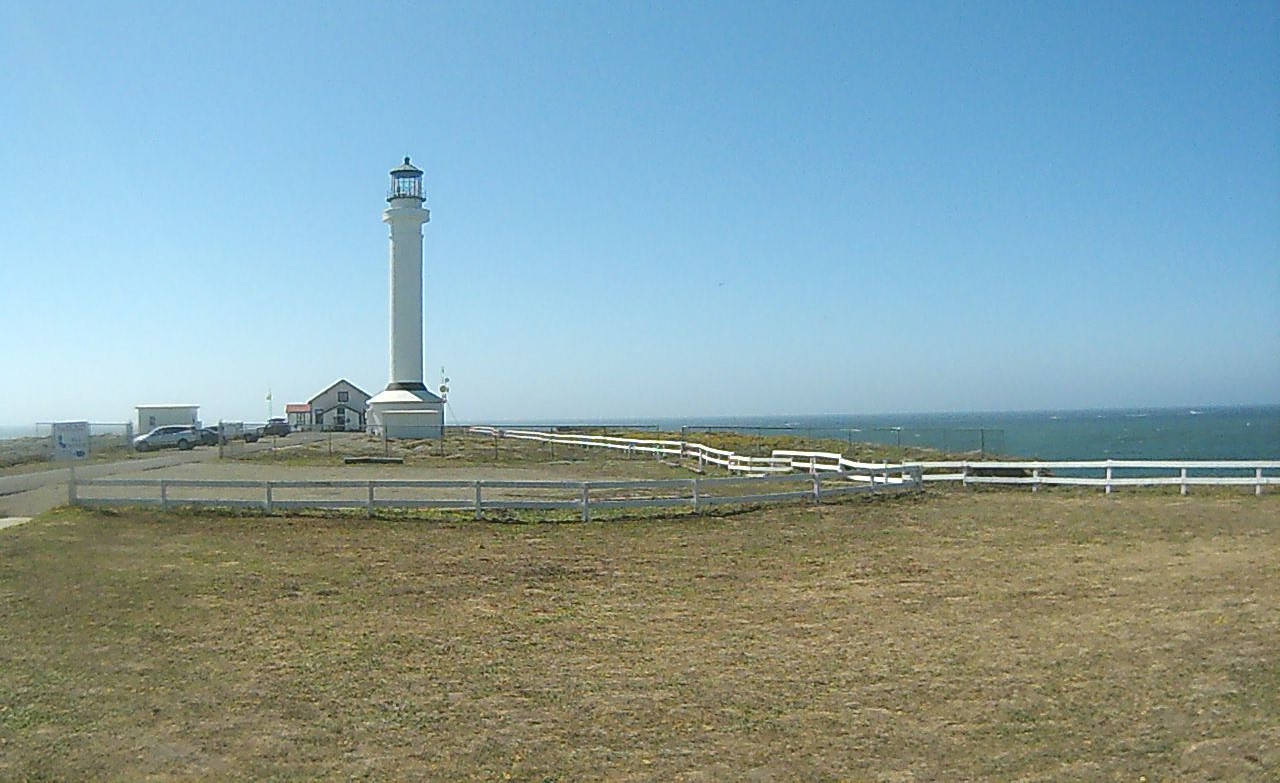} &\hspace{-0.4cm}
			\includegraphics[width = 0.09\textwidth]{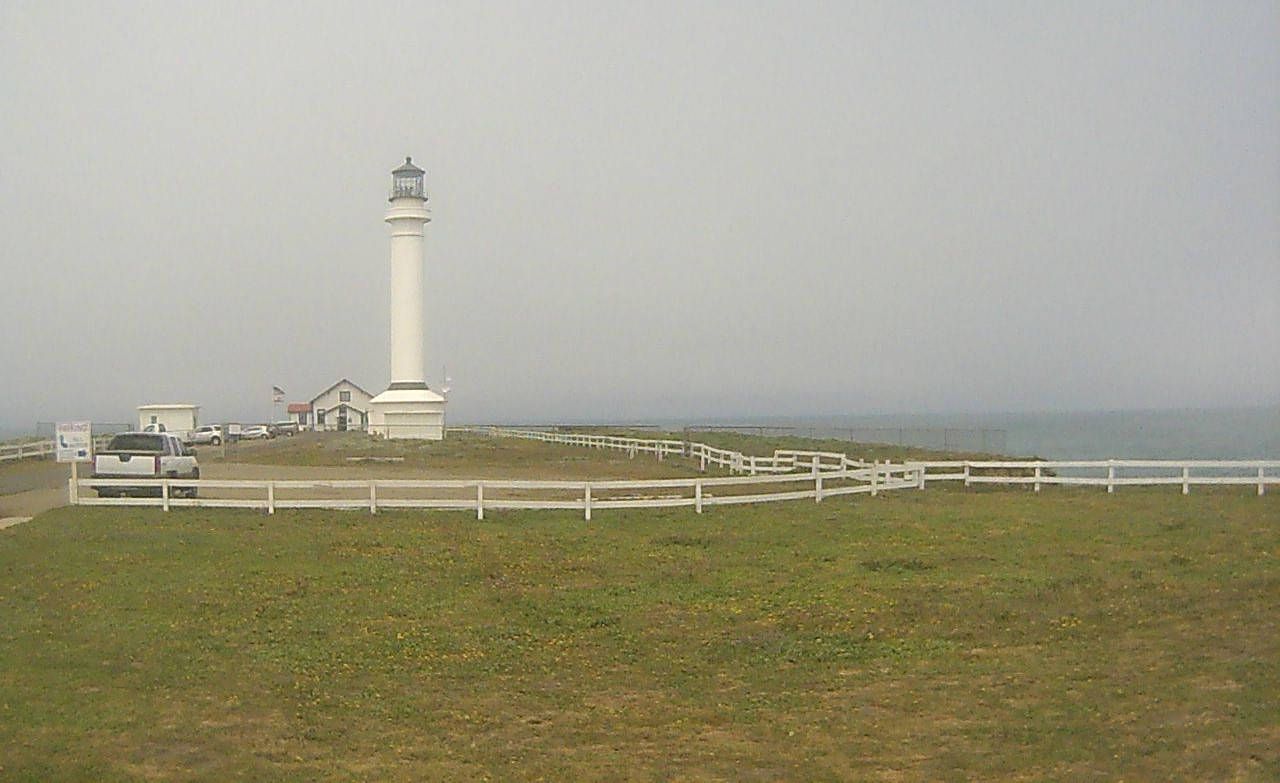} &\hspace{-0.5cm}
			\includegraphics[width = 0.09\textwidth]{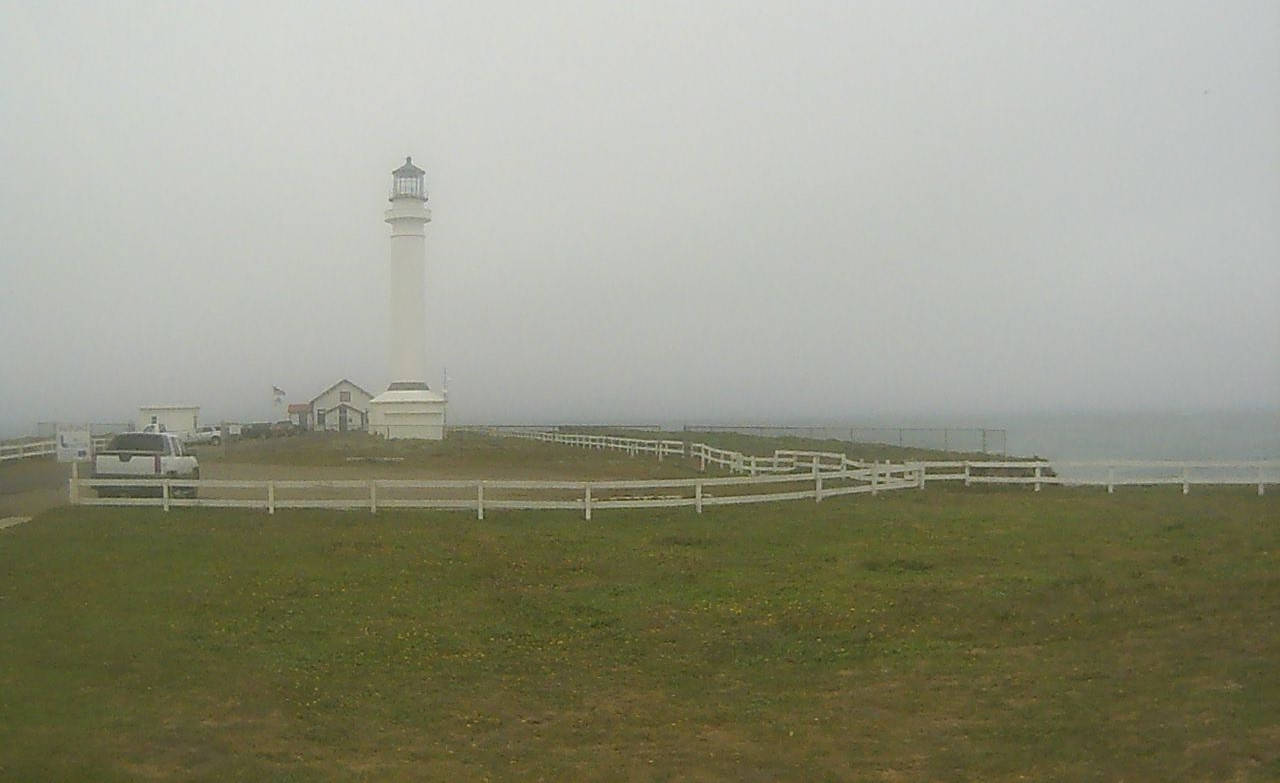} &\hspace{-0.5cm}
			\includegraphics[width = 0.09\textwidth]{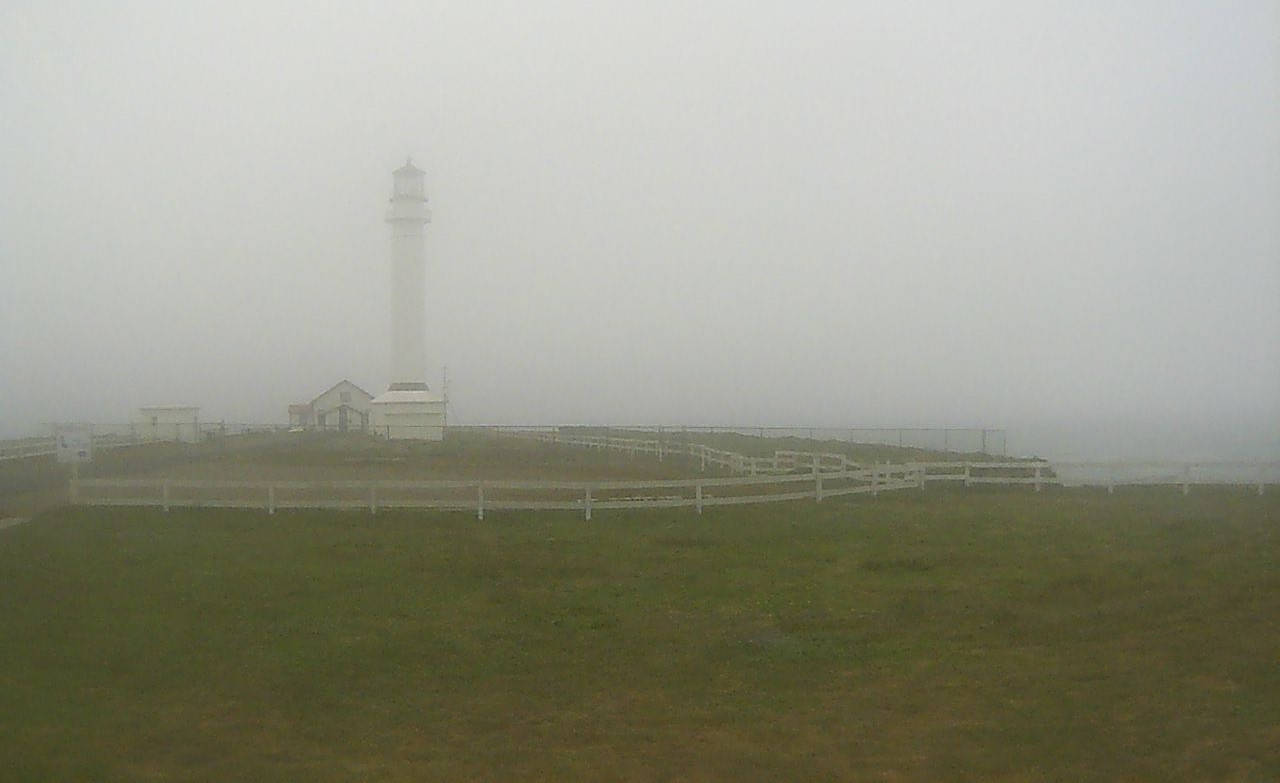} &\hspace{-0.4cm}
			\includegraphics[width = 0.09\textwidth]{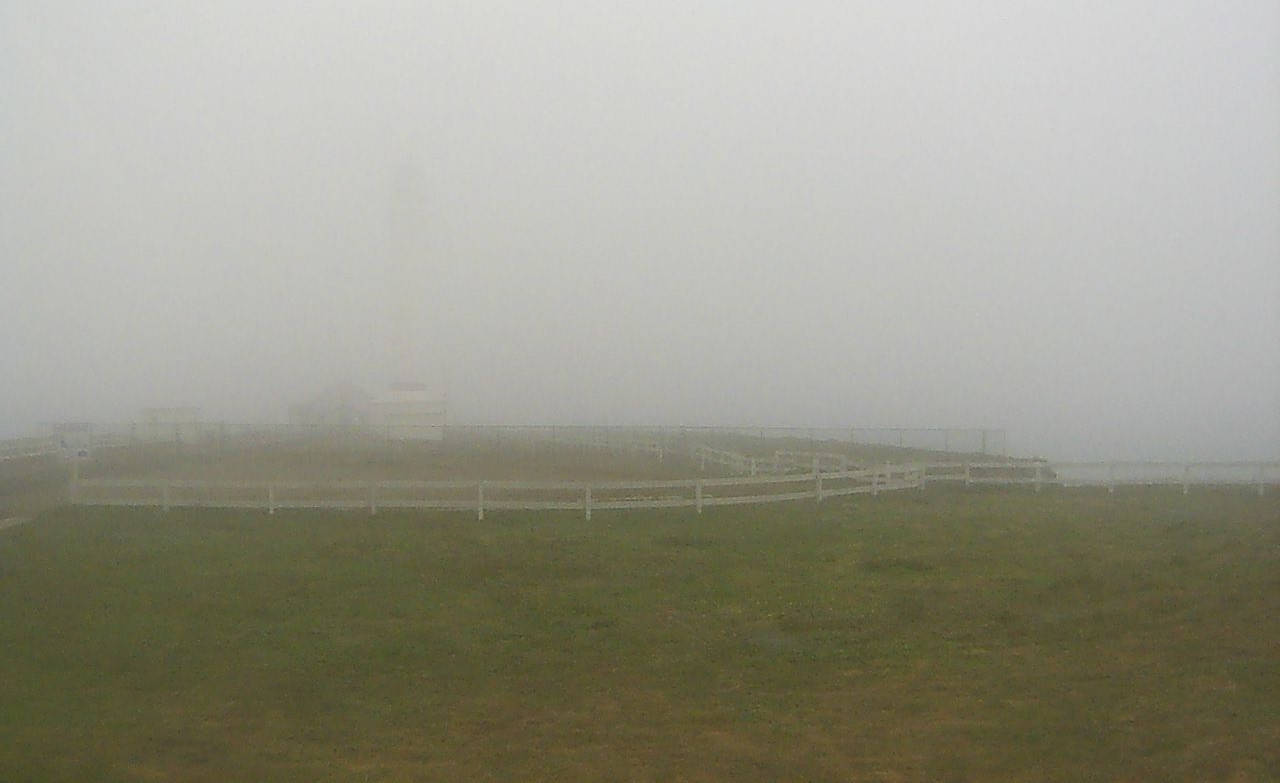} \\
			(a) Clear & \hspace{-0.4cm}
			(b) Slightly & \hspace{-0.4cm}
			(c) Moderately & \hspace{-0.4cm}
			(d) Highly & \hspace{-0.4cm}
			(e) Extremely
		\end{tabular}
	\end{center}
	\vspace{-0.5cm}
	\caption{ The image samples from MRFID.
	}
	\vspace{-0.3cm}
	\label{MFID-samples}
\end{figure}
\begin{figure*}[htbp]\scriptsize
	\begin{center}
		\begin{tabular}{@{}ccccc@{}} 
		    \rotatebox[origin=lt]{90}{\large{(a) Inputs}} & \hspace{-0.4cm}
			\includegraphics[width = 0.25\textwidth]{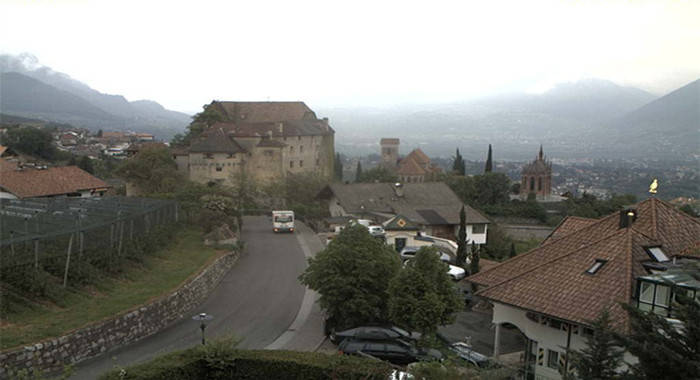} & \hspace{-0.4cm}
			\includegraphics[width = 0.25\textwidth]{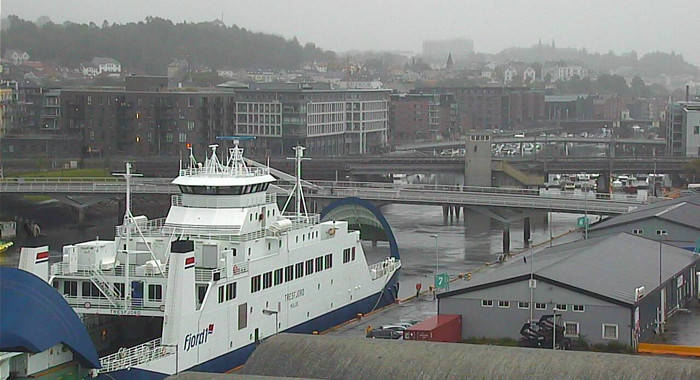} & \hspace{-0.4cm}
			\includegraphics[width = 0.25\textwidth]{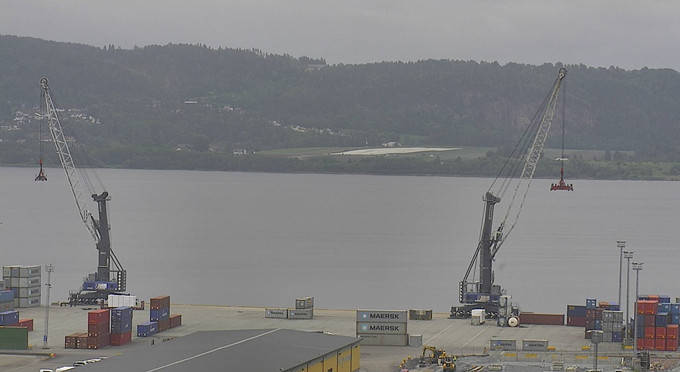} & \hspace{-0.4cm}
			\includegraphics[width = 0.25\textwidth]{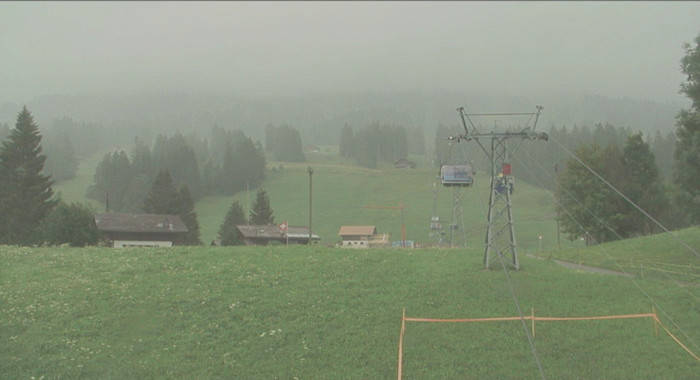} \\
			\rotatebox[origin=lt]{90}{\large{(b) He \cite{he2011PAMI}}} & \hspace{-0.4cm}
 			\includegraphics[width = 0.25\textwidth]{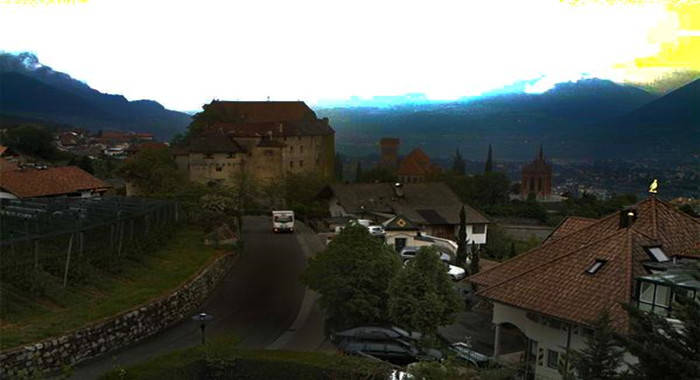} & \hspace{-0.4cm}
			\includegraphics[width = 0.25\textwidth]{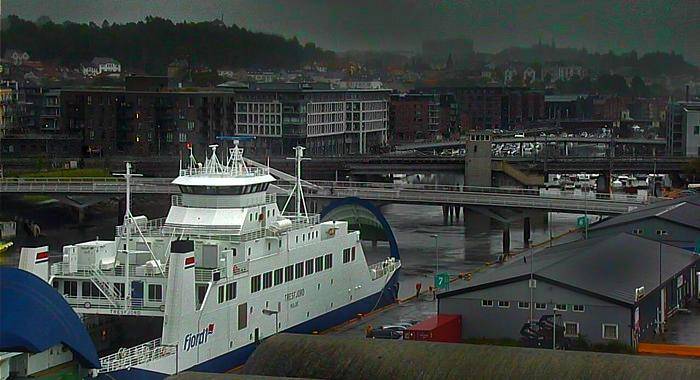} & \hspace{-0.4cm}
			\includegraphics[width = 0.25\textwidth]{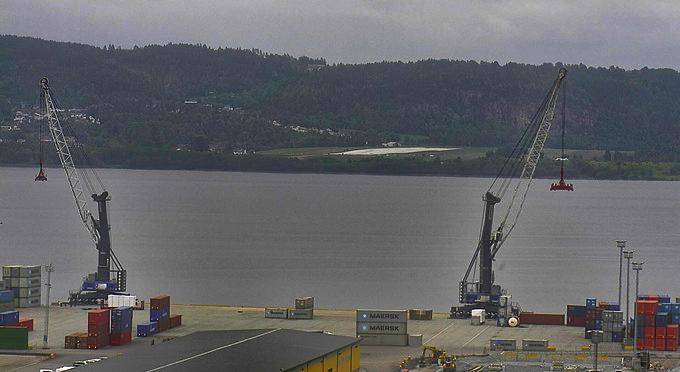} & \hspace{-0.4cm}
			\includegraphics[width = 0.25\textwidth]{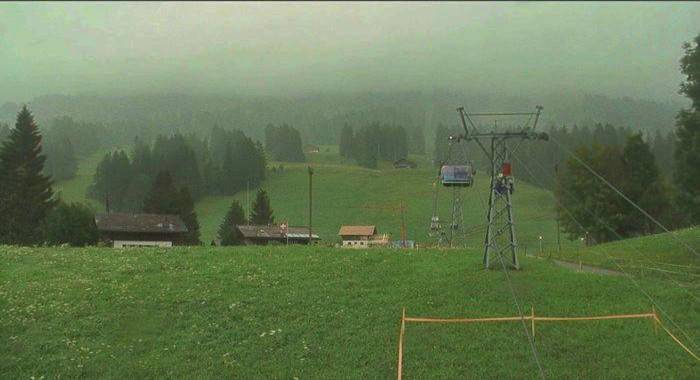} \\
			\rotatebox{90}{\large{(c) Meng \cite{meng2013ICCV}}} & \hspace{-0.4cm}
			\includegraphics[width = 0.25\textwidth]{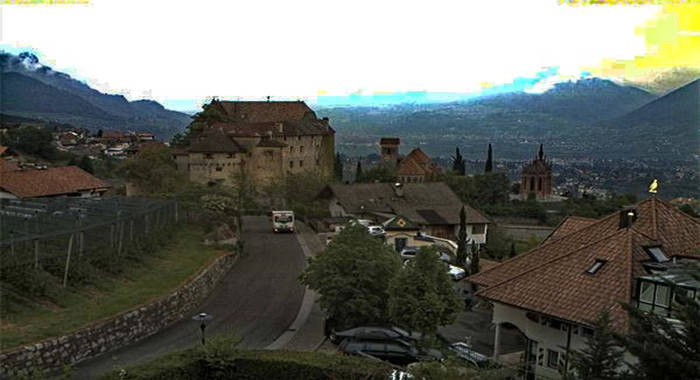} & \hspace{-0.4cm}
			\includegraphics[width = 0.25\textwidth]{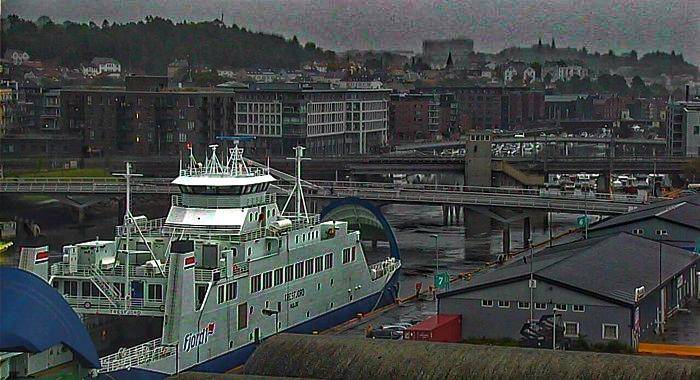} & \hspace{-0.4cm}
			\includegraphics[width = 0.25\textwidth]{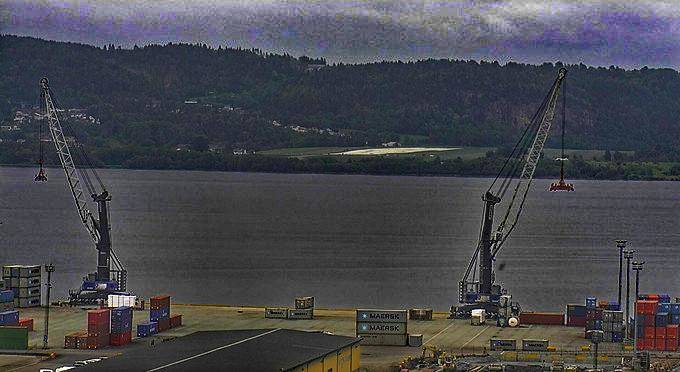} & \hspace{-0.4cm}
			\includegraphics[width = 0.25\textwidth]{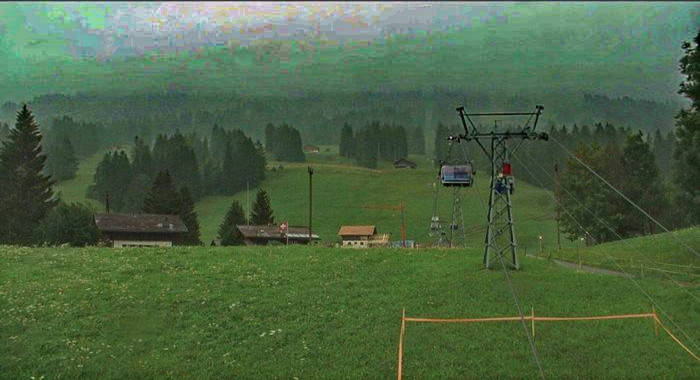} \\
			\rotatebox[origin=lt]{90}{\large{(d) Zhu \cite{zhu2015TIP}}} & \hspace{-0.4cm}
			\includegraphics[width = 0.25\textwidth]{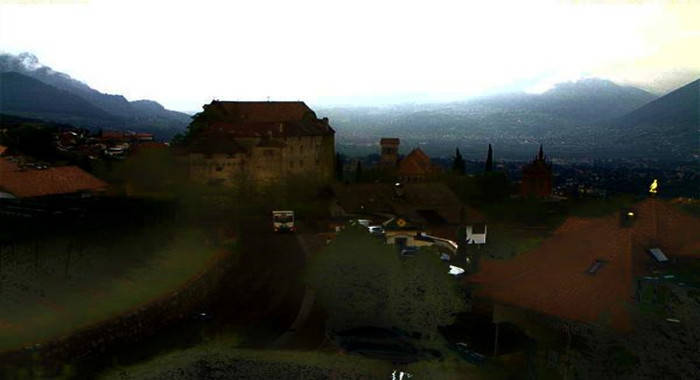} & \hspace{-0.4cm}
			\includegraphics[width = 0.25\textwidth]{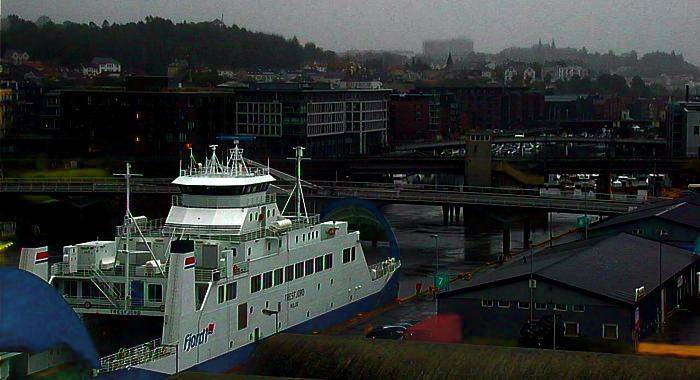} & \hspace{-0.4cm}
			\includegraphics[width = 0.25\textwidth]{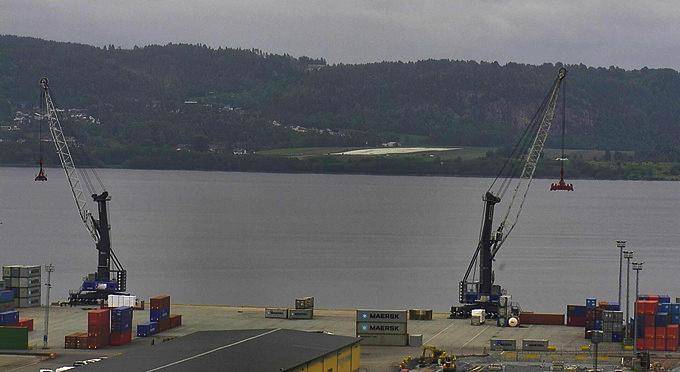} & \hspace{-0.4cm}
			\includegraphics[width = 0.25\textwidth]{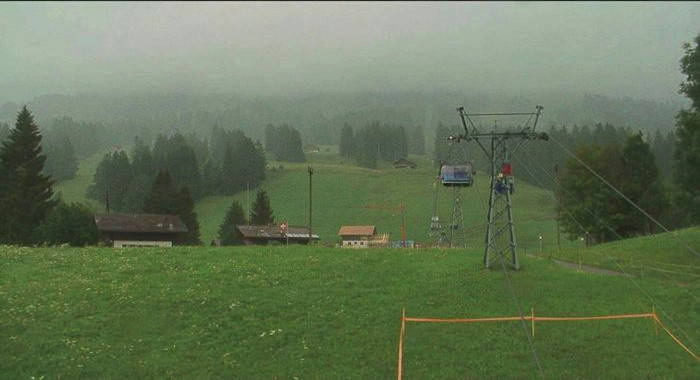} \\
			\rotatebox[origin=lt]{90}{\large{(e) Cai \cite{cai2016dehazenet}}} & \hspace{-0.4cm}
			\includegraphics[width = 0.25\textwidth]{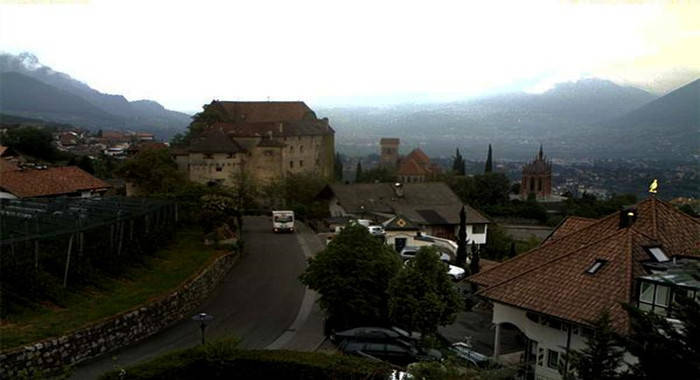} & \hspace{-0.4cm}
			\includegraphics[width = 0.25\textwidth]{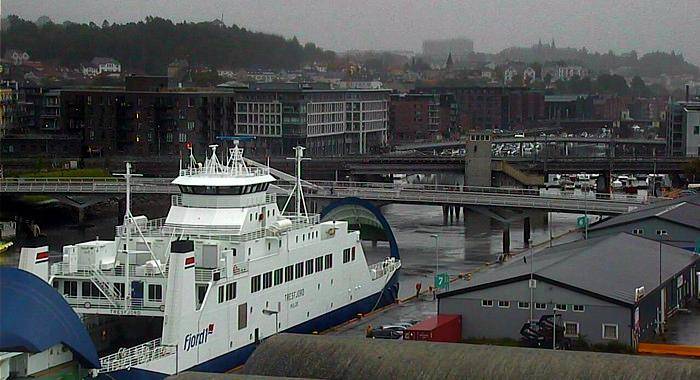} & \hspace{-0.4cm}
			\includegraphics[width = 0.25\textwidth]{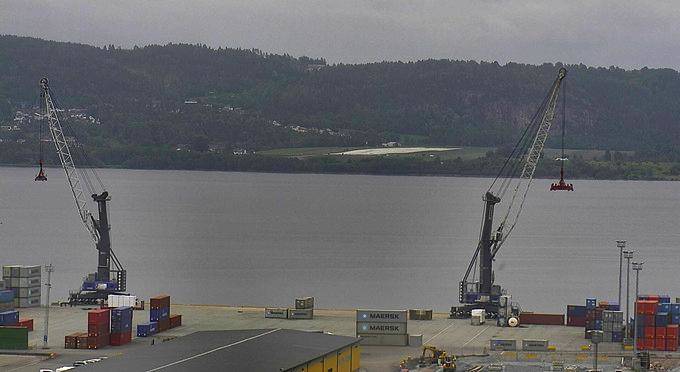} & \hspace{-0.4cm}
			\includegraphics[width = 0.25\textwidth]{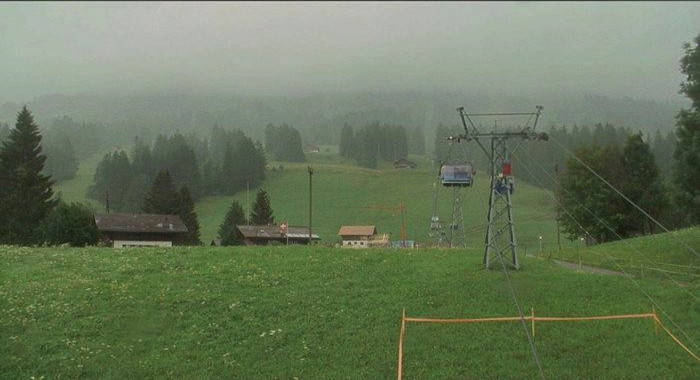} \\
			\rotatebox[origin=lt]{90}{\large{(f) Ren \cite{MSCNN2016ECCV}}} & \hspace{-0.4cm}
			\includegraphics[width = 0.25\textwidth]{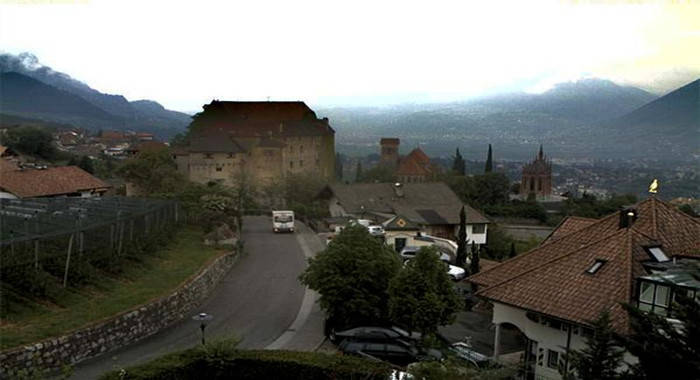} & \hspace{-0.4cm}
			\includegraphics[width = 0.25\textwidth]{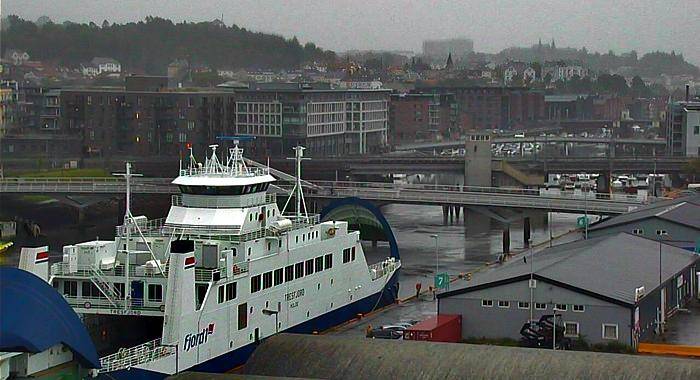} & \hspace{-0.4cm}
			\includegraphics[width = 0.25\textwidth]{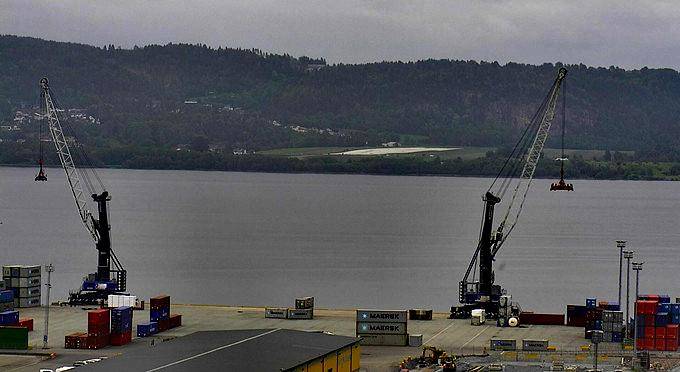} & \hspace{-0.4cm}
			\includegraphics[width = 0.25\textwidth]{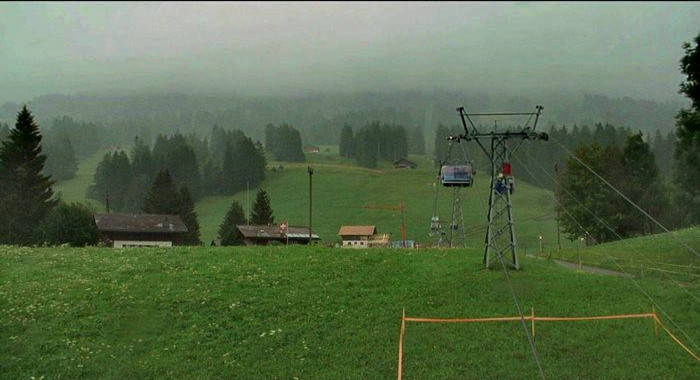} \\
			\rotatebox[origin=lt]{90}{\large{(g) Ours}} & \hspace{-0.4cm}
			\includegraphics[width = 0.25\textwidth]{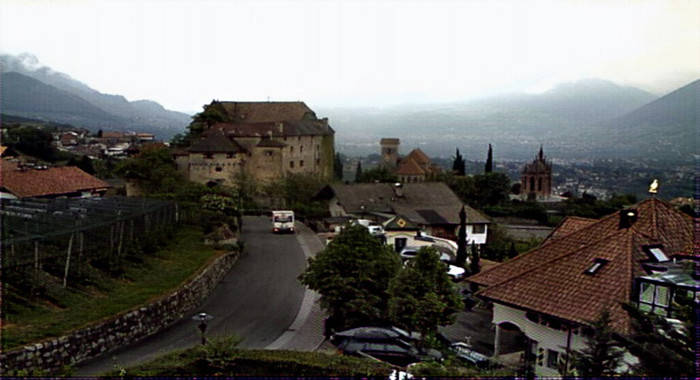} & \hspace{-0.4cm}
			\includegraphics[width = 0.25\textwidth]{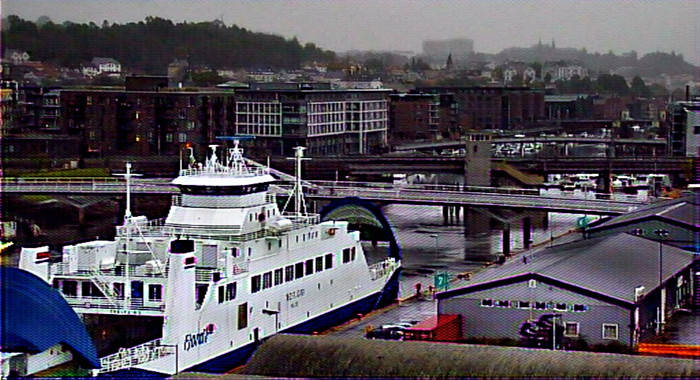} & \hspace{-0.4cm}
			\includegraphics[width = 0.25\textwidth]{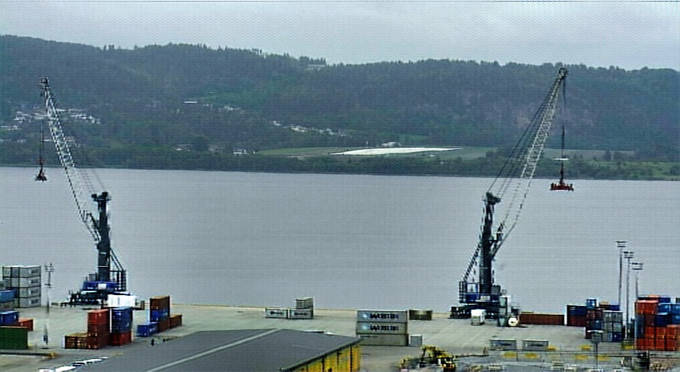} & \hspace{-0.4cm}
			\includegraphics[width = 0.25\textwidth]{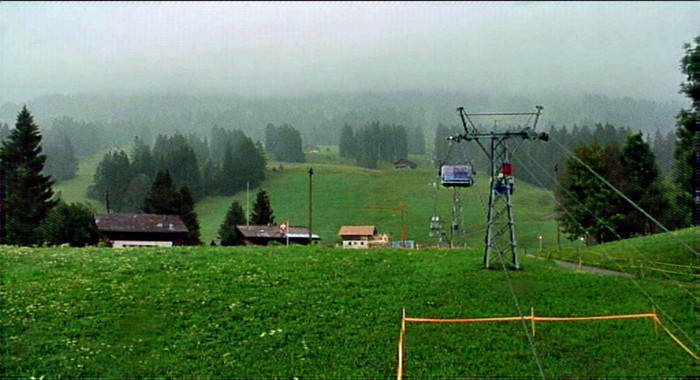} \\
		\end{tabular}
	\end{center}
	\vspace{-0.3cm}
	\caption{Qualitative comparison of different methods on the MRFID dataset.
	}
	\vspace{-0.3cm}
	\label{MFID-results}
\end{figure*}
%
\section{Experimental Results}
In this section, we qualitatively and quantitatively evaluate the defogged results by our proposed method against five state-of-the-art methods on synthetic and real-world images. The representative methods we compare our method against include Cai \textit{et.al.} \cite{cai2016dehazenet}, Ren \textit{et.al.} \cite{MSCNN2016ECCV}, He \textit{et.al.} \cite{he2011PAMI} , Zhu \textit{et.al.} \cite{zhu2015TIP} and Meng \textit{et.al.} \cite{meng2013ICCV}. Moreover, we use two evaluation criteria: Fog Aware Density Evaluator (FADE) \cite{referenceless2015TIP} and Blind Assessment based on Visibility Enhancement (BAVE) \cite{BAVE2011blind}. To the best of our knowledge, these two meastures are the few existing methods designed to quantitatively evaluate defogging performance. FADE (represented as $\mathcal{F}$) predicts the visibility of a single foggy image to evaluate the fog density. The lower value of the FADE, the better the defogged performance. In BAVE, there are three indicators to evaluate the defogged result. They are the rate of new visible edges $\textit{e}$, the quality value of the contrast restoration $\overline{\textit{r}}$ and the normalized saturate value of pixels  $\delta$. For a defogged result, the higher values of $\textit{e}$ and  $\overline{\textit{r}}$, and the smaller value of $\delta$, the better the recovered quality.

\subsection{Datasets}
Because our method does not need fog-fogfree image pairs to train the model, we do not have to synthesize the training samples, which is usually adopted by existing deep learning-based defogging methods. We collect the clear and foggy images from Google site, RESIDE dataset \cite{RESID2017reside} and our own multiple real foggy image defogging (MRFID) datasets. MRFID contains images from 200 clear outdoor scenes. For each of the clear image, there are four images of the same scene containing different densities of fog defined as slightly foggy, moderately foggy, highly foggy and extremely foggy and some examples are shown in Figure \ref{MFID-samples}. The foggy images of MRFID are selected from an image dataset called the Archive of Many Outdoor Scene (AMOS) \cite{AMOS2007CVPR}, in which images were captured by 29,945 static webcams located around the world, and contains 1,128,087,180 images from 2006 to 2017. In MRFID, images of each scene were manually selected from images taken within one year period, the image sizes range from 640$\times$480 to 22,840$\times$914. Thus, this new foggy dataset will be useful for research on deep learning based defogging methods. Moreover, in this paper, the number of the training foggy images is 12,461, and 1000 for testing. The number of the training clear images is 12,257.
\begin{center}
	\begin{table}[htbp]
		\caption{Average BAVE and FADE of defogged results on our dataset MRFID (averaged over 800 foggy images).}
		\vspace{-0.3cm}
		\setlength{\tabcolsep}{0.35mm}{
			\begin{tabular}{ccccccc}
				\toprule 
				\multirow{1}{*}{} & \multicolumn{1}{c}{He~\cite{he2011PAMI}} &\multicolumn{1}{c}{Meng~\cite{meng2013ICCV}} &\multicolumn{1}{c}{Zhu~\cite{zhu2015TIP}}
				&\multicolumn{1}{c}{Cai~\cite{cai2016dehazenet}}
				&\multicolumn{1}{c}{Ren~\cite{MSCNN2016ECCV}}
				&\multicolumn{1}{c}{Our}\\
				\midrule
				\textit{e} & 0.94 &	2.98 &	1.11 &	0.79 &	0.92 &	2.89\\
				\midrule
				$\overline{\textit{r}}$ & 1.12	& 2.09 & 0.96 &	1.08 &	1.18 & 2.58  \\
				\midrule
				$\delta(\%)$ & 0.39	& 0.79 & 3.93 &	6.86  &	2.35 &	0.35  \\
				\midrule
				$\mathcal{F}$ & 1.31 &	1.03 &	1.24 &	1.22 &	1.37 &	0.86 \\
				\bottomrule
		\end{tabular}}
		\label{MFID-quantitative}    
	\end{table}
	\vspace{-0.5cm}
\end{center}
%
\subsection{Implementation Details}
In our network, all the training samples are resized to 512$\times$512. We empirically choose $\gamma_{1}=10$, $\gamma_{2}=10$, $\gamma_{3}=8$, $\gamma_{4}=5$ and $\gamma_{5}=2$ for the loss function in generating the fog-free image. During training, ADAM optimizer is used for the generator and discriminator with the learning rate of 2$\times$$10^{-4}$ and batch size of 1. For all the defogged results in this paper, we used TensorFlow \cite{tensorflow2016} to train the network for 354,000 iterations, which takes about 126 hours on an NVidia GeForce GTX 1080 Ti GPU.

\subsection{Qualitative and Quantitative Evaluation}

For qualitative evaluation, we compare our proposed method against three prior-based and two deep learning-based state-of-the-art methods \cite{he2011PAMI} \cite{zhu2015TIP} \cite{meng2013ICCV} \cite{cai2016dehazenet} \cite{MSCNN2016ECCV} on both synthetic and real images.

\textbf{Evaluation on synthetic images:} In this experiment, the synthetic images have come from RESIDE dataset \cite{RESID2017reside}. Figure \ref{synthetic-results} shows several defogged results by our method and other defog approaches. It can be observed that there is a phenomenon of color distortion in the sky region of He's and Meng's results (as can be seen in Figure \ref{synthetic-results} (b) and Figure \ref{synthetic-results} (c)). The reason is due to these two prior-based methods overestimate the atmospheric light. Moreover, as observed in Figure \ref{synthetic-results} (d)-(g), we note that our results are similar to the defogged results of Zhu's, Cai's and Ren's methods. Although there are no artifacts in the results by using learning-based methods \cite{cai2016dehazenet}\cite{MSCNN2016ECCV}, some remaining fog is still not removed as shown in Figure \ref{synthetic-results} (e)-(f). 

In comparison, our model can generate clearer images from the single foggy images and has no artifacts, as shown in Figure \ref{synthetic-results} (g). This is because our enhancer network can improve the texture details of the outputs from generator \textit{G} and \textit{R}, which can further strengthen mapping capability and preserve the constrast and sharpness of the image. It is also reflected in the quantitative results in Table \ref{synthetic-quantitative}. It can be seen that the $\mathcal{F}$ in our results are always smaller than the other two learning based methods. Moreover, the new visible edges $\textit{e}$ and the quality value of the contrast restoration $\overline{\textit{r}}$ in our results are usually larger than others, and the normalized saturate value of pixels $\delta$ in our results are also smaller than others. It is demonstrated that our method has an overall better defog performance on synthetic images compared with other defog methods.

\begin{figure}[htbp]\scriptsize
	\begin{center}
		\begin{tabular}{@{}c@{}}
			\includegraphics[width = 0.16\textwidth]{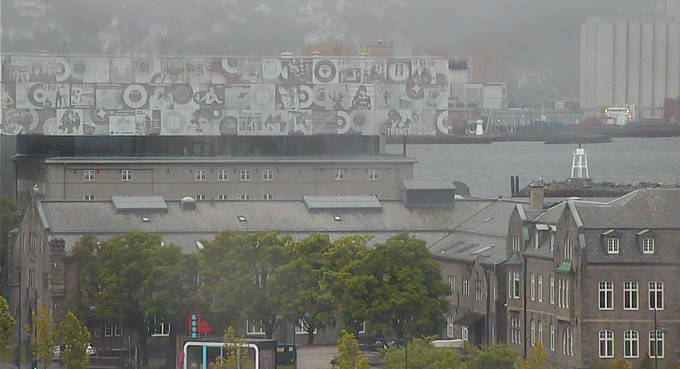} \\
			(a) Foggy images 
		\end{tabular}
	\end{center}
	\begin{center}
		\begin{tabular}{@{}ccc@{}}
			\includegraphics[width = 0.16\textwidth]{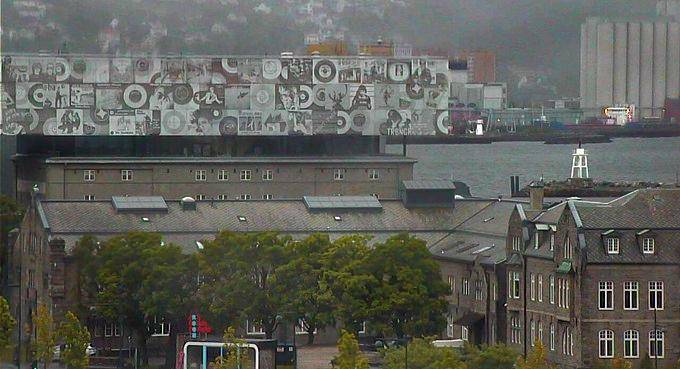} & \hspace{-0.4cm}
			\includegraphics[width = 0.16\textwidth]{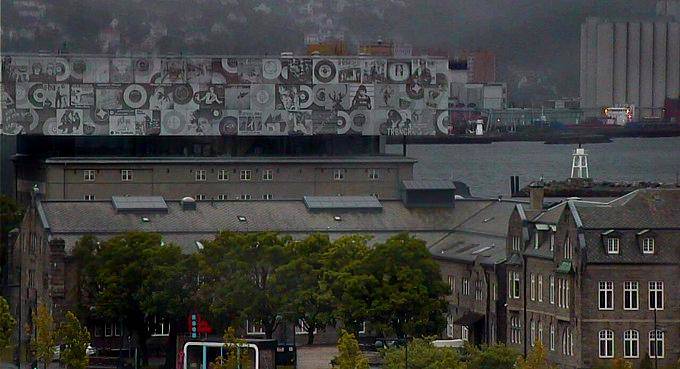} & \hspace{-0.4cm}
			\includegraphics[width = 0.16\textwidth]{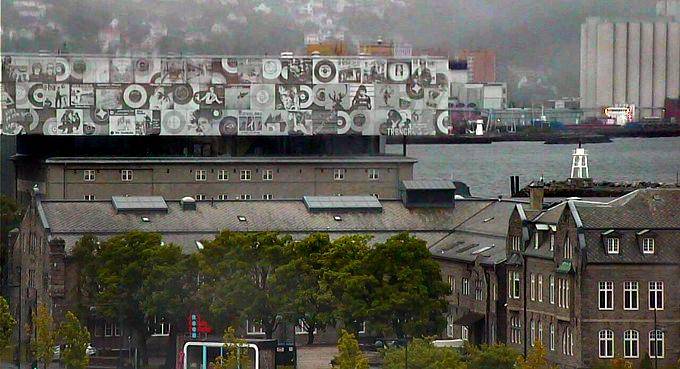}\\
			\textit{e}=0.2775,$\overline{\textit{r}}$=1.5026 &\hspace{-0.4cm}
			\textit{e}=0.3264,$\overline{\textit{r}}$=1.4168 & \hspace{-0.4cm}
			\textit{e}=0.3675,$\overline{\textit{r}}$=1.9205 \\
			\includegraphics[width = 0.16\textwidth]{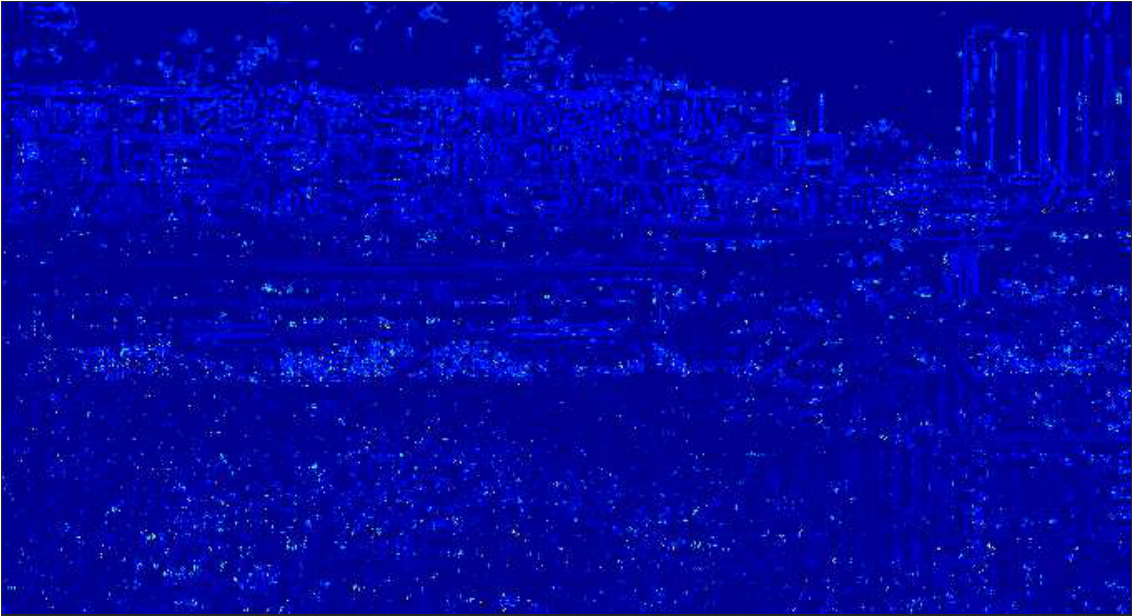} & \hspace{-0.4cm}
			\includegraphics[width = 0.16\textwidth]{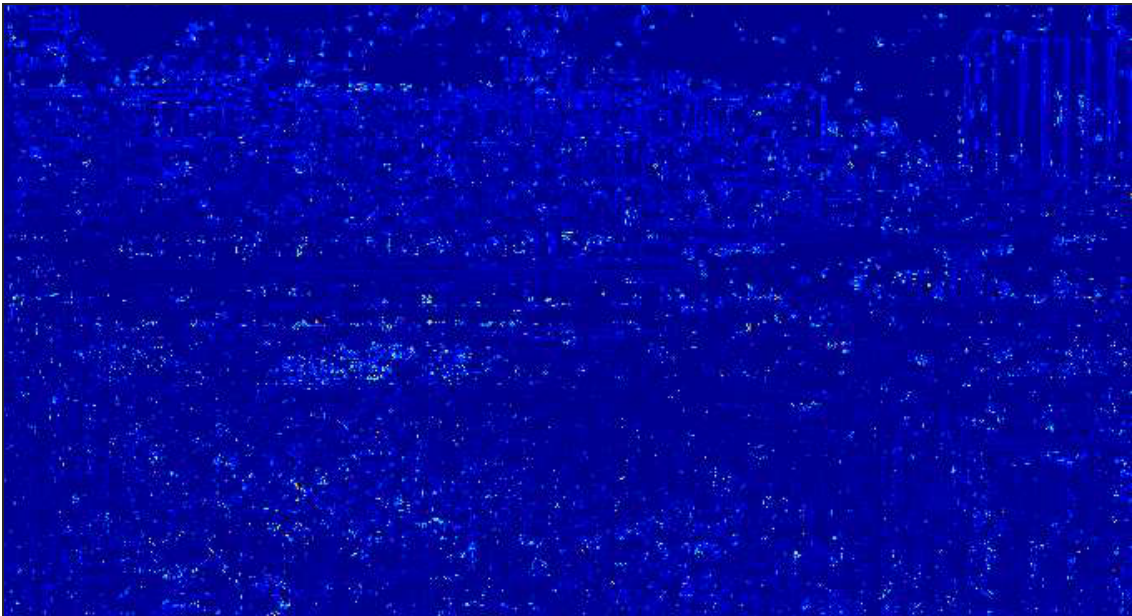} & \hspace{-0.4cm}
			\includegraphics[width = 0.16\textwidth]{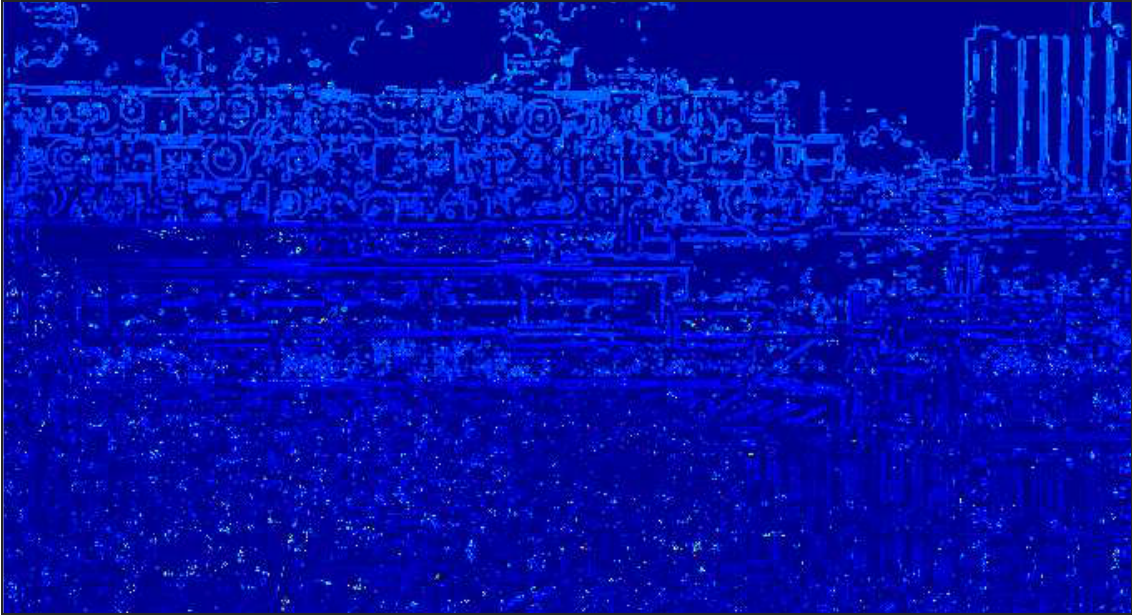} \\
			(b) He \cite{he2011PAMI}  & \hspace{-0.4cm}
			(c) Cai \cite{cai2016dehazenet} & \hspace{-0.4cm}
			(d) Ours
		\end{tabular}
	\end{center}
	\vspace{-0.3cm}
	\caption{Comparison of different defogging techniques on real world image of the MRFID. The 2nd row is the defogged results of (a) by different methods. The 3rd row is the visible gradient ratio maps for the corresponding results in 2nd row.
	}
	\vspace{-0.4cm}
	\label{MFID-ratio-compare}
\end{figure}
\begin{figure}[t]\footnotesize
	\begin{center}
		\begin{tabular}{@{}cccc@{}}
			\includegraphics[width = 0.12\textwidth]{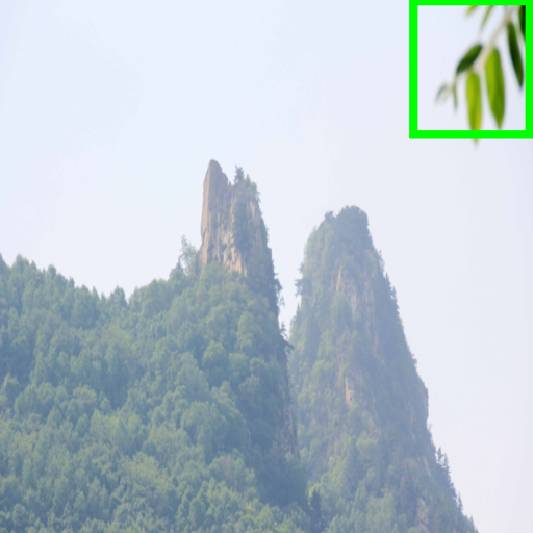} & \hspace{-0.4cm} 
			\includegraphics[width = 0.12\textwidth]{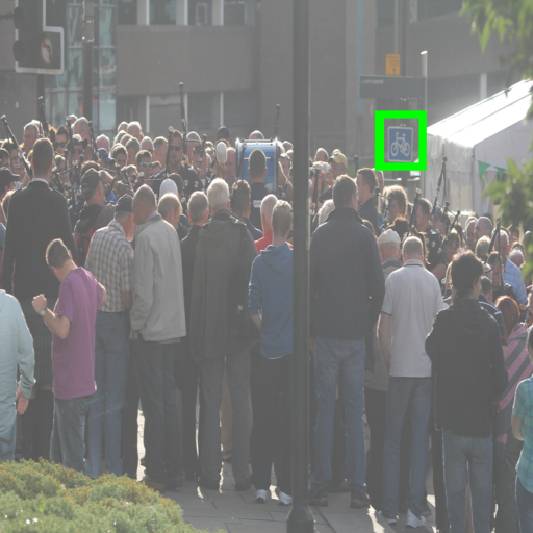}\\
			(a) & \hspace{-0.4cm}
			(b)
		\end{tabular}
	\end{center}
	\begin{center}
		\begin{tabular}{@{}cccc@{}}
			\includegraphics[width = 0.11\textwidth]{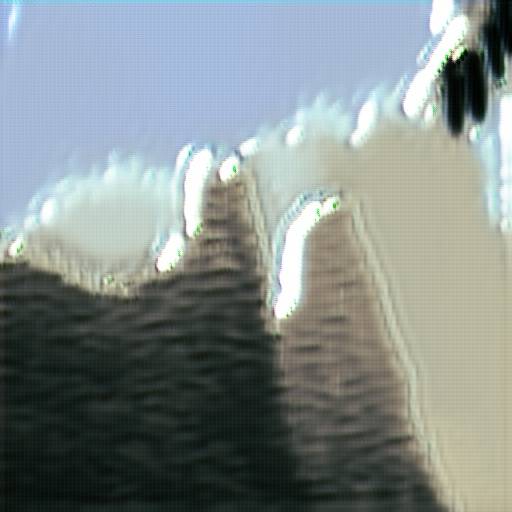} & \hspace{-0.4cm}
			\includegraphics[width = 0.11\textwidth]{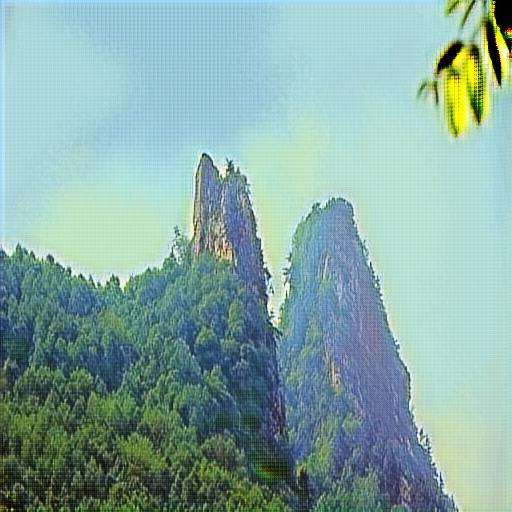} & \hspace{-0.4cm}
			\includegraphics[width = 0.11\textwidth]{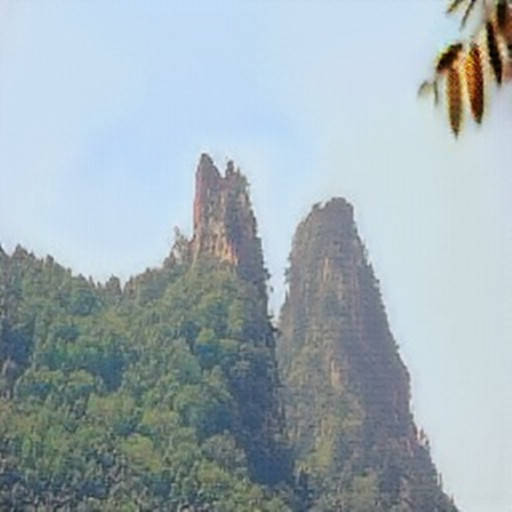} & \hspace{-0.4cm}
			\includegraphics[width = 0.11\textwidth]{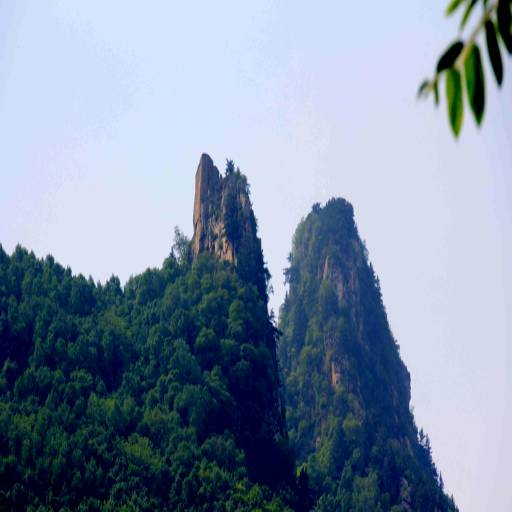} \\
			\includegraphics[width = 0.11\textwidth]{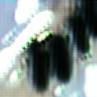}  & \hspace{-0.4cm}
			\includegraphics[width = 0.11\textwidth]{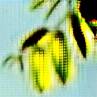}  & \hspace{-0.4cm}
			\includegraphics[width = 0.11\textwidth]{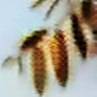}  & \hspace{-0.4cm}
			\includegraphics[width = 0.11\textwidth]{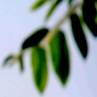} \\
			\includegraphics[width = 0.11\textwidth]{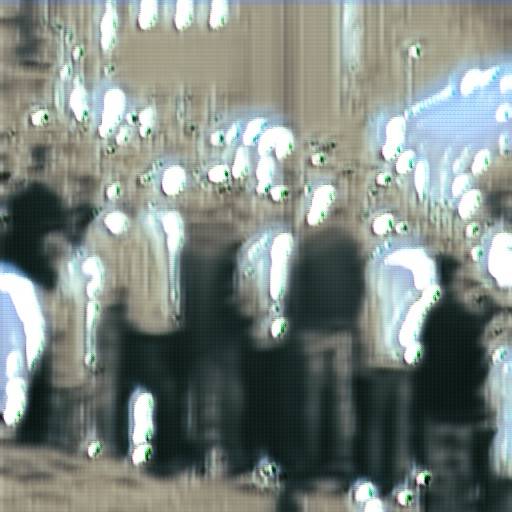} & \hspace{-0.4cm}
			\includegraphics[width = 0.11\textwidth]{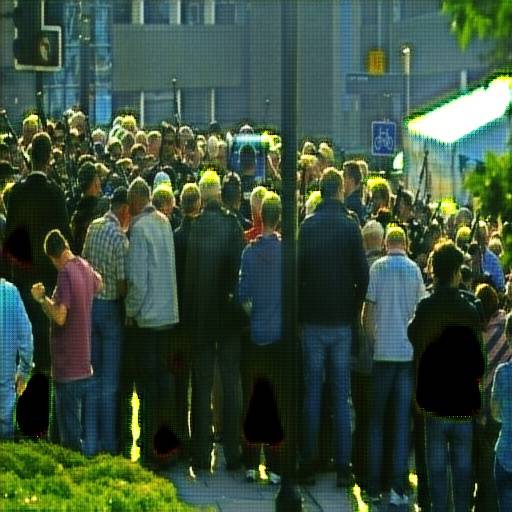} & \hspace{-0.4cm}
			\includegraphics[width = 0.11\textwidth]{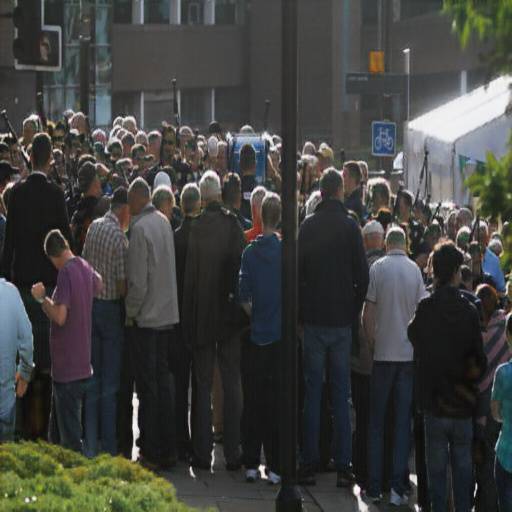} & \hspace{-0.4cm}
			\includegraphics[width = 0.11\textwidth]{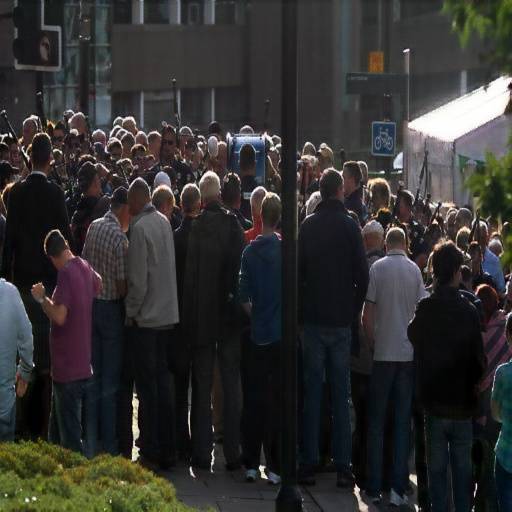} \\
			\includegraphics[width = 0.11\textwidth]{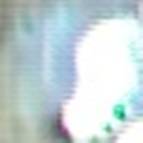}  & \hspace{-0.4cm}
			\includegraphics[width = 0.11\textwidth]{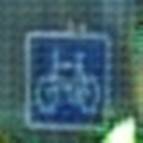}  & \hspace{-0.4cm}
			\includegraphics[width = 0.11\textwidth]{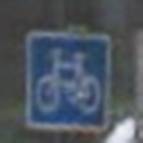}  & \hspace{-0.4cm}
			\includegraphics[width = 0.11\textwidth]{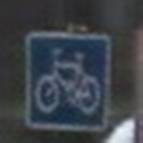} \\
			(c)  & \hspace{-0.4cm}
			(d)  & \hspace{-0.4cm}
			(e)  & \hspace{-0.4cm}
			(f) 
		\end{tabular}
	\end{center}
	\vspace{-0.5cm}
	\caption{Effectiveness of the proposed Cycle-Defog2Refog approach. The 2nd row is the defogged results of (a) by various methods. The 3rd row is an enlarged sub-image highlighted in (a) of the corresponding image in the row above. The 4th row is the defogged results of (b) by different methods. The 5th row is an enlarged sub-image highlighted in (b) of the corresponding image in the row above.
	}
	\vspace{-0.35cm}
	\label{loss-ablation}
\end{figure}
\textbf{Evaluation on real world images:} We evaluate our method on real world images which are provided by the authors of previous methods and available on the Internet \cite{fattal2014colorline}\cite{ren2018CVPR}. As shown in Figure \ref{learning-comparison}, we compare the proposed method with other learning-based methods \cite{cai2016dehazenet}\cite{MSCNN2016ECCV}. We can see that our results have a better recovered quality than others in contrast and sharpness. Figure \ref{real-results} shows the comparison of defogged results by our method and other five defog methods. As shown in Figure \ref{real-results} (b)-(c), there are still some artifacts in He's and Meng's results, e.g. color distortion in the sky region in the R2 column. In addition, the defogged results by He's method \cite{he2011PAMI} are darker than other methods’ due to overestimating the fog density. The results of Zhu \cite{zhu2015TIP} as shown in Figure \ref{real-results} (d) have a good performance, however, the defogged images are darker than ours in luminance. The results of Cai\cite{cai2016dehazenet} and Ren\cite{MSCNN2016ECCV} as shown in Figure \ref{real-results} (e)-(f) are similar. However, there are still some remaining fog which are not removed from the image. In contrast, the results of our method as shown in Figure \ref{real-results} (g) are more natural and clearer. It is also reflected in the quantitative results in Table \ref{real-quantitative} and Figure \ref{D-real-quantitative}. Regarding indicator $\delta$, its value always maintained in a small range, which means that our results have a good saturation. Regarding indicators \textit{e} and $\overline{\textit{r}}$, our results have higher values than others. It demonstrates that the local contrast and edge information are recovered well in our results. Moreover, as shown in Figure \ref{D-real-quantitative}, we can see that the FADE values of our results are smaller than other methods. It turns out that the texture details of our defogged results are recovered felicitously and clearer.
\begin{figure}[t]\footnotesize
	\begin{center}
		\begin{tabular}{@{}cccc@{}}
			\includegraphics[width = 0.11\textwidth]{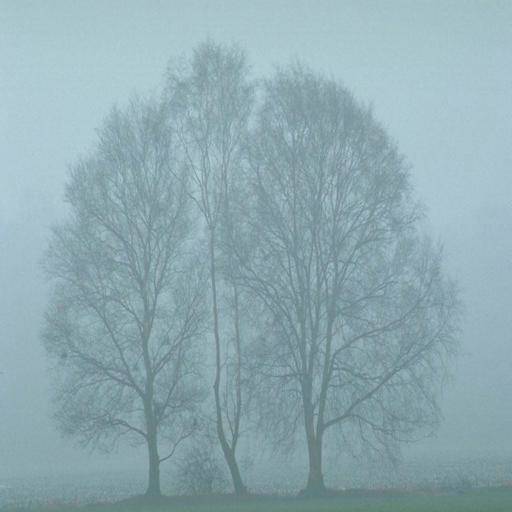} & \hspace{-0.4cm}
			\includegraphics[width = 0.11\textwidth]{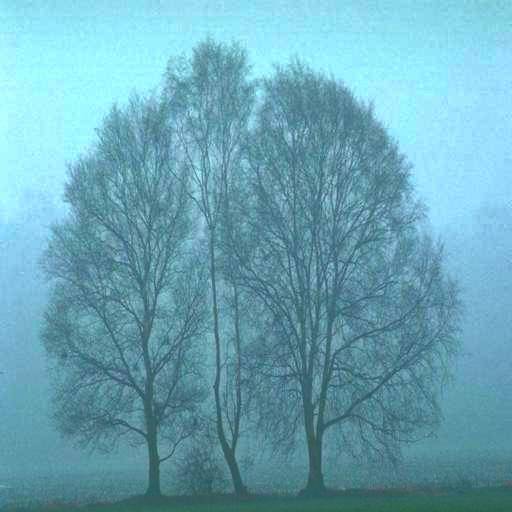} & \hspace{-0.4cm}
			\includegraphics[width = 0.11\textwidth]{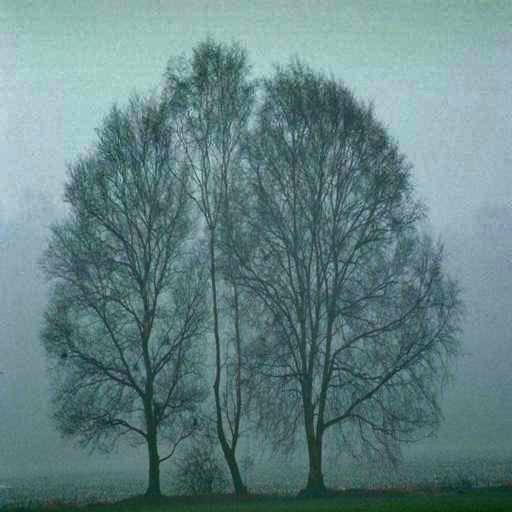} & \hspace{-0.4cm}
			\includegraphics[width = 0.11\textwidth]{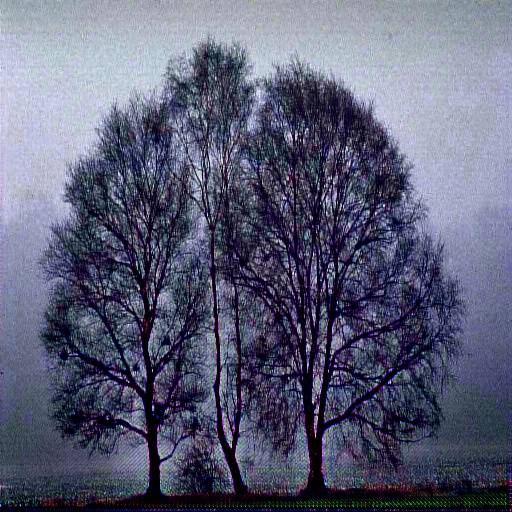} \\
			\includegraphics[width = 0.11\textwidth]{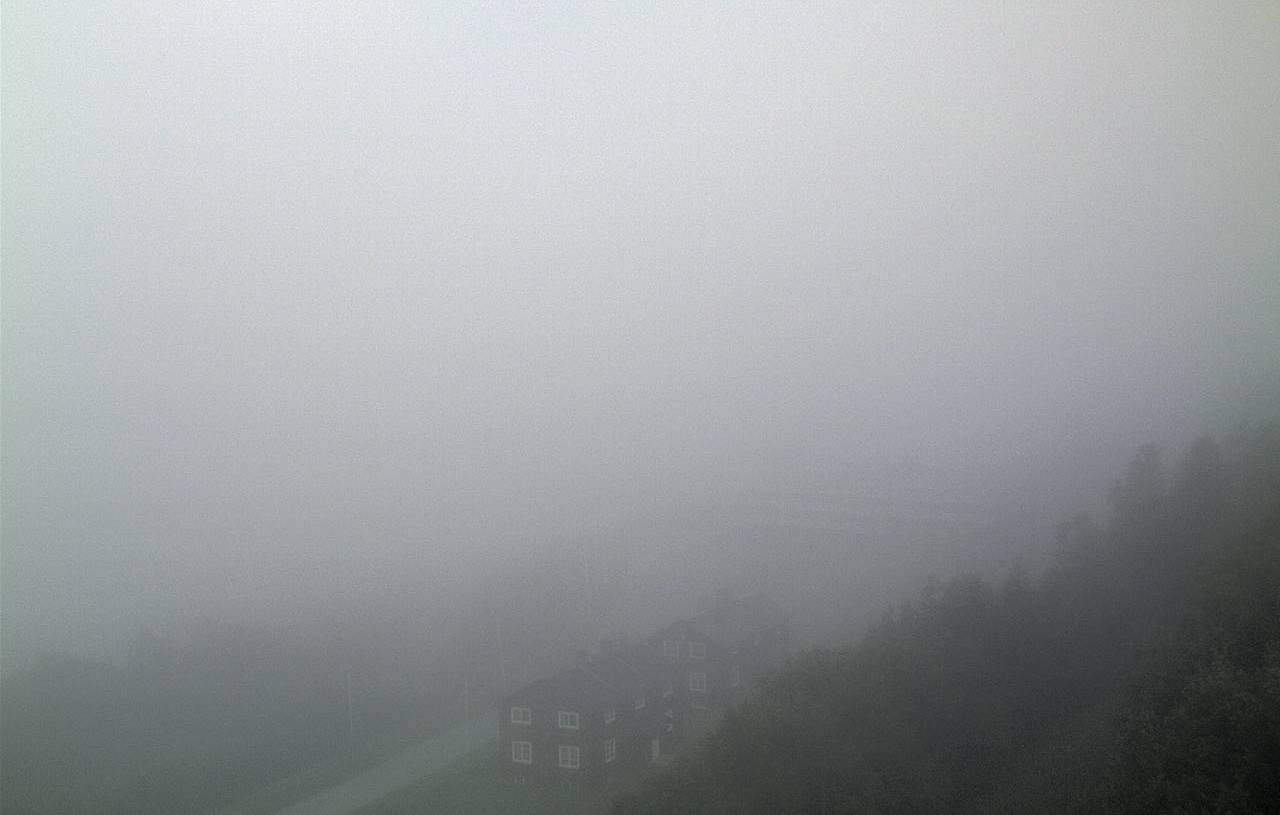} & \hspace{-0.4cm}
			\includegraphics[width = 0.11\textwidth]{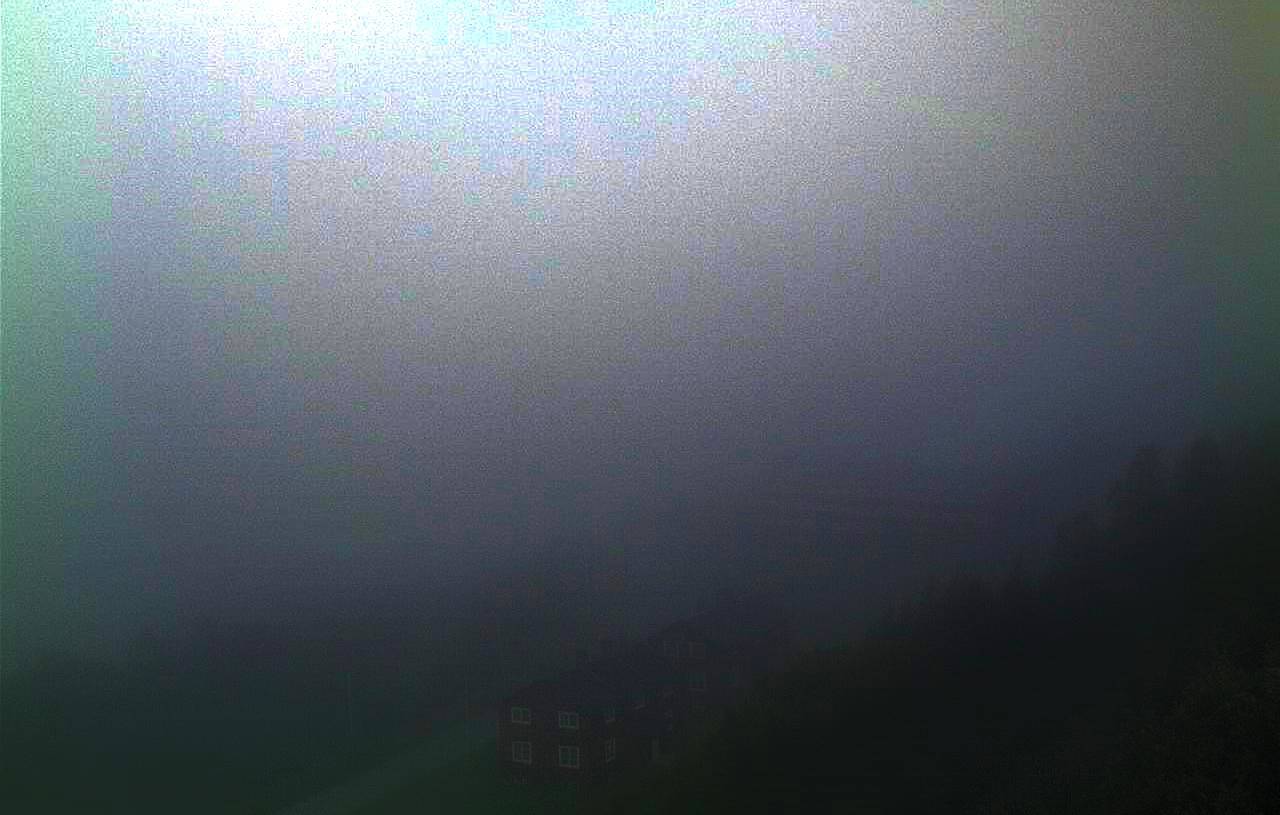} & \hspace{-0.4cm}
			\includegraphics[width = 0.11\textwidth]{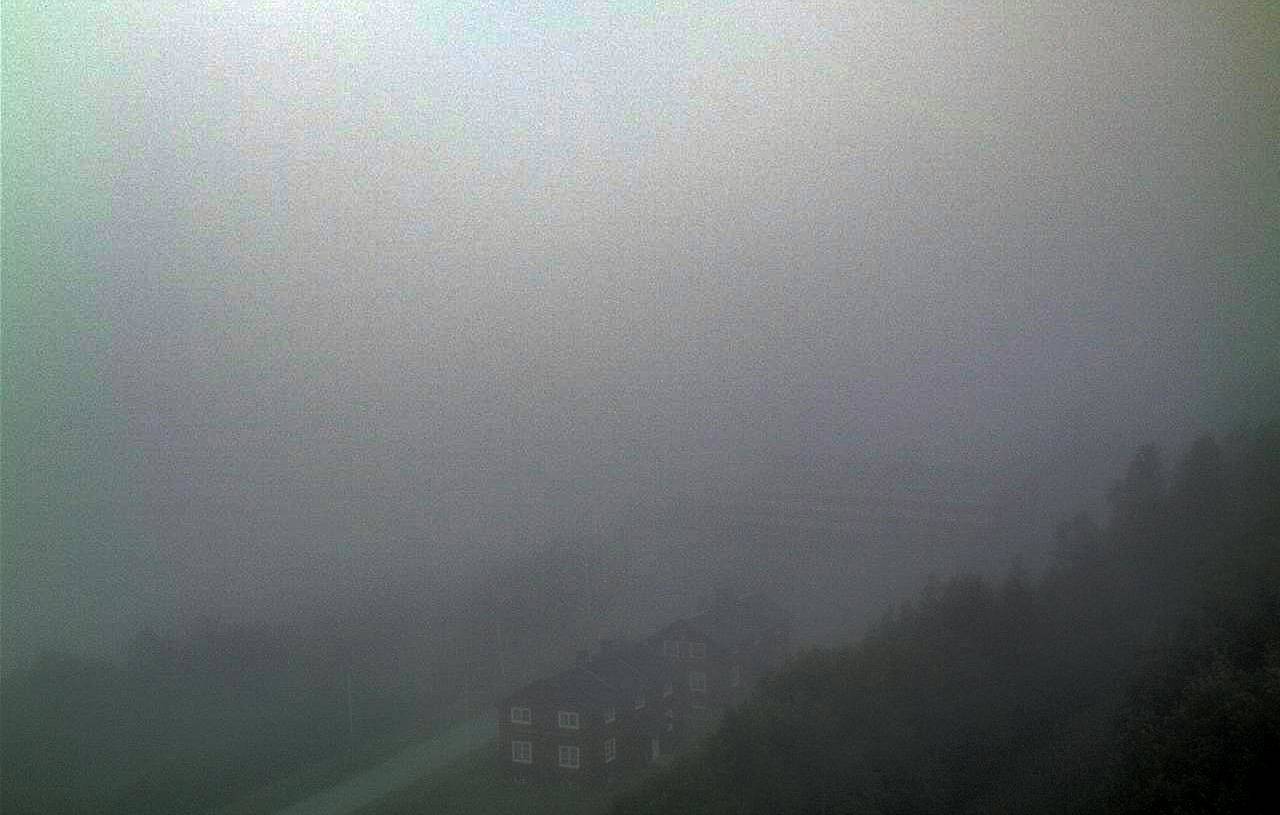} & \hspace{-0.4cm}
			\includegraphics[width = 0.11\textwidth]{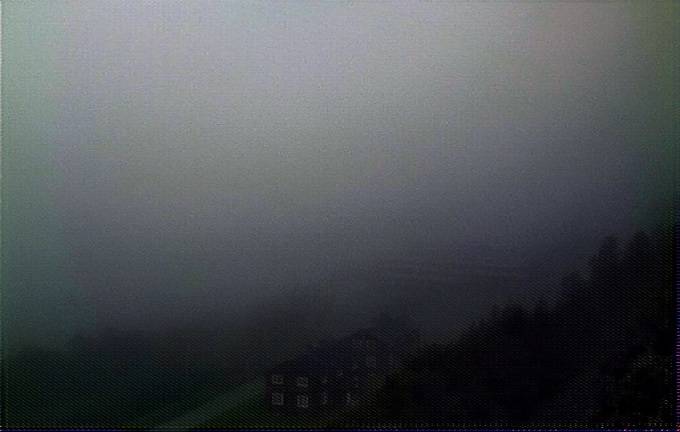} \\
			(a) Foggy inputs & \hspace{-0.4cm}
			(b) He \cite{he2011PAMI} & \hspace{-0.4cm}
			(c) Ren \cite{MSCNN2016ECCV}& \hspace{-0.4cm}
			(d) Our
		\end{tabular}
	\end{center}
	\vspace{-0.5cm}
	\caption{The proposed approach and the state-of-the-art approaches do not work well for the foggy images of heavy fog density.
	}
	\vspace{-0.5cm}
	\label{limitations-ablation}
\end{figure}

\textbf{MRFID dataset.} We further evaluate the proposed method against other five defog methods \cite{he2011PAMI} \cite{zhu2015TIP} \cite{meng2013ICCV} \cite{cai2016dehazenet} \cite{MSCNN2016ECCV} on our real foggy dataset MRFID. As shown in Figure \ref{MFID-results}, we can see that our method is superior to the others in sharpness and brightness. We further discuss this in Figure \ref{MFID-ratio-compare}, the rate of new visible edges $\textit{e}$ and the quality value of the contrast restoration $\overline{\textit{r}}$ are larger than other methods, which means our method has more competitive results in local contrast and edge information. Moreover, we can see that our method has a stronger visible gradient ratio than others (as shown in the third row of Figure \ref{MFID-ratio-compare}), which is reflected by the indicator $\overline{\textit{r}}$. It is shown that our method is less prone to artifacts, the generated defogged image is more natural and has a high contrast. We also quantitatively evaluate our method on MRFID in Table \ref{MFID-quantitative}. As shown, our method has a better defog performance compared with the other methods\cite{he2011PAMI} \cite{zhu2015TIP} \cite{meng2013ICCV} \cite{cai2016dehazenet} \cite{MSCNN2016ECCV} in this dataset.

\section{Analysis and Discussions}
\subsection{Effectiveness of Cycle-Defog2Refog Network}
In this section, we analyse how the Cycle-Defog2Refog network is more effective than the original CycleGAN in recovering the clear image from the foggy image. We compared the defogged results by using CycleGAN and our method, as shown in Figure \ref{loss-ablation}. In Figure \ref{loss-ablation} (c)-(d), the results are generated by both methods only using the generative adversarial loss function. We can see that the CycleGAN's result (as shown in Figure \ref{loss-ablation} (c)) has been distorted and the texture details are completely lost. In contrast, as shown in Figure \ref{loss-ablation} (d), although the color distortion appears in the defogged results of our method, the texture details are completely preserved. It is due to we have adopted a refog network based on atmospheric degeneration model to strengthen the mapping function, so that we can preserve more texture details for the defogged results. Moreover, Figure \ref{loss-ablation} (e) and Figure \ref{loss-ablation} (f) show the defogged results of CycleGAN and our algorithm by using full loss function. As shown the zoom-in regions in Figure \ref{loss-ablation} (e), the color of the leaves is distorted, and the bicycle sign is blurred. In our results, there is no color distortion on the leaves, and the bicycle sign has clearer edges as shown in Figure \ref{loss-ablation} (f).

\subsection{Limitations}
As with previous approaches, there are limitations in the proposed approach. One such limitation is that Cycle-Defog2Refog cannot handle heavy fog. As shown in Figure \ref{limitations-ablation}, in the first row, our method as well as those of He\cite{he2011PAMI} and Ren\cite{MSCNN2016ECCV} did not handle this image well; In the second row, all the methods have failed to deal with the heavy fog image. The reason is that current atmospheric degradation model can no longer accurately describe the foggy map in this case. In future work, we would like to solve this problem by dedicating to optimize this model and build a multifarious of foggy dataset to train our network.

\section{Concluding Remarks}

In this paper, we adopt a cycle generative adversarial network for single image defogging. Our method is trained by unpaired image data, which avoids preparing a large number of synthetic foggy images in advance. To generate a more real and clearer image, we have proposed a new refog network based on physical model and a new enhancer network to supervise the mapping from fog domain to fog-free domain. In refog network, we further presented a sky prior to estimate the atmospheric light to prevent artifacts, such as color distortion. Moreover, we introduce a new foggy dataset to train and evaluate our approach, it includes 200 clear outdoor images and 800 foggy images with different fog density. Extensive experiments on both synthetic and real-world foggy images demonstrate that our method performs favorably against several state-of-the-art methods.

{\small
\bibliographystyle{ieee}
\bibliography{refog2defog}
}

\end{document}